\pdfoutput=1
\documentclass[10pt]{article}

\usepackage[letterpaper,textwidth=6.75in,textheight=9.25in,top=0.9in,
            headheight=14pt,headsep=20pt,footskip=30pt]{geometry}
\usepackage[numbers]{natbib}
\usepackage{fancyhdr}
\fancypagestyle{plain}{\fancyhf{}\fancyfoot[C]{\thepage}}
\makeatletter
\renewcommand{\section}{\@startsection{section}{1}{\z@}%
  {-2.0ex \@plus -0.5ex \@minus -0.2ex}{1.5ex \@plus 0.3ex \@minus 0.2ex}%
  {\large\bfseries\raggedright}}
\renewcommand{\subsection}{\@startsection{subsection}{2}{\z@}%
  {-1.8ex \@plus -0.5ex \@minus -0.2ex}{0.8ex \@plus 0.2ex}%
  {\normalsize\bfseries\raggedright}}
\renewcommand{\subsubsection}{\@startsection{subsubsection}{3}{\z@}%
  {-1.5ex \@plus -0.5ex \@minus -0.2ex}{0.5ex \@plus 0.2ex}%
  {\normalsize\bfseries\raggedright}}
\renewcommand{\paragraph}{\@startsection{paragraph}{4}{\z@}%
  {1.5ex \@plus 0.5ex \@minus 0.2ex}{-1em}{\normalsize\bfseries}}
\g@addto@macro\normalsize{%
  \setlength\abovedisplayskip{7pt plus 2pt minus 4pt}%
  \setlength\belowdisplayskip{7pt plus 2pt minus 4pt}%
  \setlength\abovedisplayshortskip{0pt plus 2pt}%
  \setlength\belowdisplayshortskip{4pt plus 2pt minus 2pt}}
\renewcommand{\@maketitle}{%
  \newpage\null\vskip -2.5em
  \begin{center}
    \rule{\linewidth}{2pt}\par\vskip 0.9em
    {\LARGE\bfseries \@title\par}
    \vskip 0.7em\rule{\linewidth}{0.6pt}\par\vskip 1.6em
    {\normalsize \@author\par}
  \end{center}
  \par\vskip 1em}
\makeatother
\usepackage[font=small,labelfont=bf]{caption}
\usepackage[ragged]{footmisc}

\PassOptionsToPackage{colorlinks=true,linkcolor=blue!60!black,
  citecolor=green!45!black,urlcolor=magenta!60!black}{hyperref}
\usepackage{algorithm}
\usepackage{algpseudocode}
\newcommand{\theouralgorithm}{\arabic{algorithm}}

\newenvironment{ouralgorithm}[1]{%
  \refstepcounter{algorithm}%
  \par\addvspace{1ex}%
  \noindent\rule{\linewidth}{0.8pt}\\[4pt]%
  \noindent\textbf{Algorithm~\theouralgorithm}\quad #1\\[2pt]%
  \noindent\rule{\linewidth}{0.4pt}%
  \par\addvspace{0.5ex}%
}{%
  \par\addvspace{0.5ex}%
  \noindent\rule{\linewidth}{0.8pt}%
  \par\addvspace{1ex}%
}

\newcommand{\ResetAlgLine}{\setcounter{ALG@line}{0}}
\usepackage[utf8]{inputenc}
\usepackage[T1]{fontenc}
\usepackage{hyperref}
\usepackage{url}
\usepackage{booktabs}
\usepackage{amsfonts}
\usepackage{amsmath, amssymb, amsthm}
\usepackage{nicefrac}
\usepackage{microtype}
\usepackage{xcolor}
\usepackage{graphicx}
\usepackage{tikz}
\usepackage{multirow}
\usetikzlibrary{arrows.meta, positioning, fit, backgrounds, calc, decorations.pathreplacing}
\usepackage{pifont}

\usepackage{float}

\usepackage{etoolbox}
\usepackage{adjustbox}
\BeforeBeginEnvironment{tabular}{\begin{adjustbox}{max width=\linewidth}}
\AfterEndEnvironment{tabular}{\end{adjustbox}}
\makeatletter
\def\appsecformat#1{\expandafter\ifx\csname c@#1\endcsname\c@section
  Appendix~\fi\csname the#1\endcsname\quad}
\g@addto@macro\appendix{\gdef\thesection{\@Alph\c@section}%
  \global\let\@seccntformat\appsecformat
  \setcounter{equation}{0}%
  \@addtoreset{table}{section}\@addtoreset{figure}{section}%
  \@addtoreset{equation}{section}}
\makeatother

\begin{document}

\title{Hamilton--Jacobi Networks: Unifying Conserved-Quantity Discovery
and Symplectic Trajectory Prediction}

\author{%
  \begin{tabular}[t]{c}
    \textbf{Thomas Nektarios Saltzis}\\
    \texttt{tsaltzi@csd.auth.gr}
  \end{tabular}
  \hspace{3em}
  \begin{tabular}[t]{c}
    \textbf{Anastasios Tefas}\thanks{Corresponding author.}\\
    \texttt{tefas@csd.auth.gr}
  \end{tabular}\\[0.9em]
  Department of Informatics, Aristotle University of Thessaloniki\\
  54124 Thessaloniki, Greece}
\date{}
\maketitle

\begin{abstract}
Hamilton--Jacobi theory unifies two objectives of learning Hamiltonian dynamics, symplectic prediction and conserved-quantity discovery: it seeks a time-dependent canonical transformation that carries every state of a trajectory to the same conserved quantities and whose inverse recovers the state at any time. Learning the dynamics thus becomes representation learning. For autonomous Hamiltonian systems with complete flow, such a transformation exists, integrable or not, as a single globally defined map that neural networks can parameterise. Our Hamilton--Jacobi Network parameterises this map and its inverse as encoder and decoder, canonical and exactly invertible by construction, as the theory requires, and is trained to predict each trajectory segment from one observation. We prove that, on the training data, its prediction loss vanishes precisely when its latents are each segment's conserved quantities, a tie that a two-way bound, up to local constants, extends to any loss and segment. Prediction and discovery thus become one objective, prescribing no form for the conserved quantities. A time-independent canonical map of latents, their frame, provably leaves predictions unchanged. At the training energies of an oscillator, a pendulum and non-integrable H\'enon--Heiles, most unseen trajectories' frame-free latents vary in rms by less than one training-noise vector. In the frame that training settles on, they are operationally interpretable: low-degree polynomials, fit at time zero, recover the energy (cross-validated $R^2\ge0.944$) and, in the integrable systems, the initial phase's circle coordinates ($R^2>0.99$), and approximately reproduce their response to time shifts (median errors: energy $\le5.2\%$, phase rate $\lesssim1.3\%$).
\end{abstract}

\section{Introduction}
\label{sec:introduction}

Hamiltonian systems govern a wide range of physical phenomena, from
celestial mechanics and molecular dynamics to plasma physics, and learning
their dynamics from trajectory data has become an active area of
scientific machine learning. Unlike generic sequence data, these systems
carry geometric structure: their flow preserves the symplectic form, and they admit conserved
quantities that constrain the accessible regions of phase space~\cite{arnold1989mathematical}. A learned model that inherits this structure can make predictions faithful to the geometry of the true dynamics, and it can expose the quantities
the motion preserves.

This has motivated models that build the structure in or discover it
from data. One line of work builds Hamiltonian structure into
the model by learning the Hamiltonian itself, whose equations of motion
are then integrated at inference time~\cite{Greydanus2019,David2023}. A second line learns the flow map directly and makes it symplectic by
construction, as the true flow is~\cite{Jin2020,Horn2025,Canizares2024}. A third line identifies conserved quantities from trajectory data, either
independently of
prediction~\cite{Bondesan2019,Liu2021AIPoincare,Chen2024Deflation} or
jointly with it. Some joint approaches make no Hamiltonian-structure
assumptions, which extends them to broader system
classes but provides no symplectic guarantees when the target system is Hamiltonian~\cite{matsubara2022finde,alet2021noether,jin2022learning}.
Within the Hamiltonian framework, conserved quantities beyond the
Hamiltonian itself have been incorporated into prediction with their form
fixed in advance, either as quadratic forms~\cite{van2024noether} or
through an action--angle ansatz~\cite{Daigavane2022}.

We propose the \textbf{Hamilton--Jacobi Network (HJN)}, in which
symplectic prediction and the discovery of conserved quantities are unified under
one objective. The motivating observation comes from Hamilton--Jacobi theory, which seeks not a rule for stepping forward in time but a canonical transformation carrying every state of a trajectory to the same set of conserved quantities, from which the state at any requested time is recovered by inverting it~\cite[\S10.1]{goldstein2002classical}\cite[\S47]{arnold1989mathematical}. Because these quantities fully encode the trajectory, learning the
dynamics is recast as representation learning, with the conserved
quantities as the representation.
Discovery and prediction then use the same canonical transformation, in
opposite directions. For autonomous Hamiltonian systems whose flow is complete, such a transformation exists whether or not the system is integrable, and all such transformations share a common form: a single globally defined map that a neural network can parameterise (Section~\ref{subsec:problem_formulation}).

HJN parameterises the transformation and its inverse as a tied-weight encoder--decoder (Figure~\ref{fig:overview}) that enforces by construction the two properties the Hamilton--Jacobi form requires, canonicity and exact invertibility: the encoder composes symplectic shear maps, following~\cite{Jin2020,Horn2025}, with time as a continuous input, as in SympFlow~\cite{Canizares2024}, and the decoder is its exact algebraic inverse, so every predicted map is symplectic. HJN is trained to predict each trajectory segment, a contiguous portion of
bounded duration, from a single observation of it. Since every prediction shares the latents of one observed state and differs only in its query time, the decoder predicts the states at all requested times within a segment in a single forward pass and without a numerical integrator. Longer horizons are covered by composing segments. We show that, on the states and times the training data covers, HJN
achieves zero prediction loss precisely when its encoder takes the form
Hamilton--Jacobi theory calls for, that is, when its latents are each segment's conserved quantities, whose form nothing prescribes. At any loss, and on any segment of any trajectory, the departure of the
latents from constancy and the prediction error on that segment bound
each other, up to local constants, so the tie between prediction and discovery holds away from zero prediction loss as
well.

The theory bounds the method's scope: autonomous Hamiltonian systems with
complete flow, observed as conjugate coordinate--momentum pairs. Our experiments use three systems within it: two integrable ones (the ideal mass--spring oscillator and the ideal
pendulum) and the non-integrable H\'enon--Heiles system. On each, we train three models with different random seeds. We show that the prediction objective determines the latents only up to a
time-independent canonical transformation, which we call their frame. Two complementary experiments measure how closely these models, at the nonzero prediction loss that training reaches, carry the structure that zero prediction loss forces
exactly on the training data, both there and on states not seen during training. Measured in the frame-free setting the latents come close to constancy on
held-out trajectories at the training energies in all three systems: for
the median trajectory, they deviate from their segments' frame-free conserved quantities, in rms, by less than the rms length of the noise
vector added to each observation during training, for every trained model. In the frame that training settles on, across the training energy range and for every trained model, a polynomial of degree at most four in the latents approximates the energy (cross-validated $R^2 \ge 0.944$), and degree-one polynomials approximate the circle coordinates of the
initial phase in the two integrable systems ($R^2 > 0.99$). These read-outs, fit at $t=0$ alone, respond to shifts of the encoding
time approximately as the classical conserved quantities do: the recovered energy has
median relative errors of at most $5.2\%$, and the recovered phase
advances approximately at each integrable system's own rate, in the median to within $0.2\%$ of it
for the oscillator and about $1.3\%$ for the pendulum. The latents are thus interpretable in an operational sense. The theory and both experiments concern single segments. We additionally evaluate HJN over rollouts that compose them from a single observation, where its error, on every test set of these systems, is comparable to or below that of a reference method that also
learns conserved quantities~\cite{van2024noether}, under the same protocol.

\section{Related Work}
\label{sec:related_work}

Methods for learning dynamics from observational data differ in what mathematical object the network is asked to learn, a choice that determines the prediction mechanism and the structural guarantees.

\paragraph{Learning the Hamiltonian and Lagrangian functions.}
Methods of this family parameterise the Hamiltonian or Lagrangian as a
scalar-valued MLP, recover equations of motion via automatic
differentiation, and integrate the resulting ODE numerically at inference.
Preservation of the symplectic structure is then a property of the integrator rather than of the learned function.

Hamiltonian Neural Networks (HNNs)~\cite{Greydanus2019} introduced this paradigm; they learn a conserved quantity, the Hamiltonian, without prescribing its form. David and M\'{e}hats~\cite{David2023} showed that the forward-Euler loss introduces an irreducible $O(h^{2})$ floor, and proposed Symplectic HNNs (SHNNs) that
remove it by matching the loss to a symplectic integrator. Neural Symplectic Form~\cite{chen2021neural} learns the symplectic 2-form defining the
coordinate-free Hamilton equation, extending coverage to
systems with non-standard symplectic structures. Lagrangian Neural Networks (LNNs)~\cite{cranmer2020lagrangian} parameterise $\mathcal{L}(q,\dot{q})$ and recover accelerations by inverting the Euler--Lagrange equations, avoiding the need for canonical momenta.

Additional methods extend this paradigm. Nonseparable Symplectic Neural Networks (NSSNNs)~\cite{xiong2020nonseparable} embed an augmented-phase-space symplectic
integrator into the architecture, making the internal update map
symplectic by construction, while scaling to $N$-body systems via pairwise interaction networks. Bacsa et al.~\cite{bacsa2023symplectic} replace the encoder of a Deep Markov Model with a symplectic Neural ODE, learning a latent phase-space representation with symplectically structured inference trajectories. However, the symplectic constraint is confined to the encoder; forward prediction uses an unconstrained MLP. Reduced-order HNNs (RO-HNNs)~\cite{Friedl2025LearningHD} target
high-dimensional systems by jointly learning a low-dimensional symplectic
submanifold via a cotangent-lifted symplectic autoencoder and the reduced
Hamiltonian dynamics on it via a latent HNN, integrated at inference with a
symplectic scheme. 

\paragraph{Learning the flow map.}
Since the flow of a Hamiltonian system is symplectic, a second family of approaches learns the flow map $\phi_h$ directly and makes it symplectic by construction. A single forward pass advances $(q,p)$ by a fixed step $h$, producing predictions at integer multiples of $h$. 

SympNets~\cite{Jin2020} initiated this family by composing
elementary shear maps whose
symplectic property is guaranteed for any $C^{1}$ generating function $V$ and
proved universal approximation theorems of arbitrary symplectic maps. Each module parameterises $\nabla V$ with a single hidden layer, which eliminates automatic differentiation from the forward pass. Generalized HNNs (GHNNs)~\cite{Horn2025} unify SympNets,
H\'{e}nonNets~\cite{Burby2020}, and Symplectic RNNs~\cite{chen2020SRNN} under a
single framework of chained symplectic integration steps on independently learned separable Hamiltonians. GHNNs use deeper MLPs for the generating functions $V$, which increases per-module expressivity. Generating Function Neural Networks (GFNNs)~\cite{Chen2021GFNN} approximate the type-2 generating function of the symplectic map with an unconstrained network and obtain each prediction step by an implicit iterative solve, proving a linear error-growth bound for integrable systems; the type-2 choice fixes which variables are taken as independent. SympFlow~\cite{Canizares2024} extends this family to continuous time by
accepting time as an explicit input. Each sublayer is the exact flow of a
time-dependent separable Hamiltonian. Prediction at any query time within a training window $[0,\,\Delta t]$ requires only a single forward pass, while sequential composition is needed beyond $\Delta t$.

\paragraph{Discovery of conserved quantities.}
A parallel line of work identifies conserved quantities from trajectory data.
Bondesan and Lamacraft~\cite{Bondesan2019} use symplectic
normalising-flow layers to learn action--angle transformations, the action--angle
ansatz prescribing the form of the invariants and confining the method to
integrable systems, while Li et al.~\cite{Li2020NCT} restrict to point
transformations, pushing the latent distribution toward independent harmonic
oscillators with learnable frequencies. AI Poincar\'{e}~\cite{Liu2021AIPoincare}
uses a neural technique to estimate the number of independent
conservation laws, while
Chen et al.~\cite{Chen2024Deflation} discover complete sets of functionally
independent, Poisson-commuting invariants via neural deflation with an HNN
surrogate. In all four cases, the discovered quantities are disconnected
from trajectory prediction.

\paragraph{Connecting discovery with prediction.}
Another line of work connects conserved-quantity discovery
directly with trajectory prediction, splitting into two groups by the geometric assumptions imposed.

\textit{Methods outside the Hamiltonian framework.}
The following methods make no Hamiltonian-structure assumptions, which extends them to broader system classes but provides no symplectic
guarantees when the target system is Hamiltonian. FINDE~\cite{matsubara2022finde} jointly learns dynamics and $K$ candidate first
integrals: a neural network outputs scalar quantities
$V_1,\dots,V_K$, and the learned vector field is projected onto the tangent space of their joint level set, so that $\tfrac{d}{dt}V_k = 0$, and inference uses a numerical integrator or implicit solve. Joint training of the base dynamics and the
invariant network encourages the $V_k$
to align with true first integrals. The method improves long-horizon prediction empirically, but $K$ must be chosen in advance. Noether Networks~\cite{alet2021noether} learn a conserved embedding and enforce a conservation loss $\sum_t \lVert g_\phi(x_0)-g_\phi(\tilde{x}_t)\rVert^2$ via an inner-loop update during prediction, improving long-term accuracy across diverse settings including raw-video prediction. Their inference properties are inherited from the backbone predictor. Poisson Neural Networks (PNNs)~\cite{jin2022learning} learn Poisson dynamics via a Darboux--Lie decomposition: an invertible network maps to latent canonical and Casimir coordinates, an extended SympNet advances the canonical component while holding Casimirs fixed, and the inverse map returns to original variables; predictions proceed by sequential composition at a fixed step size, and Poisson structure is preserved by construction. However, the number of Casimirs must be chosen in advance.

\textit{Methods within the Hamiltonian framework.}
The following methods operate within the Hamiltonian formalism and incorporate conserved quantities beyond the Hamiltonian into the model that generates predictions, but in each case prescribe their form a priori. Action-Angle Networks~\cite{Daigavane2022} learn a symplectic
mapping to action--angle variables and evolve the angles linearly in the latent
space, making inference cost independent of the prediction horizon. However, constant
actions and linearly evolving angles are valid only for integrable systems,
prescribing the latent representation a priori. Noether's Razor~\cite{van2024noether} takes a Bayesian model-selection perspective: conserved quantities parameterise Hamiltonian symmetries via Noether's theorem, and an approximate marginal-likelihood objective selects those supported by the data, avoiding collapse to trivial constants without manual symmetry regularisation. As in its HNN backbone, predictions are obtained by numerical integration, so preservation of the symplectic structure is a property of the integrator. The method recovers correct symmetry structures and improves HNN generalisation, but restricts conserved quantities to quadratic forms.

\paragraph{Positioning of the present work.}
The observation at the core of HJN is that Hamilton--Jacobi theory unifies symplectic prediction with the discovery of conserved quantities. The theory seeks a time-dependent canonical transformation that carries every state of a trajectory to the same conserved quantities $(\alpha,\beta)$ and whose inverse recovers the state at any requested time; these quantities, one per phase-space dimension, determine the trajectory. This transformation carries no integrability requirement: it exists for every autonomous Hamiltonian system whose flow is complete, and, as a single globally defined map, it is one that a neural network can parameterise. HJN takes it as its target. It parameterises the transformation and its inverse as a tied-weight encoder--decoder that enforces by construction the two properties the Hamilton--Jacobi form requires, canonicity and exact invertibility: the encoder composes symplectic shear maps, following~\cite{Jin2020,Horn2025}, with time as a continuous input, as in SympFlow~\cite{Canizares2024}, and the decoder is its exact algebraic inverse, so every predicted map is symplectic by construction. We make the opening observation precise for this architecture (Section~\ref{sec:convergence}): on the states and times the training data covers, HJN achieves zero prediction loss precisely when its encoder takes the Hamilton--Jacobi form, and at any loss, on any segment of any trajectory, the departure of the latents from constancy and the prediction error on that segment bound each other, up to local constants. Prediction and the discovery of conserved quantities thus become one objective, while the form those quantities take is never prescribed.

\section{Proposed Method}
\label{sec:proposed_method}

\subsection{Problem Formulation and Physical Motivation}
\label{subsec:problem_formulation}
Given trajectory data $(q, p, t)$ from a Hamiltonian system, we seek a model that
predicts future states while preserving symplectic
structure and discovering conserved quantities from data alone. Classical mechanics offers a framework that unifies these objectives. We first review it, then show how it motivates a
neural architecture.

Consider an $n$-degree-of-freedom dynamical system with generalised coordinates
$q \in \mathbb{R}^n$, momenta $p \in \mathbb{R}^n$, and time $t \in \mathbb{R}$. We restrict attention throughout to autonomous systems, $\partial H/\partial t=0$,
and take $H$ to be of class $C^2$. We write $z=(q,p)\in\mathbb{R}^{2n}$ for a point of phase space and $\phi(z_0,t)$
for the flow of the system, the state reached at time $t$ by the
trajectory passing through $z_0$ at time $0$. The flow is complete when every trajectory is defined for all $t\in\mathbb{R}$,
so that no solution escapes to infinity, or reaches a singularity of $H$, in
finite time. Because the system
is autonomous, the motion from a given state depends only on that state and on
the elapsed time, so a complete flow forms a one-parameter group,
$\phi(\phi(z_0,t),s)=\phi(z_0,t+s)$, and in particular evaluating it at $-t$
inverts it, $\phi(\phi(z_0,t),-t)=z_0$\citep[\S4]{arnold1989mathematical}. A change of
variables defined by $2n$ functions of the form
\begin{equation}
Q_i = Q_i(q, p, t), \quad P_i = P_i(q, p, t), \quad i \in \{1,\ldots,n\}
\label{eq:canonical_transf}
\end{equation}
such that Hamilton's equations hold in the new variables for any Hamiltonian is called a \emph{canonical transformation}. Such a map is differentiable and invertible on phase space $(q, p)$ at each fixed $t$, and its $2n \times 2n$ Jacobian matrix
$M = \partial(Q,P)/\partial(q,p)\big|_{t}$ satisfies the symplectic condition
$M^\top \Omega M = \Omega$, with
$\Omega = \bigl(\begin{smallmatrix} 0 & I_n \\ -I_n & 0 \end{smallmatrix}\bigr)$ \citep[\S\S9.1 and 9.4]{goldstein2002classical}. A particularly useful canonical transformation is one for which the transformed Hamiltonian vanishes, $K=0$. In that case Hamilton's equations become $\dot{Q}_i = 0$ and $\dot{P}_i = 0$, so that $P_i = \alpha_i$ and $Q_i = \beta_i$, where $(\alpha,\beta)$ are $2n$ conserved quantities determined by the initial conditions. Knowing these conserved quantities, the original coordinates are recovered at any time~$t$ via the inverse, also canonical, transformation:

\begin{equation}
q_i(t) = q_i(\alpha_j, \beta_j, t), \qquad
p_i(t) = p_i(\alpha_j, \beta_j, t),
\label{eq:hj_solution}
\end{equation}
yielding the complete equations of motion. This framework for handling dynamics is called the Hamilton--Jacobi method.

The canonical transformation is traditionally constructed via a generating
function\footnote{For concreteness, the example assumes a type-2 generating function
$S = F_2(q_i, P_i, t)$. In general the generating function can depend on $n$ old and $n$ transformed
phase space variables, one from each conjugate pair, and on time.} $S$, where the transformed Hamiltonian satisfies
$K = H + \partial S/\partial t$. Imposing $K = 0$ yields
$H\!\left(q_i,\,\frac{\partial S}{\partial q_i},\,t\right)
+ \frac{\partial S}{\partial t} = 0$,
which is the Hamilton--Jacobi equation, whose complete solution is subject to the
non-degeneracy requirement $\det\!\left(\frac{\partial^2 S}{\partial q_i \,\partial \alpha_j}\right)\neq 0$
\citep[\S10.1]{goldstein2002classical}\citep[\S47]{arnold1989mathematical}. The transformations themselves, however, admit a direct characterisation that does not require constructing a generating function. Write $\Psi(q,p,t)=(Q,P)$ for the transformation
\eqref{eq:canonical_transf} and $\Psi^{-1}(\alpha,\beta,t)=(q,p)$ for its inverse
\eqref{eq:hj_solution}, and let $X_H$ denote the Hamiltonian vector field, so that
a trajectory obeys $\dot z=X_H(z)$. Every canonical transformation with the $K=0$
property takes a common form.\footnote{Conservation of the transformed variables gives
$\partial K/\partial Q_i=\partial K/\partial P_i=0$, that is, on a connected domain, $K$ a function of
$t$ alone. Hamilton's equations involve $K$ only through its derivatives with respect to
$Q$ and $P$, so a $K$ depending on $t$ alone gives the same dynamics as $K=0$;
this is therefore the same class of transformations.} Writing $D_z\Psi$ for the Jacobian of $\Psi$ with
respect to the phase-space variables at fixed $t$, the requirement that
$\Psi(z(t),t)$ be constant along every trajectory is equivalent to
\begin{equation}
    \frac{\partial \Psi}{\partial t} + (D_z\Psi)\,X_H = 0,
    \label{eq:cons_condition}
\end{equation}
and, provided the flow is complete, equivalent in turn to
\begin{equation}
    \Psi(z,t) = \Psi_0\bigl(\phi(z,-t)\bigr),
    \qquad \Psi_0(\cdot) := \Psi(\cdot,0),
    \label{eq:flow_form}
\end{equation}
where $\Psi_0$, the transformation at $t=0$, is itself canonical. Conversely,
every canonical $\Psi_0$ yields through \eqref{eq:flow_form} a canonical
transformation whose variables are conserved, so such transformations exist
whenever the flow is complete. These statements hold on any open set invariant
under the flow, and in particular on all of phase space
(Appendix~\ref{app:prop_flow}).

Parameterising the direct characterisation has three
advantages: First, holding on all of phase space, \eqref{eq:flow_form}
characterises the transformation as a single globally defined map, which is precisely what a neural network parameterises. Second, it imposes no representation-dependent regularity condition. Each standard
generating-function type takes a different set of $n$ old and $n$ transformed variables
as independent and exists only where that choice is locally invertible, so a fixed type
may degenerate where the canonical transformation itself is regular and another type is
well defined. This is further discussed in Appendix~\ref{app:generator_regularities}. Parameterising $\Psi$ directly only requires properties of
the dynamics, namely that $H$ be autonomous and of class $C^2$ and that its
flow be complete. Third, \eqref{eq:flow_form} carries no integrability
requirement: the existence established above rests on these assumptions alone,
none of which concerns integrability. This does not conflict with the classical obstructions to integrability, which concern a different object: the Liouville--Arnold theorem requires $n$ functionally independent,
\emph{time-independent} first integrals in
involution\citep[\S49]{arnold1989mathematical}, and non-integrability is the
statement that no such family exists in the relevant regularity class, whereas
$\Psi$ is allowed to depend explicitly on $t$, so its components need not be first
integrals of the autonomous system.

\begin{figure}[!htbp]
\centering
\begin{tikzpicture}[
    % --- Global styles ---
    font=\small,
    >=Stealth,
    % --- Color palette (muted) ---
    encoder fill/.style={fill=blue!10},
    decoder fill/.style={fill=orange!10},
    latent fill/.style={fill=green!12},
    hjl fill/.style={fill=blue!18},
    io fill/.style={fill=gray!10},
    box line/.style={draw=gray!60, line width=0.5pt, rounded corners=2.5pt},
    % --- Arrow style ---
    data arrow/.style={->, thick, draw=gray!60!black},
]

% ================================================================
%  INPUT
% ================================================================
\node[box line, io fill, minimum height=0.7cm, minimum width=1.1cm,
      align=center, inner sep=4pt]
      (input) {$q,\, p,\, t$};

% ================================================================
%  ENCODER
% ================================================================
\node[box line, encoder fill, minimum height=1.2cm, minimum width=3.5cm,
      align=center, inner sep=5pt, right=0.7cm of input]
      (encoder) {};

% Encoder label at top
\node[anchor=north, font=\scriptsize\bfseries, inner sep=0pt]
      at ([yshift=-3pt]encoder.north)
      {Mapping Network $\mathcal{M}$};

% HJL blocks inside encoder — centered via dots at midpoint
\node[font=\scriptsize, inner sep=0pt]
      at ([yshift=-3pt]encoder.center)
      (dots) {$\cdots$};

\node[box line, hjl fill, minimum height=0.45cm, minimum width=0.65cm,
      font=\scriptsize, inner sep=2pt, left=0.08cm of dots]
      (hjl1) {$\mathrm{HJL}_L$};

\node[box line, hjl fill, minimum height=0.45cm, minimum width=0.65cm,
      font=\scriptsize, inner sep=2pt, right=0.08cm of dots]
      (hjlL) {$\mathrm{HJL}_1$};
% Small arrows inside encoder
\draw[->, thin, draw=gray!50] (hjl1) -- (dots);
\draw[->, thin, draw=gray!50] (dots) -- (hjlL);

% ================================================================
%  LATENT
% ================================================================
\node[box line, latent fill, minimum height=1.2cm, minimum width=1.4cm,
      align=center, inner sep=4pt, right=0.7cm of encoder]
      (latent) {$Q,\, P$\\[-1pt]
                {\scriptsize Latents}};

% ================================================================
%  DECODER
% ================================================================
\node[box line, decoder fill, minimum height=1.2cm, minimum width=3.5cm,
      align=center, inner sep=5pt, right=0.7cm of latent]
      (decoder) {};

\node[anchor=north, font=\scriptsize\bfseries, inner sep=0pt]
      at ([yshift=-3pt]decoder.north)
      {Inverse Network $\mathcal{M}^{-1}$};

% IHJL blocks inside decoder — mirroring encoder
\node[font=\scriptsize, inner sep=0pt]
      at ([yshift=-3pt]decoder.center)
      (idots) {$\cdots$};

\node[box line, decoder fill, minimum height=0.45cm, minimum width=0.65cm,
      font=\scriptsize, inner sep=2pt, left=0.08cm of idots]
      (ihjl1) {$\mathrm{IHJL}_1$};

\node[box line, decoder fill, minimum height=0.45cm, minimum width=0.65cm,
      font=\scriptsize, inner sep=2pt, right=0.08cm of idots]
      (ihjlL) {$\mathrm{IHJL}_L$};

% Arrows inside decoder
\draw[->, thin, draw=gray!50] (ihjl1) -- (idots);
\draw[->, thin, draw=gray!50] (idots) -- (ihjlL);

% ================================================================
%  OUTPUT
% ================================================================
\node[box line, io fill, minimum height=0.7cm, minimum width=1.1cm,
      align=center, inner sep=4pt, right=0.7cm of decoder]
      (output) {$\hat{q},\, \hat{p}$};

% ================================================================
%  t_pred INPUT
% ================================================================
\node[box line, io fill, minimum height=0.55cm, minimum width=0.9cm,
      align=center, inner sep=3pt, font=\small,
      below=0.45cm of decoder]
      (tpred) {$t_{\mathrm{pred}}$};

% ================================================================
%  ARROWS
% ================================================================
\draw[data arrow] (input)   -- (encoder);
\draw[data arrow] (encoder) -- (latent);
\draw[data arrow] (latent)  -- (decoder);
\draw[data arrow] (decoder) -- (output);
\draw[data arrow] (tpred)   -- (decoder);

% ================================================================
%  SHARED WEIGHTS ANNOTATION
% ================================================================
\coordinate (sw left)  at ([yshift=0.4cm]encoder.north);
\coordinate (sw right) at ([yshift=0.4cm]decoder.north);

\draw[dashed, draw=gray!70!black, line width=0.6pt]
      (encoder.north) -- (sw left);
\draw[dashed, draw=gray!70!black, line width=0.6pt]
      (sw left) -- node[above, font=\scriptsize, text=gray!60!black,
      yshift=1pt] {shared weights} (sw right);
\draw[->, dashed, draw=gray!70!black, line width=0.6pt]
      (sw right) -- (decoder.north);
\draw[->, dashed, draw=gray!70!black, line width=0.6pt]
      (sw left) -- (encoder.north);

\end{tikzpicture}

\caption{Overview of the Hamilton--Jacobi Network. A phase-space
observation $(q, p, t)$ is encoded by the Mapping
Network~$\mathcal{M}$, a composition of $L$ Hamilton--Jacobi Layers,
into latents~$(Q, P)$, which are conserved quantities of the dynamics
along a segment exactly when the prediction loss vanishes on it
(Section~\ref{sec:convergence}). The Inverse Network~$\mathcal{M}^{-1}$ decodes them at all requested
query times~$t_{\mathrm{pred}}$ to predict
$(\hat{q}, \hat{p})$, in a single parallel forward pass. The decoder
is the exact algebraic inverse of the encoder and shares all
parameters; both maps are symplectic by construction.}

\label{fig:overview}
\end{figure}
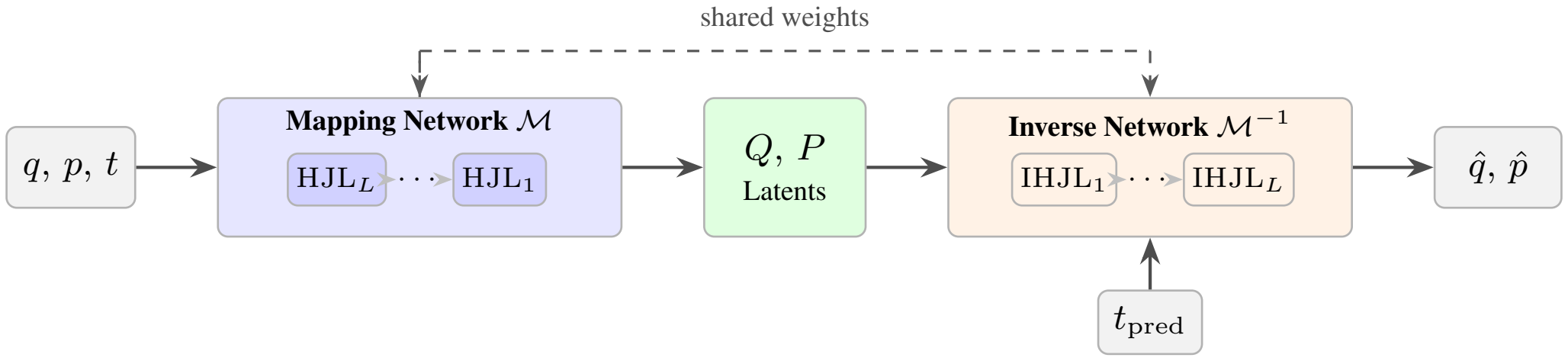

Our method learns the canonical transformation \eqref{eq:canonical_transf} and
its inverse \eqref{eq:hj_solution} from data via a symplectically constrained
encoder--decoder architecture (Figure~\ref{fig:overview}), the encoder
realising $\Psi$ and the decoder $\Psi^{-1}$. The key observation is that
Hamilton--Jacobi theory recasts learning dynamics as a \emph{representation
learning} problem: because the conserved quantities $(\alpha,\beta)$ fully encode
the phase-space trajectory, discovering the canonical transformation that
produces them and predicting the trajectory are two sides of the same problem.

\subsection{Hamilton--Jacobi Network}

\subsubsection{Overview}
\label{sec:hjn_overview}
We now translate the Hamilton--Jacobi construction into an architecture (Figure~\ref{fig:overview}), which we call the Hamilton--Jacobi Network (HJN). The model comprises two networks. A Mapping Network (encoder) $\mathcal{M}$ implements a time-dependent canonical transformation on phase space to latents:
\begin{equation}
    Q,\; P = \mathcal{M}(q,\; p,\; t),
    \qquad q, p, Q, P \in \mathbb{R}^n,\; t \in \mathbb{R}.
    \label{eq:mapping}
\end{equation}
An \textbf{Inverse Network} (decoder) $\mathcal{M}^{-1}$, the analytical inverse
of~$\mathcal{M}$, using these latents, predicts the original coordinates at arbitrary query time
$t_{\mathrm{pred}} \in \mathbb{R}$:
\begin{equation}
    \hat{q},\; \hat{p} = \mathcal{M}^{-1}(Q,\; P,\; t_{\mathrm{pred}}),
    \qquad \hat{q}, \hat{p} \in \mathbb{R}^n.
    \label{eq:inverse}
\end{equation}
Each prediction is computed from the fixed latents and its own
$t_{\mathrm{pred}}$ alone, so the decoder can be evaluated at all
requested query times from a single encoding, in one forward pass. Canonicity and exact invertibility are what \eqref{eq:flow_form}
requires of $\Psi$, and both are enforced as hard architectural
constraints: the Mapping Network is composed of layers that are symplectic by construction for every admissible parameter value and, since each sub-step has block-triangular structure, globally invertible with a closed-form algebraic
inverse (Appendix~\ref{app:invertible}). The decoder then shares all parameters
with the encoder and guarantees exact reconstruction
$\mathcal{M}^{-1}(\mathcal{M}(q, p, t)) = (q, p)$ up to numerical
precision when $t_{\mathrm{pred}} = t$. The resulting architecture is a tied-weight autoencoder over a $2n$-dimensional
latent space with the encoder--decoder pair defining a volume-preserving
($\det M = 1$; Appendix~\ref{app:invertible}), bijection at every $t$. 

\subsubsection{Composition}
\label{sec:hjl}
The Mapping Network is a composition $\mathcal{M} = \mathrm{HJL}_L \circ \cdots \circ \mathrm{HJL}_1$ of $L$ Hamilton--Jacobi Layers (HJLs). Each HJL composes symplectic shear maps parameterised by neural
networks, following~\cite{Jin2020,Horn2025}.
As in~\cite{Canizares2024}, time enters as a continuous input. We further augment each sub-step with learnable modulation parameters:
$a,a' \neq 0$ are scalar scaling factors applied to every conjugate pair,
$\gamma,\gamma'$ are scalars controlling the strength of the
neural shear terms, and $c_{1,i},c_{2,i},c'_{1,i},c'_{2,i}$ are per-coordinate
shifts. Fixing $a=a'=1$, $\gamma=\gamma'=1$, and all shifts to zero recovers the
un-modulated shear layer exactly. An HJL consists of two sub-steps applied sequentially, with the outputs of Step~1
serving as inputs to Step~2.
Each sub-step acts on every degree of freedom $i \in \{1,\ldots,n\}$:

\textbf{Step~1} (coordinate scaling, momentum shift):
\begin{align}
    Q_i^{(1)} &= a \cdot q_i + c_{1,i}, \label{eq:hjl_s1q}\\
    P_i^{(1)} &= \tfrac{1}{a}\, p_i
               - \gamma\,\tfrac{\partial G(\mathbf{q},\, t)}{\partial q_i}
               + c_{2,i}, \label{eq:hjl_s1p}
\end{align}

\textbf{Step~2} (momentum scaling, coordinate shift):
\begin{align}
    P_i^{(2)} &= a' \cdot P_i^{(1)} + c'_{1,i}, \label{eq:hjl_s2p}\\
    Q_i^{(2)} &= \tfrac{1}{a'}\, Q_i^{(1)}
               + \gamma'\,\tfrac{\partial F(\mathbf{P}^{(1)},\, t)}
                                {\partial P_i^{(1)}}
               + c'_{2,i}, \label{eq:hjl_s2q}
\end{align}

where $G(\mathbf{q}, t)$ and $F(\mathbf{P}^{(1)}\!, t)$ are scalar-valued MLPs,
each taking all $n$ components of their respective coordinate vector together with
time as input. We parameterise $G$ and $F$ as arbitrarily deep MLPs and compute the required
partial derivatives by automatic differentiation. Our motivation for using depth is parameter efficiency: there exist functions
that a deep network approximates with far fewer parameters than any shallow
network requires~\cite{Telgarsky2015RepresentationBO}. Empirical results with symplectic architectures suggest this trade-off can be relevant in practice~\cite{Horn2025}. Neither this choice nor the modulation
parameters are claimed to improve accuracy. The canonicality proof (Appendix~\ref{sec:sm_proof}) requires $C^2$ activation functions (e.g. GELU) and that $a,a' \neq 0$, which we enforce (Appendix~\ref{app:init_details}). Since each layer is canonical, so is $\mathcal{M}$ at arbitrary depth (Appendix~\ref{app:composition}). The Inverse Network reverses each layer in closed algebraic form
(details in Appendix~\ref{app:invertible}), yielding the decoder
$\mathcal{M}^{-1} = \mathrm{IHJL}_1 \circ \cdots \circ \mathrm{IHJL}_L$,
which shares all parameters with the encoder and is itself canonical
(Appendix~\ref{app:inverse}). Because these guarantees hold only under exact arithmetic, we verify numerically that both trained networks remain symplectic to within single-precision error (Appendix~\ref{app:symplecticity_diagnostic}).

\subsection{Training}
\label{sec:training}

\paragraph{Temporal segmentation and time localisation.}
The training set consists of trajectories of observations $(q,p,t)$. We partition each trajectory into contiguous segments of duration at most $T_{\max}$. Segmentation is not required by the theoretical construction: by Section~\ref{subsec:problem_formulation} the conserved quantities $(\alpha,\beta)$ parameterise the trajectory for all time, so a single segment per trajectory would suffice. The reason is one of optimisation: fitting a model to reconstruct a window of a trajectory tends to become harder as that window lengthens. \citet{Ribeiro2019OnTS} quantify
this for prediction-error estimation. Their objective maps the model parameters,
together with the initial condition the window is simulated from, to the
reconstruction error they produce; For a non-contractive model the Lipschitz and $\beta$-smoothness constants admit bounds that grow polynomially or exponentially in the window length. They observe that their resulting objectives
are dense with closely spaced local minima that trap their solver near its initial
guess. They further show the bounds are governed by the longest window rather than
by the record, so splitting a record into shorter windows is the remedy they adopt. The same remedy is standard practice for training neural differential equations with the addition of continuity across segments being imposed by a constraint or a prior~\citep{Turan2021MultipleSF,Iakovlev2022LatentNO}. What remains is the choice of $T_{\max}$.
\citet{chakraborty2024divide} argue for basing it on the fastest Lyapunov
time scale of the system:
the sensitivity of the reconstructed trajectory to the
quantities being fitted obeys a linear equation whose homogeneous part is the
linearised flow, so it is amplified at the rate the Lyapunov exponent gives, and keeping each window within that timescale is what prevents the exploding gradients and the poor loss landscape they report. We set $T_{\max}$ on the same reasoning: the minimum local Lyapunov time over the training observations, or the duration of the longest training trajectory when that is shorter. Both analyses concern models that differ from ours (a one-step map composed over the window; vector-field parameters); Appendix~\ref{sec:sampling_limitation} carries them to our setting. $T_{\max}$ is fixed from the training data before optimisation begins, not by trialling values against the evaluation. Appendix~\ref{sec:sampling_limitation} sets out the rule, and reports its application to each task; Appendix~\ref{app:lyap_estimator} specifies the estimator. We claim neither the rule nor the estimator as optimal, and
Appendix~\ref{sec:tmax_sweep} reports its sensitivity.

Moreover, within a segment starting at $t_{\mathrm{start}}$ we replace $t$ by
$t' \;=\; t - t_{\mathrm{start}} \;\in\; [0,T_{\max}]$, so that the network is
trained on segment-local elapsed time rather than on absolute time along the
original trajectory. The same
parameterisation is used at inference (Section~\ref{sec:autoregressive}), where
the target prediction interval is partitioned into windows of duration at most
$T_{\max}$. Consequently, the network is never evaluated at a time input outside
$[0,T_{\max}]$, the range covered in training, addressing extrapolation in the time argument as a failure mode. The shift is also consistent with the autonomy assumption in
Section~\ref{subsec:problem_formulation}: the motion from
a given state depends only on that state and the elapsed time, so removing the
absolute time origin does not discard information relevant to prediction.

\paragraph{Prediction loss.}
Each batch draws one observation from each of $N$ segments.
The encoder produces latents
$Q, P \in \mathbb{R}^{N \times n}$ via~\eqref{eq:mapping}.
For each segment, $N_{\mathrm{pred}}$ query times are sampled, forming
$t_{\mathrm{pred}} \in \mathbb{R}^{N \times N_{\mathrm{pred}}}$.
The latents are broadcast along the query-time dimension and the
Inverse Network produces predictions at all query times in parallel:
\begin{equation}
    \hat{q},\; \hat{p}
    = \mathcal{M}^{-1}(Q_{\mathrm{full}},\; P_{\mathrm{full}},\; t_{\mathrm{pred}}),
    \qquad \hat{q}, \hat{p} \in \mathbb{R}^{N \times N_{\mathrm{pred}} \times n}.
    \label{eq:decode_batch}
\end{equation}

The prediction loss is the MAE over all valid (segment, query-time,
degree-of-freedom) entries. Writing $M_{jm}\in\{0,1\}$ for a mask that is
$1$ when query time $m$ is a real sample of segment $j$ and $0$ for
padding,
\[
\mathcal{L}_{\mathrm{pred}}
= \frac{\sum_{j,m,i} M_{jm}\,
  \bigl(|\hat{q}_{jmi} - q_{jmi}| + |\hat{p}_{jmi} - p_{jmi}|\bigr)}
  {n\sum_{j,m} M_{jm}}.
\]
When all segments share a common length $N_{\mathrm{pred}}$ this reduces to
$\tfrac{1}{N\,N_{\mathrm{pred}}\,n}\sum_{j,m,i}(\cdot)$. We use MAE rather than MSE because it is less sensitive to coordinate
magnitudes, yielding more balanced gradient contributions across the
energy range.

\paragraph{Prediction and discovery as one objective.}
\label{sec:convergence}
With canonicity and exact invertibility supplied by the architecture, two statements make precise the sense in which prediction and discovery of conserved quantities are one objective, as anticipated in Section~\ref{subsec:problem_formulation}. Times within a segment are segment-local, so a segment consists of observations $\bigl(\phi(z_0,t),t\bigr)$ at a finite set of times $t\in\mathcal{T}\subset[0,T_{\max}]$, where $z_0$ is the state the segment occupies at local time~$0$ and $0\in\mathcal{T}$ because the segment's first observation defines its local time origin.

For the first statement, what has to be shown is that the prediction loss selects, among canonical maps, those of the form~\eqref{eq:flow_form} on the training data. A batch encodes one observation of the segment, drawn at some anchor time, and decodes it at the $N_{\mathrm{pred}}$ query times drawn for that segment, where the target at query time $t_{\mathrm{pred}}$ is $\phi(z_0,t_{\mathrm{pred}})$. We characterise the maps at which the sampling-averaged objective $\mathbb{E}\,\mathcal{L}_{\mathrm{pred}}$, the expectation of $\mathcal{L}_{\mathrm{pred}}$ over the draw of the batch (segments, anchor and query times), vanishes. $\mathcal{L}_{\mathrm{pred}}$ is a sum of non-negative terms over a positive normaliser and the draws are from finite sets, so the expectation vanishes exactly when every term arising with positive probability does; terms belonging to different segments enter the sum without coupling, so, every segment being drawn with positive probability, it suffices to fix one segment. Provided the sampling covers each segment, in that every ordered pair of its observation times arises as an (anchor, query) pair with positive probability, vanishing of the objective means that encoding any one observation of the segment and decoding at any of its times returns the observed state:
$$
\mathcal{M}^{-1}\bigl(\mathcal{M}(\phi(z_0,t),\,t),\;t_{\mathrm{pred}}\bigr)= \phi(z_0,\,t_{\mathrm{pred}}),\qquad t,\,t_{\mathrm{pred}}\in\mathcal{T}.
$$
The right-hand side does not depend on $t$, so observations taken at different times of the segment are decoded to the same state at a common query time; and since $\mathcal{M}^{-1}(\cdot,t_{\mathrm{pred}})$ is the inverse of the bijection $\mathcal{M}(\cdot,t_{\mathrm{pred}})$ and hence injective (Appendix~\ref{app:invertible}), the latents they are decoded from must themselves agree, $\mathcal{M}(\phi(z_0,t),t)=\mathcal{M}(\phi(z_0,t'),t')$ for all $t,t'\in\mathcal{T}$; injectivity at a single shared query time already forces this. The encoder output is thus constant along the segment; evaluating at $t=0$, which $0\in\mathcal{T}$ permits, identifies its value as $\Psi_0(z_0)$, where $\Psi_0:=\mathcal{M}(\cdot,0)$ is canonical by construction. An observation of the segment is $z=\phi(z_0,t)$, so $z_0=\phi(z,-t)$ by the group property, and the constancy therefore reads $\mathcal{M}(z,t)=\Psi_0\bigl(\phi(z,-t)\bigr)$ at every observed pair $(z,t)$, which is the restriction of \eqref{eq:flow_form}, condition~(iii) of
Appendix~\ref{app:prop_flow}, to the states and times the training data covers. Every observation of the segment then encodes to $\Psi_0(z_0)$, the $2n$ conserved quantities that parameterise the segment through \eqref{eq:hj_solution}. Symplecticity is what makes those constants a canonical set, so that $\mathcal{M}(\cdot,t)$ is a canonical transformation with $K=0$, the Hamilton--Jacobi transformation of Section~\ref{subsec:problem_formulation}, and not merely an invertible change of variables to constants; it also confines the predicted map $\mathcal{M}^{-1}(\cdot,t')\circ\mathcal{M}(\cdot,t)$ to canonical maps, as every Hamiltonian flow is. The converse also holds. If $\mathcal{M}$ has the form \eqref{eq:flow_form} at the states and times the training data covers, then every observation of the segment, anchor and target alike, encodes to $\Psi_0(z_0)$, since $\phi(\phi(z_0,s),-s)=z_0$ for every $s$. Decoding the anchor's latents at $t_{\mathrm{pred}}$ therefore returns $\phi(z_0,t_{\mathrm{pred}})$, the target, by exact invertibility of $\mathcal{M}(\cdot,t_{\mathrm{pred}})$ (Appendix~\ref{app:invertible}). Every term of the objective vanishes; and since $z\mapsto\Psi_0(\phi(z,-t))$ is canonical for canonical $\Psi_0$ (Appendix~\ref{app:prop_flow}), the symplectic constraint excludes no map of the form \eqref{eq:flow_form} with canonical $\Psi_0$. Vanishing of the objective constrains the latents only within a segment. Terms of different segments do not interact, and the latents a segment receives are $\Psi_0(z_0)$, fixed by that segment's own state at local time~$0$; since $\Psi_0$ is injective, segments with distinct initial states receive distinct latents, consecutive segments of a single trajectory included. In conclusion, the first statement is that on the states and times the training data covers, zero prediction loss and \eqref{eq:flow_form} are the same condition on $\mathcal{M}$: the transformations to conserved quantities are precisely the zero set of the prediction objective.

The condition above concerns exact vanishing. The second statement is that
constancy of the latents and prediction error remain tied at nonzero loss as
well. Nothing in the following argument refers to the training data or to the value of the loss, only to exact invertibility and the smoothness of the activations, so it holds on any segment of any trajectory. Fix a
segment with initial state $z_0$ and two of its observation times $t,t'$, and
write $\ell_t:=\mathcal{M}(\phi(z_0,t),t)$ and
$\ell_{t'}:=\mathcal{M}(\phi(z_0,t'),t')$ for the latents of the two
observations, so that $\lVert\ell_t-\ell_{t'}\rVert$ is their departure from
constancy. Decoding $\ell_t$ at $t'$ gives the model's prediction
$\hat z:=\mathcal{M}^{-1}(\ell_t,t')$ of the state $\phi(z_0,t')$, and
$\lVert\hat z-\phi(z_0,t')\rVert$ is the prediction error. Because
$\mathcal{M}(\cdot,t')$ is the exact inverse of $\mathcal{M}^{-1}(\cdot,t')$,
encoding the prediction at $t'$ returns $\ell_t$, while encoding the true state
at $t'$ returns $\ell_{t'}$; hence
\begin{equation}
\ell_t-\ell_{t'}
=\mathcal{M}(\hat z,t')-\mathcal{M}\bigl(\phi(z_0,t'),t'\bigr):
\label{eq:constancy_error_identity}
\end{equation}
the departure from constancy is the image under $\mathcal{M}(\cdot,t')$ of the
pair (prediction, target). The two quantities are distances: the departure from constancy is the distance
between the values of $\mathcal{M}(\cdot,t')$ at two points, and the prediction
error is the distance between those points. The mean value inequality relates
exactly these. For a map $f$ differentiable on a convex open set $B$ containing $a$ and $b$, with $\lVert Df(x)\rVert\le L$ in operator norm for all $x\in B$,
\[
\lVert f(a)-f(b)\rVert \;\le\; L\,\lVert a-b\rVert
\]
\citep[Thm.~9.19]{rudin1976principles}. Take $f=\mathcal{M}(\cdot,t')$, which is
continuously differentiable because the activations are $C^2$
(Section~\ref{sec:hjl}), $a=\hat z$, $b=\phi(z_0,t')$, $B$ an open ball
containing both, and $L$ the maximum of $\lVert D_z\mathcal{M}(\cdot,t')\rVert$
over its closure, finite because the Jacobian is continuous and the closure
compact. Then \eqref{eq:constancy_error_identity} gives
\begin{equation}
\lVert \ell_t-\ell_{t'}\rVert \;\le\; L\,\lVert \hat z-\phi(z_0,t')\rVert .
\label{eq:constancy_le_error}
\end{equation}
Applied instead to $f=\mathcal{M}^{-1}(\cdot,t')$, also continuously
differentiable, at $a=\ell_t$ and $b=\ell_{t'}$, whose images are $\hat z$ and
$\phi(z_0,t')$, the same inequality gives, with $L'$ the corresponding bound
on $\lVert D\mathcal{M}^{-1}(\cdot,t')\rVert$ over the closure of an open ball containing $\ell_t$ and $\ell_{t'}$,
\begin{equation}
\lVert \hat z-\phi(z_0,t')\rVert \;\le\; L'\,\lVert \ell_t-\ell_{t'}\rVert .
\label{eq:error_le_constancy}
\end{equation}
The departure of the latents from constancy and the prediction error therefore bound each other, the first by $L$ times the second and the second by $L'$ times the first. In conclusion, the second statement carries this tie off the zero set and in both directions: up to the local bounds $L$ and $L'$, the encoder $\mathcal{M}$ holds the latents it produces constant along a segment to the degree that the decoder $\mathcal{M}^{-1}(\cdot,t_{\mathrm{pred}})$ predicts that segment from them, and the decoder predicts that segment to the degree that those latents are constant, at whatever loss the model has reached and on whatever segment it is applied to.

\paragraph{Independence loss.}
The equivalence above fixes the conserved information the latents must carry, but
not the $\Psi_0$ of \eqref{eq:flow_form}. Composing
encoder and decoder gives $\Psi^{-1}(\Psi(z,t),t')=\phi(z,t'-t)$, in which $\Psi_0$
cancels (Appendix~\ref{app:prop_flow}). This invariance is not confined to the
optimum: at every parameter setting, encoders differing by a frame, that is, a
$t$-independent canonical map $S$ with $\mathcal{M}'=S\circ\mathcal{M}$, receive the same prediction loss in exact
arithmetic (Appendix~\ref{app:frame_invariance}). The frame is therefore
selected not by $\mathcal{L}_{\mathrm{pred}}$ but by the initialisation, the
optimiser and the descent path, none of which accounts for the quality of the
resulting map. One such quality is componentwise conditioning. Componentwise ill-conditioning appears
as dependence within a conjugate pair: when $Q^{(i)}$ and $P^{(i)}$ are nearly
functionally related, recovering one from the stored pair requires a cancellation
between two nearly equal quantities. Rounding errors in the finite-precision latents can be amplified when the decoder inverts a componentwise ill-conditioned map, and the amplified errors can accumulate over the repeated re-encodings performed during inference (Section~\ref{sec:autoregressive}). For instance, the frame
sweep of Appendix~\ref{app:hsic_frame_sweep} constructs a frame that leaves the predictor unchanged, yet multiplies the single-precision rollout error by $144$. Nothing guarantees that training avoids such a frame, so as a safeguard we penalise the statistical dependence between
the paired outputs $(Q^{(i)}, P^{(i)})$ of each degree of freedom $i$, measured by the
Hilbert--Schmidt Independence Criterion (HSIC)~\cite{Gretton2005MeasuringSD}. We use it with RBF kernels, so that the penalty is not restricted to linear dependence, the transformations~\eqref{eq:canonical_transf} being generically nonlinear. The independence loss is
$\mathcal{L}_{\mathrm{HSIC}} = \frac{1}{n}\sum_{i=1}^{n}
\mathrm{HSIC}(Q^{(i)}, P^{(i)})$. In Appendix~\ref{app:hsic_frame_sweep} we
motivate the term by exhibiting the failure mode it is designed to prevent and by showing that the penalty is non-zero on every frame in that sweep exhibiting it.

\paragraph{Total loss.}
Because the prediction and independence objectives have different scales and
units, both terms are normalised by data-driven baselines
($\bar{\mathcal{L}}_{\mathrm{pred}}$ from a calibration epoch,
$H_{\max}$ from a maximally correlated reference) so that their gradient
contributions start balanced:
\begin{equation}
    \mathcal{L}_{\mathrm{total}}
    = \lambda_{\mathrm{pred}}
      \frac{\mathcal{L}_{\mathrm{pred}}}
           {\bar{\mathcal{L}}_{\mathrm{pred}}}
    + \lambda_{\mathrm{HSIC}}
      \frac{\mathcal{L}_{\mathrm{HSIC}}}
           {H_{\max}},
    \label{eq:loss_total}
\end{equation}
where $\lambda_{\mathrm{pred}}$ and $\lambda_{\mathrm{HSIC}}$ are fixed
weighting coefficients. The complete training algorithm, including the calibration procedure
for $\bar{\mathcal{L}}_{\mathrm{pred}}$ and $H_{\max}$, is provided in Algorithm~\ref{alg:training} (Appendix~\ref{app:training_algorithm}).

\subsection{Inference}
\label{sec:autoregressive}
Since training covers only segment-local times in $[0,T_{\max}]$ (Section~\ref{sec:training}), the target timeline is partitioned into segments of duration at most $T_{\max}$. Within each segment, $\mathcal{M}$ encodes a single observation into its latents, which $\mathcal{M}^{-1}$ decodes at all query times in that segment in parallel, in a single forward pass. At each segment boundary, the predicted boundary state $(\hat{q}, \hat{p})$ is re-encoded by $\mathcal{M}$, with the segment-local time reset to $0$ when predicting forward in time and to $T_{\max}$ when predicting backward in time. The complete inference algorithm and a cost analysis are provided in Appendix~\ref{app:handoff}.

\section{Experiments}
\label{sec:experiments}

\subsection{Tasks and training}
\label{sec:tasks}

We evaluate HJN on four systems: the analytically integrable ideal mass--spring oscillator, the numerically integrable nonlinear ideal pendulum, a real-world non-Hamiltonian dissipative system (a physical pendulum subject to air resistance) that lies outside the assumptions of our theory, and the non-integrable H\'enon--Heiles system, which exhibits chaotic trajectories. The first three tasks follow the experimental setup of~\citet{Greydanus2019}: we
adopt their systems, physical constants, training energy ranges, and
observation noise level. For Task~3 we use the same measured pendulum
trajectory of~\citet{schmidt2009} under their train/test split. In every task, each batch step decodes every sample of every segment.

\paragraph{Task~1 -- Ideal mass--spring.}
$H=\tfrac{1}{2}(p^2+q^2)$ with $k=m=1$,
yielding a period $T\approx 6.283$. Trajectories are sampled
from the analytic solution. Training uses 24 trajectories at
$E\in[0.2,\,1]$ that span $t=6.493$ (32 samples each), each treated as a single training segment ($T_{\max}=6.493$, $N_{\mathrm{pred}}=32$). Two test sets probe rollout extrapolation (Test~A: horizon $t{\sim}3.1\,T_{\max}$ at the training energies) and energy extrapolation (Test~B: 25 trajectories at $E\in[1.118,\,1.942]$, with horizon $t=T_{\max}$).

\paragraph{Task~2 -- Ideal pendulum.}
$H=\tfrac{1}{2}p^2+3(1-\cos q)$ with $m=l=1$ and $g=3$, restricted to
the oscillatory regime. Ground-truth trajectories are generated with an adaptive high-order integrator (DOP853; tolerances in Appendix~\ref{app:task_specs}). Training uses 25
trajectories at $E\in[1.3,\,2.3]$ with horizon $4.06$ (30 samples
each, full trajectory as a single segment: $T_{\max}=4.06$,
$N_{\mathrm{pred}}=30$). Test~A: horizon $t{\sim}4.9\,T_{\max}$, drawn from the same
energy distribution as training. Test~B: $E\in[2.4,\,4.43]$, horizon $t{\sim}1.2\,T_{\max}$.

\paragraph{Task~3 -- Real pendulum~\cite{schmidt2009,schmidt2009data}.}
A single measured trajectory from a physical pendulum (556
mostly uniformly spaced samples over ${\sim}94.5$\,s).
Following~\cite{Greydanus2019} we adopt $E=p^2+2.4(1-\cos q)$ as the
nominal energy. Training uses the first 444 samples. The
estimator of Section~\ref{sec:tmax_selection} returns a $T_{\max}=15.973$, below
$T_{\mathrm{record}}$, so the record is divided into
$\lceil 75.833/15.973\rceil=5$ contiguous segments, whose sample counts differ. The final 112 samples, the last $18.5$\,s of the record, form the test set.

\paragraph{Task~4 -- H\'enon--Heiles.}
$H=\tfrac{1}{2}(p_x^2+p_y^2)+\tfrac{1}{2}(x^2+y^2)
+(x^2y-\tfrac{1}{3}y^3\bigr)$. Ground-truth trajectories are integrated with
an adaptive high-order Taylor scheme (heyoka~\cite{heyoka_sw,biscani2021heyoka}; details in Appendix~\ref{app:task_specs}). Training uses 48
trajectories over $E\in[0.08,\,0.125]$, each spanning $t=50$ ($334$ samples). The estimator of
Section~\ref{sec:tmax_selection} returns $T_{\max}=8.563$, so each training trajectory is divided into $\lceil 50/8.563\rceil=6$ contiguous segments. Test~A: horizon $t{\sim}5.8\,T_{\max}$ at the training energies. Test~B: $E\in[0.13,\,\tfrac{1}{6})$, up to the escape energy, at horizon $t{\sim}2\,T_{\max}$.

\label{sec:implementation_details}
Tests~A and~B of Tasks~1, 2 and~4 form the six extrapolation settings, all energy-conserving and within the scope of the method, while the test set of Task~3 is neither. For Tasks~1 and~2, zero-mean Gaussian noise with variance $\sigma^2 = 0.1$ is added to each observation during training. For Task~4, the noise variance is reduced to $\sigma^2 = 0.0045$ to
match the signal-to-noise ratio of Tasks~1--2, since the H\'enon--Heiles
coordinates have substantially smaller variance (Appendix~\ref{app:training_details}). Task~3 adds no noise, since the data already contain real measurement noise.

\paragraph{Architecture and optimisation.}
An identical HJN backbone is used across all tasks: $L=10$ Hamilton--Jacobi Layers (Section~\ref{sec:hjl}) with MLP potentials $F,G$ of hidden widths $[33,60,32]$ and GELU activations, except those of the final HJL, which use a linear output so the transformation can realise unbounded coordinate scalings. Training uses AdamW with weight decay applied to MLP weights only. We repeat training with three random seeds (42, 52, 62); all hyperparameters, including the epoch budgets, were selected on seed~42 and held fixed for the other two. Architecture details, initialisation, and per-task hyperparameters are provided in Appendix~\ref{app:experimental_details} (Tables~\ref{tab:hparams_task}--\ref{tab:hparams_train}).

\subsection{Evaluation}
\label{sec:eval_protocol}

The equivalence of Section~\ref{sec:convergence} identifies the zero set of $\mathcal{L}_{\mathrm{pred}}$
with the maps of the form~\eqref{eq:flow_form}, whose latents are the conserved
quantities on the states and times the training data covers. Training, however, reaches a nonzero prediction loss and adds noise to the anchors. To what degree the trained models carry the structure that zero prediction loss forces exactly on the training data, both there and on states not seen during training, is what the first two experiments measure, by complementary means.

That equivalence concerns the prediction objective alone, whereas training also
optimises $\mathcal{L}_{\mathrm{HSIC}}$, which acts directly on the latents; any
property we observe in them could in principle be its product rather than the
prediction objective's. To attribute correctly, we retrain with
$\lambda_{\mathrm{HSIC}}=0$, using the same architecture, data and
checkpoint-selection criterion and repeat the experiments
(Appendix~\ref{zero_hsic_experiments}).

\paragraph{Constancy of the learned latents.}
\label{sec:latent_conservation}
At zero prediction loss, on the states and times the training data covers,
$\mathcal{M}$ maps every observation of a segment to the same latent $\Psi_0(z_0)$
(Section~\ref{sec:convergence}). This experiment measures
whether, in the regime training reaches, the latents of a segment remain approximately constant for both training and test sets. Raw distances between latents cannot carry this measurement, because latent space offers no scale against which a deviation from constancy counts as small. Encoders differing by a $t$-independent canonical map $S$, $\tilde{\mathcal{M}}=S\circ\mathcal{M}$, make the same predictions (Appendix~\ref{app:frame_invariance}), yet $S$ can change raw latent distances arbitrarily. We therefore remove the frame by composing the encoder with its own decoder at $t=0$, itself a $t$-independent canonical map (Appendix~\ref{app:inverse}), which gives the gauge-fixed encoder $\mathcal{M}_{\mathrm{g}}:=\mathcal{M}^{-1}(\cdot,0)\circ\mathcal{M}$. For any $\tilde{\mathcal{M}}=S\circ\mathcal{M}$, whose own decoder is $\tilde{\mathcal{M}}^{-1}=\mathcal{M}^{-1}\circ S^{-1}$,
\begin{equation*}
    \tilde{\mathcal{M}}^{-1}(\cdot,0)\circ\tilde{\mathcal{M}}
    = \mathcal{M}^{-1}(\cdot,0)\circ S^{-1}\circ S\circ\mathcal{M}
    = \mathcal{M}^{-1}(\cdot,0)\circ\mathcal{M},
\end{equation*}
so $\mathcal{M}_{\mathrm{g}}$ is the same for every $S$. For a segment $s$ with observations $z_{s,t}$ at segment-local times $t$, its gauge-fixed latents are
\begin{equation}
    g_{s,t} := \mathcal{M}^{-1}\bigl(\mathcal{M}(z_{s,t},\,t),\,0\bigr).
    \label{eq:gauge_latent}
\end{equation}
Consider a segment of any trajectory, training or held-out, on whose observations $\mathcal{M}$ has the form \eqref{eq:flow_form} with $\Psi_0=\mathcal{M}(\cdot,0)$; since $\mathcal{M}^{-1}(\cdot,0)=\Psi_0^{-1}$, the gauge-fixed encoder there is $\mathcal{M}_{\mathrm{g}}(z,t)=\Psi_0^{-1}\bigl(\Psi_0(\phi(z,-t))\bigr)=\phi(z,-t)$, the $\Psi_0=\mathrm{id}$ solution of Section~\ref{subsec:problem_formulation}. Applied to an observation $z_{s,t}=\phi(z_{s,0},t)$, the group property gives $g_{s,t}=z_{s,0}$: under $\mathcal{M}_{\mathrm{g}}$, a segment's conserved quantities are its initial state. The frame-free setting achieves two properties. First, since $\mathcal{M}^{-1}(\cdot,0)$ is a bijection, $g_{s,t}=g_{s,t'}$ exactly when $\mathcal{M}(z_{s,t},t)=\mathcal{M}(z_{s,t'},t')$, so constancy of $g$ is constancy of the raw latents. Second, $g_{s,t}$ is a point of phase space, so its deviations can be compared against scales of the system itself. Note that $g_{s,t}$ is the decoder's prediction of the segment's initial state from the latent of its observation at $t$, so constancy measurements using $g_{s,t}$ are not independent of prediction quality. Nor could any measure of constancy be: Section~\ref{sec:convergence} shows that zero prediction loss and constant latents are one condition on $\mathcal{M}$, and that at any loss the departure of the latents from constancy and the prediction error bound each other.

Each trajectory is partitioned as in training. The first observation of each segment defines its local time origin, so that $g_{s,0}=z_{s,0}$ up to the round-trip error $\lVert g_{s,0}-z_{s,0}\rVert$ of $\mathcal{M}$, measured at most $5.6\times10^{-6}$ in phase-space units over all tasks, seeds and sets. Being a round trip, this equality holds on every segment, whether or not $\mathcal{M}$ has the form~\eqref{eq:flow_form} there: to within that error, $g_{s,0}$ equals the segment's conserved quantities under $\mathcal{M}_{\mathrm{g}}$, the value every $g_{s,t}$ of the segment takes when $\mathcal{M}$ has that form. We then form, for every other observation,
\begin{equation}
    \delta_{s,t} := g_{s,t}-g_{s,0},
    \label{eq:deviation}
\end{equation}
which measures both the deviation of the gauge-fixed latents from constancy and, to within the round-trip error, their deviation from these frame-free conserved quantities. Every segment contains at least two observations, so the set $\mathcal{I}$ of a trajectory's eligible observations consists of all its observations except each segment's first. With Euclidean norms in raw $(q,p)$, we compare the deviation with two scales of phase space. The first is the segment's motion, through the displacement $\Delta_{s,t} := z_{s,t}-z_{s,0}$, taking:
\begin{equation}
    D_0 := \Biggl(\frac{\sum_{(s,t)\in\mathcal{I}}\lVert\delta_{s,t}\rVert^2}
                       {\sum_{(s,t)\in\mathcal{I}}\lVert\Delta_{s,t}\rVert^2}\Biggr)^{1/2}.
    \label{eq:D0}
\end{equation}
$D_0$ is the rms deviation of the gauge-fixed latents from their segments' origins,
relative to the rms displacement of the observed states from the same origins. It is $0$ exactly when the latents are constant across the observations of every segment, $g_{s,t}=g_{s,0}$ for every $(s,t)\in\mathcal{I}$. It is exactly $1$ for an encoder that ignores its time input, $\mathcal{M}(z,t)=\mathcal{M}(z,0)$, for which $g_{s,t}=z_{s,t}$ and hence $\delta_{s,t}=\Delta_{s,t}$. The sums are pooled over the whole trajectory rather than taken as per-observation ratios, which diverge near a return to the initial state. Pooling also weights each segment by its motion, so a short final segment cannot dominate. The second metric compares the same deviations from constancy with each task's training-noise scale $\sigma\sqrt{2n}$:
\begin{equation}
    \delta_{\mathrm{rms}} := \frac{1}{\sigma\sqrt{2n}}\Bigl(\frac{1}{|\mathcal{I}|}
        \sum_{(s,t)\in\mathcal{I}}\lVert\delta_{s,t}\rVert^2\Bigr)^{1/2}.
    \label{eq:delta_rms}
\end{equation}
Here $\sigma\sqrt{2n}$ is the rms length of one training-noise vector. The training noise $\varepsilon$ has $2n$ independent coordinates of variance $\sigma^2$, so $\mathbb{E}\lVert\varepsilon\rVert^2=2n\sigma^2$. $\sigma\sqrt{2n}$ is $0.447$ on
Tasks~1 and~2 and $0.134$ on Task~4. We use this scale because training presented the predictor with anchors perturbed by noise of this size and with clean targets, so $\delta_{\mathrm{rms}}<1$ means that a segment's observations agree on its initial state more closely than the perturbations the model was trained to absorb. The comparison is a reference point, not a resolution limit. Each trajectory yields one $D_0$ and one $\delta_{\mathrm{rms}}$; we report their median and interquartile range over trajectories, separately for each seed. Task~3 is excluded from this test: the measured pendulum is dissipative, so the method's premise does not supply quantities that are conserved along its trajectories. We evaluate three sets per task, all with clean observations: the training trajectories (Train), Test~A and Test~B. Train is where the zero set is characterised, so it measures how far the models are from that set and provides the in-distribution reference for every metric.

\begin{table}[htbp]
\centering
\caption{Within-segment constancy relative to segment motion. $D_0$ is the rms deviation of the gauge-fixed latents from that of their segment's $t=0$ observation, divided by the rms displacement of the observed states from it, pooled over each trajectory's eligible observations (all except each segment's first). Entries: median $[\,p_{25},p_{75}\,]$ over trajectories, per
trained model.}
\label{tab:constancy_D0}
\setlength{\tabcolsep}{4pt}
\begin{tabular}{llccc}
\toprule
Task & Seed & Train & Test~A & Test~B \\
\midrule
\multirow{3}{*}{Task~1} & 42 & $0.0235\,[0.0219,\,0.0307]$ & $0.0232\,[0.0214,\,0.0275]$ & $0.0358\,[0.0269,\,0.0573]$ \\
 & 52 & $0.0274\,[0.0218,\,0.0312]$ & $0.0262\,[0.0218,\,0.0318]$ & $0.0413\,[0.0341,\,0.0673]$ \\
 & 62 & $0.0266\,[0.0232,\,0.0363]$ & $0.0304\,[0.0248,\,0.0356]$ & $0.0485\,[0.0320,\,0.0700]$ \\
\midrule
\multirow{3}{*}{Task~2} & 42 & $0.0381\,[0.0272,\,0.0518]$ & $0.0379\,[0.0333,\,0.0491]$ & $0.203\,[0.159,\,0.323]$ \\
 & 52 & $0.0698\,[0.0537,\,0.0847]$ & $0.0651\,[0.0575,\,0.0764]$ & $0.223\,[0.177,\,0.388]$ \\
 & 62 & $0.0411\,[0.0298,\,0.0633]$ & $0.0508\,[0.0332,\,0.0579]$ & $0.213\,[0.159,\,0.320]$ \\
\midrule
\multirow{3}{*}{Task~4} & 42 & $0.0673\,[0.0616,\,0.0839]$ & $0.0908\,[0.0640,\,0.136]$ & $0.308\,[0.194,\,0.382]$ \\
 & 52 & $0.0762\,[0.0627,\,0.0928]$ & $0.102\,[0.0708,\,0.136]$ & $0.287\,[0.204,\,0.341]$ \\
 & 62 & $0.117\,[0.105,\,0.135]$ & $0.140\,[0.110,\,0.175]$ & $0.287\,[0.213,\,0.366]$ \\
\bottomrule
\end{tabular}
\end{table}

\begin{table}[htbp]
\centering
\caption{Within-segment constancy relative to the training noise. $\delta_{\mathrm{rms}}$
is the rms length of the deviations $\delta_{s,t}=g_{s,t}-g_{s,0}$, over the same
observations as Table~\ref{tab:constancy_D0}, divided by the rms length $\sigma\sqrt{2n}$
of one training-noise vector ($0.447$ on Tasks~1 and~2, $0.134$ on Task~4). Entries: median $[\,p_{25},p_{75}\,]$ over trajectories, per
trained model.}
\label{tab:constancy_abs}
\setlength{\tabcolsep}{4pt}
\begin{tabular}{llccc}
\toprule
Task & Seed & Train & Test~A & Test~B \\
\midrule
\multirow{3}{*}{Task~1} & 42 & $0.0833\,[0.0718,\,0.0988]$ & $0.0791\,[0.0657,\,0.0957]$ & $0.212\,[0.145,\,0.324]$ \\
 & 52 & $0.0816\,[0.0724,\,0.104]$ & $0.0825\,[0.0693,\,0.0984]$ & $0.223\,[0.175,\,0.394]$ \\
 & 62 & $0.0933\,[0.0865,\,0.0982]$ & $0.0953\,[0.0890,\,0.101]$ & $0.260\,[0.162,\,0.380]$ \\
\midrule
\multirow{3}{*}{Task~2} & 42 & $0.175\,[0.153,\,0.222]$ & $0.187\,[0.175,\,0.233]$ & $1.50\,[1.04,\,2.22]$ \\
 & 52 & $0.327\,[0.255,\,0.380]$ & $0.325\,[0.288,\,0.347]$ & $1.65\,[1.18,\,2.80]$ \\
 & 62 & $0.207\,[0.149,\,0.257]$ & $0.226\,[0.188,\,0.255]$ & $1.58\,[0.973,\,2.20]$ \\
\midrule
\multirow{3}{*}{Task~4} & 42 & $0.303\,[0.268,\,0.403]$ & $0.451\,[0.274,\,0.659]$ & $1.80\,[1.01,\,2.01]$ \\
 & 52 & $0.352\,[0.264,\,0.448]$ & $0.512\,[0.315,\,0.642]$ & $1.60\,[1.06,\,1.99]$ \\
 & 62 & $0.547\,[0.465,\,0.649]$ & $0.695\,[0.492,\,0.880]$ & $1.66\,[1.10,\,2.06]$ \\
\bottomrule
\end{tabular}
\end{table}

\paragraph{Results.}
On the training trajectories, the median $D_0$ is $0.024$--$0.027$ on Task~1, $0.038$--$0.070$ on Task~2
and $0.067$--$0.117$ on Task~4 (Table~\ref{tab:constancy_D0}), with no upper quartile above $0.135$. For the median trajectory, the gauge-fixed latents thus deviate from their segments' frame-free conserved quantities by $2$--$12\%$ of the observed motion, in rms; an encoder that ignored its time input would deviate by $100\%$. In units of the
training noise, the median $\delta_{\mathrm{rms}}$ is $0.082$--$0.093$ on Task~1,
$0.175$--$0.327$ on Task~2 and $0.303$--$0.547$ on Task~4 (Table~\ref{tab:constancy_abs}), with no upper quartile above $0.649$. Thus, for the median trajectory of every model, a segment's observations agree on its initial state, in rms, more closely than the rms length of one training-noise vector. By both measures, on the training trajectories, where the zero set is characterised, the trained models are therefore close to it. On Tasks~1 and~2, Test~A shows no
held-out effect beyond the spread across trajectories: every Test~A median of both metrics lies within the
interquartile range of the same seed's Train values. On Task~4 we observe a mild held-out effect: in both metrics, every Test~A median lies above the upper quartile of the same seed's Train values, exceeding the Train median by factors of
$1.20$--$1.35$ in $D_0$ and $1.27$--$1.49$ in $\delta_{\mathrm{rms}}$. On every task
and seed, the upper quartile of $\delta_{\mathrm{rms}}$ on Test~A remains below $1$, at most $0.880$. Taken together, these comparisons show that on held-out trajectories at the training energies the gauge-fixed latents remain approximately constant in rms in every trained model, for the median trajectory and up to the upper quartile. On Test~B, outside the training energies, the median $D_0$ on Task~1 rises by factors of
$1.51$--$1.82$ relative to Train and the median $\delta_{\mathrm{rms}}$ rises to $0.212$--$0.260$, from $0.082$--$0.093$:
the models are less constant at energies they were not trained on, though still well within the training-noise scale. On Tasks~2 and~4, however, the median $\delta_{\mathrm{rms}}$ exceeds $1$, at
$1.50$--$1.65$ and $1.60$--$1.80$, with $D_0$ at $0.203$--$0.223$ and
$0.287$--$0.308$. By both measures and on every task, constancy therefore weakens outside the training energies. Within Train and Test~A, the median $D_0$ varies across seeds by at most a factor of
$1.31$ on Task~1 and by up to $1.83$ on Tasks~2 and~4. On Test~B, the three seeds' median $D_0$ agree to within $10\%$ on Tasks~2 and~4, a smaller relative spread than on Train and Test~A; the absolute spread also decreases, from $0.032$ and $0.050$ on Train to $0.020$ and $0.021$, so on these tasks the differences between runs seen within the training energies do not grow outside them. On Task~1 they vary by up to a factor of $1.35$, slightly more than on Train ($1.17$) and Test~A ($1.31$).

Section~\ref{sec:convergence} identifies zero prediction loss with constancy of the latents within each segment, on the states and times the training data covers. These results show that the models training produces, at nonzero prediction loss, come close to within-segment constancy, on the training trajectories and on held-out trajectories at the training energies. Outside the training energies the departure from constancy grows. The same
conclusions hold for a model trained without $\mathcal{L}_{\mathrm{HSIC}}$
(Table~\ref{tab:nohsic_consistency}, Appendix~\ref{zero_hsic_experiments}), consistent with the structure being the result of the prediction objective. Within the training energies, an encoder trained to predict is thus, as measured by its gauge-fixed latents, approximately
a map to quantities conserved along each segment. This structure emerges without the form of the latents being prescribed, and it holds on the non-integrable
Task~4, consistent
with~\eqref{eq:flow_form} carrying no integrability requirement.

\paragraph{Recovery of classical conserved quantities from the learned latents.}
\label{sec:latent_recovery} 
Section~\ref{sec:latent_conservation} established that the latents remain approximately constant within each trajectory segment, consistent with their interpretation as the conserved quantities $(\alpha,\beta)$. It does so, however, by removing the frame, leaving open which frame training settles on and what quantities its latents represent. To find out, 
we relate the latents to the conserved quantities classical mechanics uses to characterise a trajectory: the energy and the initial phase, a set long established and immediately interpretable. Beyond interpretability, relating to them offers a second advantage: in the systems we consider their behaviour under a shift of the observation time is known in closed form. This lets us test whether $\mathcal{M}$'s dependence on its time input reproduces the known time-dependence of $(E,\phi)$---that is, whether the encoder's action in $t$ is what \eqref{eq:flow_form} prescribes, and not merely some map that holds a quantity fixed.

The two integrable systems supply the $(E,\phi)$ set analytically, letting us test
recovery directly. We additionally include the Hénon--Heiles system, restricted to its energy: being non-integrable it admits no
global and analytic
$(E,\phi)\!\leftrightarrow\!(q,p)$ map. Its energy, however, is a global invariant known in closed form, so the energy recovery test extends to a non-integrable system. For the ideal mass-spring (ims) the energy is $E^{\mathrm{ims}}=\tfrac12(q^2+p^2)$
and the initial phase $\phi^{\mathrm{ims}}=\operatorname{atan2}(q_0,p_0)$, while for
the ideal pendulum (ip) the energy is
$E^{\mathrm{ip}}=\tfrac12 p^{2}+\omega_0^{2}(1-\cos q)$, with natural frequency
$\omega_0=\sqrt3$, and the initial
action--angle phase $\phi^{\mathrm{ip}}=\frac{\pi}{2K(m)}\,F(\psi\,|\,m)$, where
$m=E^{\mathrm{ip}}/2\omega_0^{2}$,
$\psi=\operatorname{atan2}\!\big(\sin(q_0/2)/\sqrt{m},\,p_0/\sqrt{2E^{\mathrm{ip}}}\big)$,
and $K(m)$, $F(\psi\,|\,m)$ are the complete and incomplete elliptic integrals
of the first kind. For the two integrable systems, we form a regular $40\times40$ grid of $1600$ pairs $(E,\phi)$, evenly spaced in $E$ over the training range ($[0.2,1.0]$ for the ideal mass-spring, $[1.3,2.3]$ for the ideal pendulum) and in $\phi$ over $[-\pi,\pi)$.

Each pair is mapped to phase space by the analytical solution of each system:
Ideal mass-spring:
\begin{equation}
q = \sqrt{2E}\,\sin\phi, \qquad p = \sqrt{2E}\,\cos\phi,
\label{eq:ic_map}
\end{equation}
Ideal pendulum:
\begin{equation}
q = 2\arcsin\!\big(k\,\operatorname{sn}(\xi,m)\big), \qquad
p = \sqrt{2E}\,\operatorname{cn}(\xi,m), \qquad
\xi = \frac{2K(m)}{\pi}\,\phi,
\label{eq:ic_map_pend}
\end{equation}
with $k=\sqrt{m}$, and $\operatorname{sn},\operatorname{cn}$ the Jacobi elliptic
functions.
For Hénon--Heiles we instead sample the phase space directly on energy shells. We take $40$
energies evenly spaced over the training range $E\in[0.08,0.125]$ and, for each, draw
$200$ states on the exact shell $H(q,p)=E$ that fall inside the training coordinate
box, giving $8000$ states in total each being the $t=0$ state of a trajectory of that energy.

By construction~\eqref{eq:ic_map}--\eqref{eq:ic_map_pend}, each $(q_j,p_j)$ is the $t=0$ state of the trajectory with conserved quantities $(E_j,\phi_j)$; we therefore encode every grid point at $t=0$,
\begin{equation}
    (Q_j,\,P_j) = \mathcal{M}(q_j,\,p_j,\,0),
    \qquad j=1,\dots,1600,
    \label{eq:encode_grid}
\end{equation}
to obtain the paired data $\{(Q_j,P_j),(E_j,\phi_j)\}$. On these pairs we fit polynomial read-outs of $(E,\phi)$ on $(Q,P)$ of increasing degree, starting from degree~$1$. We repeat the analysis for each of the three trained models (seeds $42$, $52$, $62$) on the same states; every read-out is fit on that model's own latents. We use polynomials for two reasons: First, the fit is a convex least-squares problem, so a recovery failure reflects the latents rather than the read-out. Second, the degree is an interpretable complexity dial: sweeping it reveals the minimal complexity needed and whether added capacity helps or harms generalisation. For energy we regress $\hat E=f(Q,P)$ directly. Phase wraps at $\pm\pi$ and is therefore discontinuous in $(Q,P)$, which a smooth polynomial cannot represent. We instead regress the continuous circle coordinates $\cos\phi$ and $\sin\phi$ as polynomials in $(Q,P)$ and recover $\hat\phi=\operatorname{atan2}(\widehat{\sin\phi},\widehat{\cos\phi})$, letting $\operatorname{atan2}$ handle the wrap. We summarise each fit by its $5$-fold cross-validated $R^2$, computed from out-of-fold predictions, and by the median of its per-point errors. For energy the $R^2$ is that of $\hat E = f(Q,P)$ and the error is the relative residual $|\Delta E|/E = |\hat E - E|/E$. For phase the $R^2$ values are those of $\cos\hat\phi$ and $\sin\hat\phi$ against $\cos\phi$ and $\sin\phi$, and the error is the wrapped angular error
$\Delta\phi = \big|\operatorname{atan2}\!\big(\sin(\hat\phi-\phi),\,
\cos(\hat\phi-\phi)\big)\big|$, reported in degrees.

\begin{table}[htbp]
\centering
\caption{Energy recovery $\hat E=f(Q,P)$ by polynomial degree, all three
energy-conserving systems. $R^2$ is $5$-fold cross-validated; $|\Delta E|/E$ is the
median relative error in \% ($1600$ grid states for the integrable systems, $8000$
shell-sampled states for Hénon--Heiles). Each entry is the range [min, max] over the three seeds,
or a single value where they agree at the shown precision.}
\label{tab:energy_recovery}
\setlength{\tabcolsep}{4pt}
\begin{tabular}{ccccccc}
\toprule
& \multicolumn{2}{c}{Ideal mass-spring} & \multicolumn{2}{c}{Ideal pendulum}
& \multicolumn{2}{c}{Hénon--Heiles} \\
\cmidrule(lr){2-3}\cmidrule(lr){4-5}\cmidrule(lr){6-7}
Degree & $R^2$ & $|\Delta E|/E$ & $R^2$ & $|\Delta E|/E$ & $R^2$ & $|\Delta E|/E$ \\
\midrule
1 & $[-0.001, 0.000]$ & $[30.8, 31.0]$ & $-0.001$ & $13.8$ & $[-0.001, 0.000]$ & $11.1$ \\
2 & $[0.963, 0.996]$ & $[1.9, 6.1]$ & $[0.800, 0.815]$ & $[5.5, 5.7]$ & $[0.471, 0.597]$ & $[5.8, 6.9]$ \\
3 & $0.999$ & $0.7$ & $[0.949, 0.974]$ & $[1.8, 2.6]$ & $[0.845, 0.950]$ & $[1.8, 3.6]$ \\
4 & $1.000$ & $[0.3, 0.6]$ & $[0.990, 0.995]$ & $[0.9, 1.2]$ & $[0.944, 0.971]$ & $[1.4, 2.1]$ \\
\bottomrule
\end{tabular}
\end{table}

\begin{table}[htbp]
\centering
\caption{Phase recovery by polynomial degree, for both integrable systems. The
angle is recovered as $\hat\phi=\operatorname{atan2}(\widehat{\sin\phi},\widehat{\cos\phi})$ from a joint regression of $(\cos\phi,\sin\phi)$ on $(Q,P)$;
$R^2$ ($5$-fold cross-validated) is that of $\cos\hat\phi$ and $\sin\hat\phi$, and
$\Delta\phi$ is the median wrapped angular error $\big|\operatorname{atan2}(
\sin(\hat\phi-\phi),\cos(\hat\phi-\phi))\big|$ in degrees. Each entry is the range [min, max]
over the three seeds, or a single value where they agree at the shown precision.}
\label{tab:phase_recovery}
\setlength{\tabcolsep}{6pt}
\begin{tabular}{ccccccc}
\toprule
& \multicolumn{3}{c}{Ideal mass-spring} & \multicolumn{3}{c}{Ideal pendulum} \\
\cmidrule(lr){2-4}\cmidrule(lr){5-7}
& \multicolumn{2}{c}{$R^2$} & $\Delta\phi$ (deg) & \multicolumn{2}{c}{$R^2$} & $\Delta\phi$ (deg) \\
\cmidrule(lr){2-3}\cmidrule(lr){5-6}
Degree & $\cos\phi$ & $\sin\phi$ & median
       & $\cos\phi$ & $\sin\phi$ & median \\
\midrule
1 & $[0.998, 1.000]$ & $0.999$ & $[0.78, 1.83]$ & $[0.995, 0.998]$ & $[0.996, 0.998]$ & $[1.62, 2.50]$ \\
2 & $1.000$ & $1.000$ & $[0.30, 0.52]$ & $[0.999, 1.000]$ & $[0.999, 1.000]$ & $[0.76, 1.37]$ \\
3 & $1.000$ & $[0.999, 1.000]$ & $[0.36, 1.19]$ & $1.000$ & $1.000$ & $[0.81, 0.91]$ \\
4 & $1.000$ & $1.000$ & $[0.15, 0.19]$ & $1.000$ & $1.000$ & $[0.20, 0.48]$ \\
\bottomrule
\end{tabular}
\end{table}

For each quantity we select, on seed~$42$, the lowest polynomial degree reaching $R^2 \ge 0.90$ under cross-validation (for phase, on both circle coordinates), and hold it fixed for the other seeds, as for all training hyperparameters. This gives degree~$2$ (oscillator), degree~$3$ (pendulum) and degree~$4$ (Hénon--Heiles) for energy, and degree~$1$ for phase in both integrable systems; every seed clears the threshold at these degrees (Tables~\ref{tab:energy_recovery}--\ref{tab:phase_recovery}). Separately, for each model, each chosen polynomial is refit once on all points to serve as the fixed read-out instrument used in the following test. The recovery test fixed the encoder at $t=0$; it leaves untested how the mapping behaves
at other encoding times~$\tau$. Holding a state $(q_0,p_0)$ fixed and increasing the
observation time $\tau$ leaves the true energy unchanged and shifts the true initial phase
by $-\omega\tau$ for the oscillator ($\omega=\sqrt{k/m}=1$) and by
$-\Omega(E)\,\tau$ for the pendulum, whose anharmonic rate $\Omega(E)=\pi\omega_0/2K(m)$
decreases with energy and so varies from state to state. If training has brought $\mathcal{M}$ near a map of the form Section~\ref{sec:convergence} characterises, its latents, through the read-outs, should inherit the time-shift behaviour of the known $(E,\phi)$ conserved quantities. We test exactly this.

We reuse the same $40\times40$ grid introduced above. Holding
each state fixed, we re-encode it at $n_\tau$ times $\tau$ evenly spaced over $[0,T_{\max}]$,
\begin{equation}
    Q(\tau),\,P(\tau) = \mathcal{M}(q_0,\,p_0,\,\tau),
    \label{eq:time_sweep}
\end{equation}
using $n_\tau=32$ over $T_{\max}\approx6.49$ for the oscillator and $n_\tau=30$
over $T_{\max}\approx4.06$ for the pendulum. To each
latent produced, we apply the fixed read-outs, obtaining
a recovered energy $\hat E(\tau)=f(Q(\tau),P(\tau))$ and a recovered phase $\hat\phi(\tau)$ at every sampled time and state---$1600\times n_\tau$ recovered
$(\hat E,\hat\phi)$ values in total. For each fixed state this gives two sequences (one for energy and one for phase) of
$n_\tau$ time-shifted recoveries, which we summarise by fitting a straight line to
each by least squares. Hénon--Heiles contributes an energy
sequence only: its $8000$ shell states are re-encoded at $n_\tau=58$ times over $T_{\max}\approx8.563$. For energy, treated identically across all three systems, we report the per-state fitted slope
$\mathrm{d}\hat E/\mathrm{d}\tau$ (median and IQR over states), which constant
$E$ would drive to $0$. We also check the value itself: the per-$\tau$ median of the relative error $|\hat E-E|/E$,
summarised over the $n_\tau$ times, in the units of the $t=0$ error of Table~\ref{tab:energy_recovery} (Table~\ref{tab:time_shift_energy}).

For phase the two systems differ in kind. The oscillator advances at a single frequency
$-\omega=-1$ for every state, so we report the median and IQR of the per-state fitted slope
$\mathrm{d}\hat\phi/\mathrm{d}\tau$ over the $1600$ states, with target $-1$
(Table~\ref{tab:time_shift_phase}). The pendulum advances at the per-state anharmonic
rate $-\Omega(E)$, so a single slope is not the target; instead, for each state we form the
absolute deviation $|\mathrm{d}\hat\phi/\mathrm{d}\tau+\Omega(E)|$ between its fitted slope
and its own $-\Omega(E)$, and report the median and IQR of these $1600$ deviations,
with target $0$ (Table~\ref{tab:time_shift_phase}). To make the energy dependence itself
visible we additionally bin the $1600$ states into five equal-width energy bins
spanning the training range and take the median fitted slope within each bin, comparing it against the theoretical $-\Omega(E)$
evaluated at the bin centre; sweeping the bins traces out the anharmonic frequency
curve (Table~\ref{tab:anharmonic_curve}). In both systems the recovered value $\hat\phi(\tau)$ is an
angle returned on $(-\pi,\pi]$ and therefore jumps by $2\pi$ each time it crosses
that boundary; before fitting we add back these $2\pi$ steps so that $\hat\phi(\tau)$
is a continuous function of $\tau$ (unwrapping). We additionally report the
linear-fit residual (in degrees), measuring how closely $\hat\phi$ follows the
fitted line, and the discrepancy between the fitted intercept and the true initial
phase $\phi_0$, which tests whether $\hat\phi(0)$ recovers the correct absolute phase
rather than only the correct rate.

\begin{table}[htbp]
\centering
\caption{Time-shift test, energy, all three energy-conserving systems, per seed. Each grid or
shell-sampled state is re-encoded across the $\tau$ sweep of \eqref{eq:time_sweep}
and $\hat E(\tau)$ is read out by that system's recovered energy relation (degree~$2$
oscillator, degree~$3$ pendulum, degree~$4$ Hénon--Heiles). The per-state slope
$\mathrm{d}\hat E/\mathrm{d}\tau$ is summarised across states; the relative
error $|\hat E-E|/E$ (\%) is the per-$\tau$ median across states, summarised over $\tau$.
Energy preservation gives zero slope and error. Entries: median $[\,p_{25},p_{75}\,]$.}
\label{tab:time_shift_energy}
\setlength{\tabcolsep}{6pt}
\begin{tabular}{lccc}
\toprule
Metric & Seed~42 & Seed~52 & Seed~62 \\
\midrule
\multicolumn{4}{l}{\textit{Ideal mass-spring}} \\
\quad Slope $\mathrm{d}\hat E/\mathrm{d}\tau$ & $0.001\,[-0.003,\,0.003]$ & $0.000\,[-0.003,\,0.003]$ & $0.000\,[-0.002,\,0.003]$ \\
\quad Error $|\hat E-E|/E$ & $5.0\,[4.7,\,5.5]$ & $2.8\,[2.2,\,3.5]$ & $4.3\,[3.6,\,5.3]$ \\
\addlinespace
\multicolumn{4}{l}{\textit{Ideal pendulum}} \\
\quad Slope $\mathrm{d}\hat E/\mathrm{d}\tau$ & $-0.009\,[-0.018,\,0.016]$ & $0.003\,[-0.020,\,0.044]$ & $-0.002\,[-0.015,\,0.009]$ \\
\quad Error $|\hat E-E|/E$ & $3.9\,[3.2,\,5.3]$ & $5.1\,[4.4,\,6.5]$ & $3.6\,[3.1,\,4.9]$ \\
\addlinespace
\multicolumn{4}{l}{\textit{H\'enon--Heiles}} \\
\quad Slope $\mathrm{d}\hat E/\mathrm{d}\tau$ & $0.0000\,[-0.0003,\,0.0003]$ & $0.0000\,[-0.0002,\,0.0002]$ & $0.0001\,[-0.0009,\,0.0011]$ \\
\quad Error $|\hat E-E|/E$ & $2.4\,[2.2,\,2.8]$ & $2.3\,[2.1,\,2.6]$ & $5.2\,[4.1,\,5.8]$ \\
\bottomrule
\end{tabular}
\end{table}

\begin{table}[htbp]
\centering
\caption{Time-shift test, phase, both integrable systems, per seed. Same sweep as
Table~\ref{tab:time_shift_energy}; $\hat\phi(\tau)$ is read out by the degree-$1$ phase
relation, unwrapped in $\tau$, and line-fitted per state. Oscillator: the per-state fitted
slope $\mathrm{d}\hat\phi/\mathrm{d}\tau$, with target $-\omega=-1$. Pendulum: the per-state
deviation $|\mathrm{d}\hat\phi/\mathrm{d}\tau+\Omega(E)|$ (rad/time) of the fitted slope from
the state's own anharmonic rate $-\Omega(E)=-\pi\omega_0/2K(m)$, with target $0$. The
intercept error is the wrapped angular distance between the fitted intercept and the true
initial phase $\phi_0$; the residual is the per-point deviation of $\hat\phi$ from the fitted
line across all $(\text{state},\tau)$ pairs; both have target $0$. Entries: median
$[\,p_{25},p_{75}\,]$ across the $1600$ states.}
\label{tab:time_shift_phase}
\setlength{\tabcolsep}{4pt}
\begin{tabular}{lccc}
\toprule
Metric & Seed~42 & Seed~52 & Seed~62 \\
\midrule
\multicolumn{4}{l}{\textit{Ideal mass-spring}} \\
\quad Slope $\mathrm{d}\hat\phi/\mathrm{d}\tau$ & $-0.998\,[-1.004,\,-0.991]$ & $-1.001\,[-1.005,\,-0.997]$ & $-1.000\,[-1.003,\,-0.997]$ \\
\quad Intercept err.\ (deg) & $1.33\,[0.58,\,2.07]$ & $0.80\,[0.44,\,1.15]$ & $0.80\,[0.45,\,1.16]$ \\
\quad Residual (deg) & $1.80\,[0.87,\,3.10]$ & $1.03\,[0.45,\,1.97]$ & $1.21\,[0.57,\,2.13]$ \\
\addlinespace
\multicolumn{4}{l}{\textit{Ideal pendulum}} \\
\quad $|\mathrm{d}\hat\phi/\mathrm{d}\tau+\Omega(E)|$ & $0.020\,[0.010,\,0.032]$ & $0.020\,[0.010,\,0.032]$ & $0.020\,[0.010,\,0.032]$ \\
\quad Intercept err.\ (deg) & $1.39\,[0.68,\,2.33]$ & $1.11\,[0.48,\,1.81]$ & $1.28\,[0.41,\,2.89]$ \\
\quad Residual (deg) & $2.05\,[1.02,\,3.46]$ & $2.10\,[1.01,\,3.55]$ & $1.93\,[0.92,\,3.40]$ \\
\bottomrule
\end{tabular}
\end{table}

\begin{table}[!htbp]
\centering
\caption{Anharmonic frequency curve, ideal pendulum. The $1600$ states are binned by energy into five equal-width bins spanning the
training range. For each bin we report (in rad/time) the theoretical phase slope
$-\Omega(E)=-\pi\omega_0/2K(m)$ evaluated at the bin centre and, per seed, the median over
the states in that bin of the per-state least-squares slopes
$\mathrm{d}\hat\phi/\mathrm{d}\tau$ (the slopes underlying Table~\ref{tab:time_shift_phase}).}
\label{tab:anharmonic_curve}
\setlength{\tabcolsep}{8pt}
\begin{tabular}{ccccc}
\toprule
& & \multicolumn{3}{c}{Fitted slope (median)} \\
\cmidrule(lr){3-5}
Energy bin & $-\Omega(E)$ (theory) & Seed~42 & Seed~52 & Seed~62 \\
\midrule
$[1.300,1.497]$ & $-1.623$ & $-1.592$ & $-1.593$ & $-1.591$ \\
$[1.497,1.695]$ & $-1.606$ & $-1.590$ & $-1.592$ & $-1.589$ \\
$[1.695,1.892]$ & $-1.588$ & $-1.587$ & $-1.591$ & $-1.583$ \\
$[1.892,2.090]$ & $-1.570$ & $-1.584$ & $-1.589$ & $-1.579$ \\
$[2.090,2.287]$ & $-1.551$ & $-1.581$ & $-1.586$ & $-1.575$ \\
\bottomrule
\end{tabular}
\end{table}

\begin{figure}[!htbp]
\centering
\includegraphics[width=\textwidth]{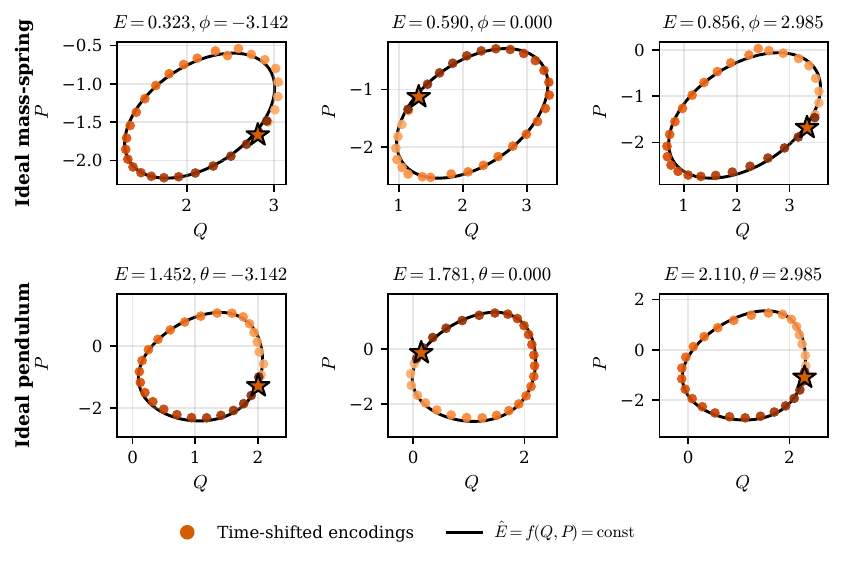}
\caption{Energy and phase signatures for representative states. Each panel fixes one state
$(q_0,p_0)$ (its $E$ and initial phase $\phi_0$ printed above, shown as $\phi$ for the oscillator and $\theta$ for the pendulum) and re-encodes it across
$\tau\in[0,T_{\max}]$ as in \eqref{eq:time_sweep} using the seed 42 model; top row, ideal mass-spring; bottom row,
ideal pendulum; three representative states per row, one per energy tertile at spread-out
phases. The black curve is the level set $\hat E=f(Q,P)=E_c$ of the recovered relation at
the state's independently known true energy $E_c$ 
(closed-form ellipse from the degree-$2$ relation for the oscillator; numerical contour of the degree-$3$ relation for the pendulum). Time-shifted encodings
$(Q(\tau),P(\tau))$ are coloured by $\tau$ (light to dark), the star marking $\tau=0$.}
\label{fig:time_shift_energy}
\end{figure}

For the harmonic-oscillator energy relation (degree~$2$), the symmetric
matrix of its $Q^2,\,QP,\,P^2$ coefficients is positive-definite in every seed. A positive-definite quadratic part guarantees closed
elliptical level sets $\hat E=c=\text{const}$ for every $c$ above the relation's
minimum $\hat E_{\min}$, which is below $0.025$ in every seed; all training energies ($E\ge0.2$) lie in this range, so each corresponds to a single closed ellipse in the $(Q,P)$ plane. The pendulum relation (degree~$3$) is cubic and carries no such guarantee:
a degree-$3$ surface may have open or multi-branch level sets, so closedness cannot
be asserted a priori. For the seed-$42$ model displayed in Figure~\ref{fig:time_shift_energy}, we nonetheless find the level set to be a single closed contour at the energies we examine. The experiment overlays two independently produced
objects to test this geometry. The level set is the curve $\{(Q,P):\hat E(Q,P)=E_c\}$
of the recovered relation, drawn at that state's independently known true
energy $E_c$---obtained in closed form as the conic root for the oscillator, and
numerically (the recovered relation evaluated on a dense $(Q,P)$ mesh, contoured at
$E_c$) for the cubic pendulum. The encodings are the time-shifted latents
$(Q(\tau),P(\tau))$ from \eqref{eq:time_sweep}, computed separately by re-encoding
the fixed state across $\tau\in[0,T_{\max}]$. If $\mathcal{M}$ preserves energy under its time action, the encodings should stay close to the contour drawn at the true energy, up to the read-out error of Table~\ref{tab:energy_recovery}, rather than drifting across level sets; this is the geometric counterpart of the zero energy slope. At fixed energy, the remaining freedom is the position along the contour, which the phase read-out maps to the phase (Table~\ref{tab:phase_recovery}). The true initial phase shifts by $2\pi$ over one period, and each sweep spans approximately one period ($T_{\max}\approx6.49$ against $2\pi/\omega\approx6.28$ for the oscillator; $T_{\max}\approx4.06$ against $2\pi/\Omega(E)\approx3.87$–-$4.05$ across the energy bins of Table~\ref{tab:anharmonic_curve} for the pendulum). If $\mathcal{M}$ also reproduces the phase shift, the encodings should therefore travel approximately once around the contour. Figure~\ref{fig:time_shift_energy} shows this
for three representative states per system, spanning the training energy range (one
per energy tertile) at spread-out initial phases. This planar level-set picture is specific to the two-dimensional latent space of the integrable systems; Hénon--Heiles, with four latents, admits no such visualisation.

\paragraph{Results.}
For all three systems a linear read-out captures essentially none of the energy's
variance ($R^2\approx0$),
while a low-order polynomial captures most of it. For the oscillator a quadratic already reaches $R^2=0.963$--$0.996$; the pendulum falls short of the $R^2\ge0.90$ threshold at degree~$2$ ($R^2=0.800$--$0.815$) and clears it at degree~$3$ ($R^2=0.949$--$0.974$); and Hénon--Heiles clears it at degree~$4$ ($R^2=0.944$--$0.971$), in seed~$52$ already at degree~$3$ ($R^2=0.950$)
(Table~\ref{tab:energy_recovery}). Using these relations, the per-state slope of the recovered energy, $\mathrm{d}\hat E/\mathrm{d}\tau$,
scatters about zero in every system and seed, with an interquartile range containing zero
(Table~\ref{tab:time_shift_energy}). Every read-out was fit at $t=0$ alone, so for $\tau>0$ the errors
along the sweep are out-of-sample in $\tau$. Summarised over the sweep, the median relative errors $|\hat E-E|/E$ are $2.8$--$5.0\%$ for the oscillator,
$3.6$--$5.1\%$ for the pendulum and $2.3$--$5.2\%$ for Hénon--Heiles. For the oscillator this range
overlaps that of the $t=0$ errors ($1.9$--$6.1\%$, Table~\ref{tab:energy_recovery}); for the pendulum
and Hénon--Heiles every value exceeds every $t=0$ error (at most $2.6\%$ and $2.1\%$). On its own, zero energy-slope is necessary but not sufficient: an encoder that ignored
its time input would produce latents constant in $\tau$ and satisfy it trivially. The phase results
below exclude this directly for the two integrable systems: the same encodings
$(Q(\tau),P(\tau))$ yield a recovered phase advancing at a nonzero rate. For all three systems, including H\'enon--Heiles, which has no phase read-out, Section~\ref{sec:latent_conservation} excludes such an encoder as well: it would give $D_0=1$, whereas every trained model has an upper quartile of $D_0$ of at most $0.175$ on the held-out trajectories at the training energies.

Phase recovers in a complementary way. In both systems it is already captured to
within a few degrees of arc by the degree-$1$ read-out (median errors of $0.78$--$1.83^\circ$ for the oscillator and $1.62$--$2.50^\circ$ for the pendulum, at $R^2>0.99$
on both circle coordinates). Degree~$4$ lowers them to $0.15$--$0.19^\circ$ and $0.20$--$0.48^\circ$, but not monotonically
in degree: the oscillator's degree-$3$ errors reach $1.19^\circ$, above every degree-$2$ value, so added capacity
does not reliably help (Table~\ref{tab:phase_recovery}). The time-shift test is where
the analyses of the two systems separate. For the
oscillator the recovered phase decreases at a single slope, with medians of $-1.001$ to $-0.998$ across seeds and every
interquartile range bracketing the expected $-\omega=-1$; median linear-fit residuals of $1.03$--$1.80^\circ$ show it does so at a steady rate, and median
intercept errors of $0.80$--$1.33^\circ$ show that the fitted lines start near the true initial phase (Table~\ref{tab:time_shift_phase}). For the pendulum no single slope is expected:
each state should advance at its own anharmonic rate $-\Omega(E)$. The recovered per-state slopes lie close to these rates: the median deviation from each
state's own $-\Omega(E)$ is $0.020$~rad/time in every seed, about $1.3\%$ of the rate
$\Omega\approx1.6$, with residuals ($1.93$--$2.10^\circ$) and intercept errors ($1.11$--$1.39^\circ$) comparable to the
oscillator's (Table~\ref{tab:time_shift_phase}). Binning by energy then isolates the energy dependence of the rate: across the five bins the median fitted slope decreases
monotonically in magnitude with energy in every seed, in the same direction as the theoretical
slowdown $-\Omega(E)$ (Table~\ref{tab:anharmonic_curve}), consistent with a real energy dependence rather than a single-bin artefact. Its magnitude, however, is underestimated: across the five bin centres the true rate
falls by $0.072$~rad/time (from $1.623$ to $1.551$, about $4\%$), while the fitted medians fall by $0.007$--$0.016$. So, the recovered rates reproduce the direction of this gentle gradient
while flattening it to roughly $10$--$22\%$ of its extent. These measurements do not separate whether the
compression originates in the latents or in the degree-$1$ read-out. In every seed the bin-level discrepancies are largest in the two extreme bins ($0.024$--$0.035$) and smallest in the
middle one ($0.001$--$0.005$), the pattern consistent with a trend flattened about the centre of the range produces. Figure~\ref{fig:time_shift_energy} (seed~$42$) illustrates the energy and phase signatures for representative states of the two integrable systems: the time-shifted encodings stay close to the level set drawn at each state’s independently known true energy, and travel approximately once around it over the sweep. This is consistent with the near zero energy slopes and the near-true phase rates of Tables~\ref{tab:time_shift_energy} and~\ref{tab:time_shift_phase}.

The prediction objective does not determine which frame the encoders adopt (Appendix~\ref{app:frame_invariance}), and no
training term references $E$ or $\phi$. A frame in which the classical
conserved quantities are highly nonlinear functions of the latents would therefore incur no
penalty under
$\mathcal{L}_{\mathrm{pred}}$. Because the read-outs act on encodings at $t=0$, they probe exactly the frame each run adopts. Nevertheless, every run settles on a frame in which the latents are interpretable in an operational sense: low-degree polynomial read-outs, fit at $t=0$ alone, approximately recover the classical conserved quantities and track their response to a shift of the encoding time. On unseen states at the training energies, a polynomial of the latents of degree at most four approximates the energy (cross-validated $R^2\ge0.944$), and degree-one polynomials approximate the circle coordinates of the initial phase in the two integrable systems ($R^2>0.99$). This outcome does not depend on $\mathcal{L}_{\mathrm{HSIC}}$: when the model
is retrained without it, all five read-outs (energy in the three systems,
phase in the two integrable ones) still reach $R^2 \ge 0.90$ at their
selected degrees (Table~\ref{tab:nohsic_recovery}). That the result holds across all three seeds and all three systems, and
persists without $\mathcal{L}_{\mathrm{HSIC}}$, points to a systematic
inductive bias rather than chance. We have not isolated which part
of the model or training setup is responsible, and leave this open. The time-shift test adds the dynamical half: re-encoding a fixed state across $\tau$ and reading the latents out through
the fitted polynomials reproduces the signatures of the analytically known
conserved quantities, with energy slopes scattered about zero and phase
advancing at close to the true rate, although for the pendulum the energy
dependence of that rate is recovered in direction but compressed in
magnitude. These results show that the models training produces, at nonzero prediction loss, carry, approximately, the structure that zero prediction loss forces exactly on the training data, on states not seen during training that span the training energy range.

\paragraph{Prediction over composed segments.}
\label{sec:prediction_accuracy}
The statements of Section~\ref{sec:convergence} concern single segments, and so do the preceding experiments. The experiment of section~\ref{sec:latent_conservation} in particular encodes every observation of a segment and decodes each latent only at a single time, $t=0$. At inference HJN does the opposite: it encodes a single observation and predicts the remaining states from it. Over horizons longer than $T_{\max}$ it composes segments, each after the first encoded from the state predicted at the preceding boundary (Section~\ref{sec:autoregressive}), so that errors can accumulate across segments. We evaluate HJN in this mode, asking how its error and predicted energy behave as segments are composed, under energy extrapolation and, on Task~3, outside the theory's scope. To give the error
values a scale, we trained as a published reference the symmetry-learning
Hamiltonian neural network of Noether’s Razor~\cite{van2024noether}, chosen because, like HJN,
it exploits discovered conserved quantities within the Hamiltonian formalism. Its configuration, the modifications aligning it with our protocol, and a check that the trained model reaches the optimisation regime of the original work are given in Appendix~\ref{app:baseline_details}. A single comparator supports no ranking
among Hamiltonian learners, in accuracy or in cost, and we draw none.

\begin{figure}[!htbp]
\centering
\includegraphics[width=\textwidth]{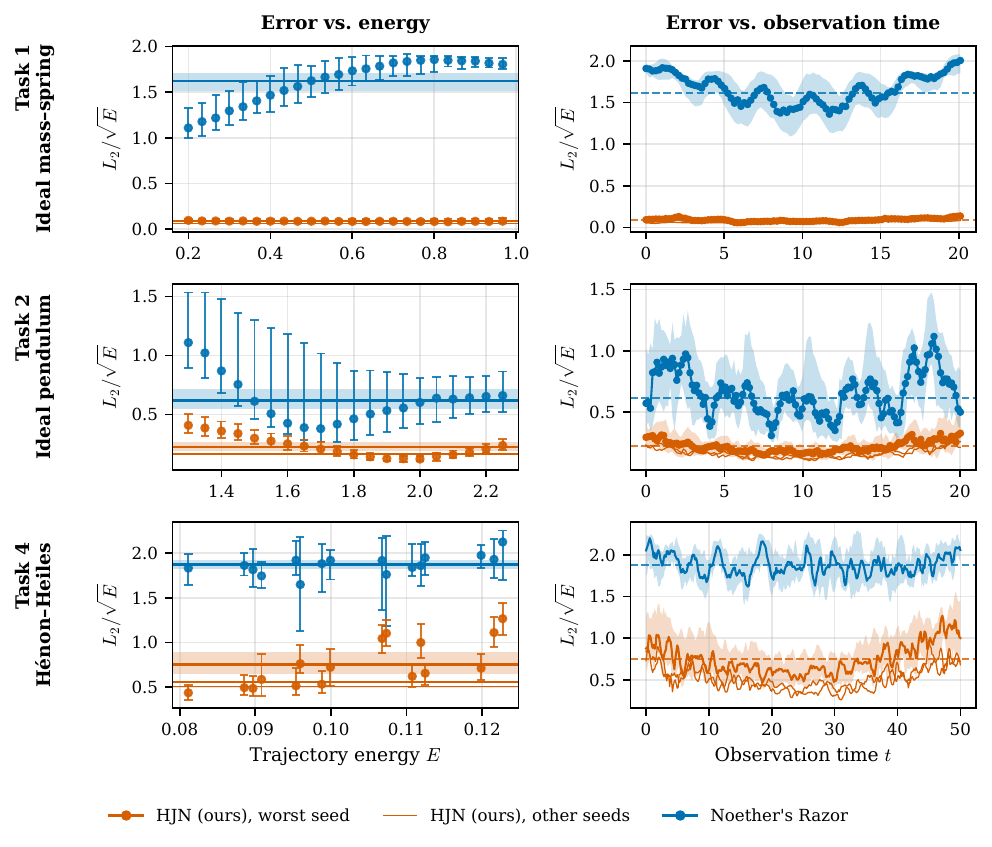}
\caption{Per-task accuracy structure on the Test~A
(rollout-extrapolation) sets, HJN
(\textcolor[HTML]{D55E00}{orange}), Noether’s Razor (\textcolor[HTML]{0072B2}{blue}). Rows: Tasks~1, 2 and~4. Columns, left to right: (a)~$\sqrt{E}$-normalised
reconstruction error vs.\ trajectory energy, where each marker is a
trajectory's median error across observation choices, error bars give
the across-observation interquartile range (IQR), the solid horizontal line marks the
cross-trajectory mean of these per-trajectory medians, and the shaded
band is its $95\%$ bias-corrected and accelerated (BCa) bootstrap confidence interval (CI) ($10{,}000$ resamples); (b)~error
vs.\ observation time, with the median across trajectories (solid),
the across-trajectory IQR (shaded band), and the cross-trajectory mean of medians (dashed). HJN is shown for the seed with the
largest cross-trajectory mean (seed~52 on Tasks~1--2, seed~62 on Task~4); thin
orange lines mark the other two seeds' means in (a) and their medians across
trajectories in (b).
Test~B (energy-extrapolation) figures are shown in
Appendix~\ref{app:supplementary_figures}.}
\label{fig:summary_statistics}
\end{figure}

Since the predicted trajectory depends on the selected observation point
$z_{\mathrm{obs}}$, we evaluate inference from
every observation index to avoid conditioning the reported accuracy on a
particular choice.
For each observation index
$m\in\{1,\dots,N_{\mathrm{obs}}\}$, each method's inference protocol (Algorithm~\ref{alg:handoff} for the proposed method) is executed as a batched inference call over all
$N_{\mathrm{traj}}$ trajectories, yielding one reconstructed trajectory per
trajectory-observation pair. For a trajectory with ground-truth states $(q_t,p_t)$ and predictions
$(\hat q_t,\hat p_t)$, we
define the reconstruction error as the mean over prediction times of the
per-time-point Euclidean distance in the $2n$-dimensional phase space,
$\bigl\|(\hat q_t - q_t,\; \hat p_t - p_t)\bigr\|_2
= \sqrt{\sum_{i=1}^{n}(\hat q_{t,i} - q_{t,i})^2
      + \sum_{i=1}^{n}(\hat p_{t,i} - p_{t,i})^2}$. Collecting these errors
over all trajectory--observation pairs yields a loss matrix
$\mathcal{L}\in\mathbb{R}^{N_{\mathrm{obs}}\times N_{\mathrm{traj}}}$, whose
entry $\mathcal{L}_{mj}$ is the reconstruction error for trajectory $j$ when
conditioned on observation index $m$. Because the state magnitude  $\|(q,p)\|_2$ scales approximately as $\sqrt{E}$ on
Tasks~1 and~2, we divide each trajectory's error by $\sqrt{E}$ so that errors
are comparable across energies. The normalisation controls for the energy
scale only; the Test~A and Test~B settings also differ in horizon. For consistency, we report the same metric for Task~4 too. To characterise sensitivity to the observation choice, each trajectory is
summarised by the median reconstruction error across observation indices,
while the interquartile range (IQR) across observation indices is visualised in
Fig.~\ref{fig:summary_statistics}. Reported accuracy values in Table~\ref{tab:main_results} correspond to the
cross-trajectory mean of these per-trajectory median errors together with a
$95\%$ bias-corrected and accelerated (BCa) bootstrap confidence interval computed from 10{,}000 resamples, for each trained model separately. For completeness, we report inference efficiency and maximum incremental GPU memory
allocation of the seed-42 models for this aggregate experiment in Appendix~\ref{sec:inference_efficiency}.

Aggregate accuracy can mask how error behaves over a long rollout, so we
next ask how prediction error, energy conservation, and phase-space
fidelity evolve over time. We assess these aspects on one representative trajectory per extrapolation setting, and on the held-out tail of Task~3. We define the representative trajectory as the test trajectory whose per-trajectory median reconstruction error (defined above) is closest to the median of the corresponding errors across all test trajectories in the same test set, as evaluated by the seed-42 model. The same trajectory is used when evaluating the other seeds and the reference. As in the aggregate evaluation, we run inference from every observation $z_m=(q_m,p_m,t_m)$, $m\in\{1,\dots,N_{\mathrm{obs}}\}$, of this trajectory, each run producing a predicted trajectory $(\hat q^{(m)}(t), \hat p^{(m)}(t))$ over the full temporal domain. At every prediction time, we summarise the pointwise reconstruction
error across all $N_{\mathrm{obs}}$ evaluations using the median and interquartile
range (IQR). The pointwise reconstruction error is defined as the squared error in phase space,
\[
\mathrm{SE}^{(m)}(t_i)
=
\sum_{k=1}^{n}\bigl(\hat q^{(m)}_k(t_i)-q_k(t_i)\bigr)^2
+
\sum_{k=1}^{n}\bigl(\hat p^{(m)}_k(t_i)-p_k(t_i)\bigr)^2 .
\]

We also track the predicted energy $E(\hat q^{(m)}(t), \hat p^{(m)}(t))$
throughout the rollout and compare it against the ground-truth energy
trajectory $E(q(t), p(t))$. As a scalar summary we report the relative
energy error
\[
\frac{\overline{\lvert \hat E - E\rvert}}{\overline{\lvert E\rvert}},
\qquad
\overline{(\cdot)} \;=\; \frac{1}{N_t}\sum_{i=1}^{N_t}(\cdot)(t_i),
\]
i.e.\ the time-averaged absolute energy deviation normalised by the
time-averaged absolute ground-truth energy. We further report trajectory-level statistics: for each evaluation $m$, the
pointwise SE is averaged over all prediction times to obtain the scalar MSE
$\mathcal{L}_m$, and $\{\mathcal{L}_m\}_{m=1}^{N_{\mathrm{obs}}}$ is summarised
across evaluations by its median and IQR. For an illustration alongside the aggregate statistics, we display one evaluation in full: the one initialised from the trajectory's first
observation, which yields the longest rollout horizon and therefore gives error and energy drift the greatest opportunity to accumulate.

\begin{figure}[!htbp]
\centering
\includegraphics[width=0.85\textwidth]{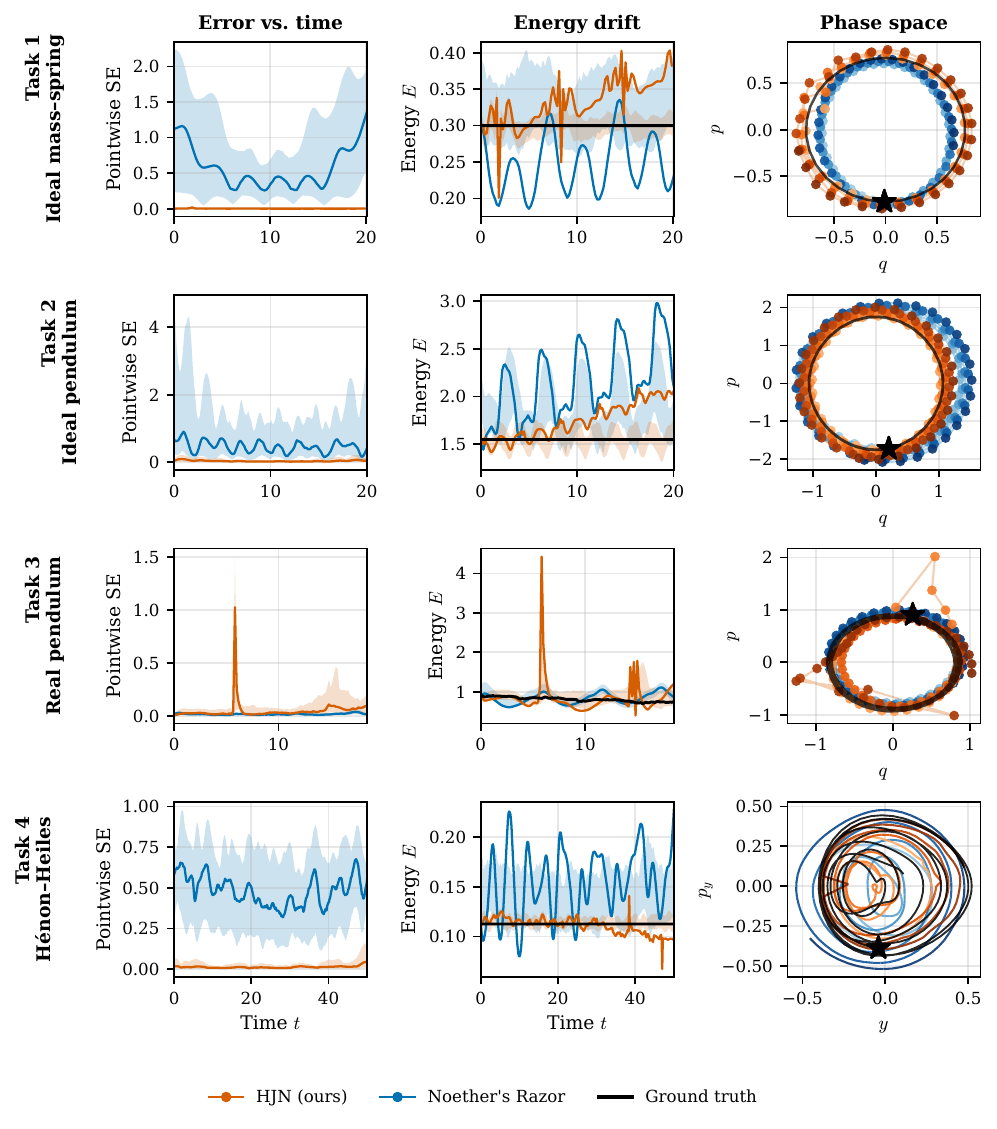}
\caption{Single-trajectory inference diagnostics for one representative
trajectory per task (Tasks~1--2 and 4: Test~A; Task~3: its held-out tail), HJN's seed-42 model (\textcolor[HTML]{D55E00}{orange}), Noether’s Razor (\textcolor[HTML]{0072B2}{blue}), ground truth (black). Inference is run from every observation point; the displayed energy and
phase-space curves use the run initialised from the first observation
(obs.\ index~$0$), the longest rollout and hence the most drift-prone.
Columns, left to right: (a)~pointwise squared error vs.\ time, with the median (solid)
and interquartile range (IQR; shaded) across observation choices; (b)~predicted energy
$E(\hat q,\hat p)$ for the obs-$0$ run (solid) with the across-observation
IQR band (shaded), against the ground-truth energy (black) -- conserved
on Tasks~1--2 and~4, decaying on Task~3; (c)~phase-space orbit for
the obs-$0$ run, colour-graded light$\to$dark with time, with the start
($t{=}0$, the conditioning observation) marked by a star. For Task~4 the panel shows the $(y,p_y)$ projection of the four-dimensional phase space. Test~B figures for Tasks~1--2 and~4 are shown in Appendix~\ref{app:supplementary_figures}.}

\label{fig:obs_sensitivity}
\end{figure}

\begin{table}[H]
\centering
\footnotesize
\caption{Aggregate prediction accuracy on the test sets: Brackets on the
$\mathrm{L}_2/\sqrt{E}$ column denote $95\%$ BCa bootstrap confidence
intervals (10{,}000 resamples). HJN is
reported per training seed. Rows Tn-A and Tn-B denote Task~$n$ under rollout
(A) and energy (B) extrapolation. Ref: the reference, included to give the errors a scale.}
\label{tab:main_results}
\setlength{\tabcolsep}{3pt}
\begin{tabular}{lcccc}
\toprule
& \multicolumn{4}{c}{$\mathrm{L}_2/\sqrt{E}$ (mean [95\% CI])} \\
\cmidrule(lr){2-5}
& \multicolumn{3}{c}{HJN (seed)} & \\
\cmidrule(lr){2-4}
Task & 42 & 52 & 62 & Ref \\
\midrule
T1-A & 0.083 [0.080, 0.087] & 0.087 [0.086, 0.088] & 0.061 [0.059, 0.065] & 1.620 [1.516, 1.706] \\
T1-B & 0.064 [0.056, 0.074] & 0.074 [0.062, 0.091] & 0.061 [0.054, 0.069] & 1.109 [1.071, 1.141] \\
T2-A & 0.162 [0.131, 0.195] & 0.226 [0.190, 0.269] & 0.165 [0.137, 0.195] & 0.618 [0.546, 0.718] \\
T2-B & 0.316 [0.248, 0.385] & 0.370 [0.294, 0.455] & 0.302 [0.241, 0.369] & 0.304 [0.259, 0.365] \\
T4-A & 0.508 [0.396, 0.648] & 0.557 [0.446, 0.688] & 0.752 [0.641, 0.894] & 1.874 [1.824, 1.926] \\
T4-B & 0.575 [0.450, 0.757] & 0.585 [0.467, 0.759] & 0.693 [0.562, 0.896] & 1.628 [1.513, 1.707] \\
\bottomrule
\end{tabular}
\end{table}

\begin{table}[H]
\centering
\footnotesize
\caption{Single-trajectory inference diagnostics on the representative
trajectory of each task and test set. Trajectory-level MSE is summarised
across observation choices by its median with IQR in brackets; energy
relative error $\overline{|\hat E - E|}/\overline{|E|}$ is reported for
the first-observation (longest-rollout) run. HJN is reported per training
seed. Rows Tn-A and Tn-B denote
Task~$n$ under rollout (A) and energy (B) extrapolation, T3 the
real-pendulum tail. Ref: the reference, included to give the errors a scale.}
\label{tab:obs_sensitivity}
\setlength{\tabcolsep}{2.5pt}
\begin{tabular}{lcccccccc}
\toprule
& \multicolumn{4}{c}{MSE (median [IQR])} & \multicolumn{4}{c}{Energy rel.\ err.\ (\%)} \\
\cmidrule(lr){2-5}\cmidrule(lr){6-9}
& \multicolumn{3}{c}{HJN (seed)} & & \multicolumn{3}{c}{HJN (seed)} & \\
\cmidrule(lr){2-4}\cmidrule(lr){6-8}
Task & 42 & 52 & 62 & Ref & 42 & 52 & 62 & Ref \\
\midrule
T1-A & 0.0028 [0.0021, 0.0041] & 0.0035 [0.0026, 0.0047] & 0.0023 [0.0018, 0.0030] & 0.706 [0.550, 0.922] & 12.7 & 5.8 & 5.2 & 18.9 \\
T1-B & 0.0065 [0.0044, 0.0089] & 0.0035 [0.0032, 0.0049] & 0.0039 [0.0030, 0.0049] & 1.870 [1.600, 2.419] & 2.8 & 2.5 & 2.8 & 16.6 \\
T2-A & 0.059 [0.038, 0.089] & 0.164 [0.111, 0.252] & 0.052 [0.033, 0.092] & 0.490 [0.290, 3.086] & 13.7 & 8.7 & 5.9 & 38.8 \\
T2-B & 0.442 [0.341, 0.542] & 0.626 [0.457, 1.020] & 0.454 [0.357, 0.536] & 0.407 [0.120, 0.540] & 4.7 & 8.0 & 7.0 & 9.9 \\
T4-A & 0.022 [0.013, 0.050] & 0.037 [0.023, 0.067] & 0.068 [0.042, 0.088] & 0.494 [0.409, 0.576] & 5.5 & 7.1 & 54.5 & 40.8 \\
T4-B & 0.064 [0.057, 0.129] & 0.057 [0.050, 0.126] & 0.076 [0.060, 0.181] & 0.479 [0.400, 0.663] & 4.9 & 15.0 & 7.6 & 29.3 \\
T3   & 0.098 [0.062, 0.194] & --- & --- & 0.017 [0.014, 0.021] & 23.7 & --- & --- & 16.5 \\
\bottomrule
\end{tabular}
\end{table}

\paragraph{Results.}
Under rollout extrapolation (Tests~A), at horizons of $3.1$ to $5.8\,T_{\max}$, the largest of HJN's three per-seed $\mathrm{L}_2/\sqrt{E}$ errors is $0.087$, $0.226$ and $0.752$ on Tasks~1, 2 and~4; under energy extrapolation (Tests~B) it is $0.074$, $0.370$ and $0.693$ (Table~\ref{tab:main_results}). Within each setting, HJN's $\mathrm{L}_2/\sqrt{E}$ error varies across seeds by at most a factor of $1.5$.  Composition of segments is not error-free: HJN's error rises toward both ends of the observation-time axis (Fig.~\ref{fig:summary_statistics}b), where rollouts are longest, most visibly on Tasks~2 and~4. On Task~2, HJN's error rises from Test~A to Test~B for every seed ($0.162\!\to\!0.316$, $0.226\!\to\!0.370$, $0.165\!\to\!0.302$), as it does on Task~4 for two of the three ($0.508\!\to\!0.575$, $0.557\!\to\!0.585$), although the Test~B horizons are shorter. On Task~2 this is consistent with HJN's Test~B error arising from the change of energy band rather than from accumulation across handoffs. The reference's error, in contrast, falls from Test~A to Test~B on all three tasks ($1.620\!\to\!1.109$, $0.618\!\to\!0.304$, $1.874\!\to\!1.628$). For every seed, HJN's mean error is below the reference's in five of the six settings; on Task~2 Test~B the per-seed means ($0.302$ to $0.370$) bracket the reference's ($0.304$). Similar results are present for a model trained without $\mathcal{L}_{\mathrm{HSIC}}$
(Table~\ref{tab:nohsic_prediction}, Appendix~\ref{zero_hsic_experiments}).

On the six extrapolation settings, in the first-observation run, HJN's relative energy error lies between $2.5\%$ and $15.0\%$ in seventeen of the eighteen (setting, seed) pairs and is below the reference's in each of them; the exception is seed~62 on Task~4 Test~A, at $54.5\%$ against the reference's $40.8\%$ (Table~\ref{tab:obs_sensitivity}). HJN's relative energy error is smaller on Test~B than on Test~A for every seed on Task~1, and for two of the three seeds on Tasks~2 and~4. Test~B combines the shorter horizon with the
higher energy band, and the latter enlarges the normaliser $\overline{|E|}$, so
these differences cannot be attributed to horizon alone. On the representative trajectories, trajectory-level MSE is consistent with the aggregate accuracy findings of Table~\ref{tab:main_results}: for every seed, HJN's IQR lies below the reference's on five of the six extrapolation settings and overlaps it on Task~2 Test~B, where HJN's median is the higher of the two. In the displayed first-observation runs of Tasks~1, 2 and~4 (Fig.~\ref{fig:obs_sensitivity}c), HJN's predicted orbit stays near the ground-truth orbit, with an outward drift on Tasks~1 and~2 that grows over the rollout, matching the rising predicted energy. The displayed seed-42 model has the largest energy error of the three seeds on Test~A of Tasks~1 and~2 and is the best on Task~4 Test~A in both MSE and energy error (Table~\ref{tab:obs_sensitivity}). The displayed model's dependence on the observation choice is visible in the across-observation IQR bands (Fig.~\ref{fig:obs_sensitivity}a); in Table~\ref{tab:obs_sensitivity}, the width of its trajectory-level IQR is below its median MSE on both test sets of Tasks~1 and~2 and above it on both test sets of Task~4; the single-trajectory diagnostics thus resolve in time the behaviour across conditioning points already observed in aggregate. We take from the comparison only that, on the six extrapolation settings, HJN's prediction error is comparable to or below that of a published conserved-quantity learner under the same evaluation protocol.

Task~3 probes how HJN behaves when the premise of Section~\ref{subsec:problem_formulation} fails. The measured pendulum is dissipative, so its flow is not volume-preserving and therefore not
canonical. A canonical family $\mathcal{M}(\cdot,t)$ exactly constant along segments would make
the flow itself canonical (Appendix~\ref{app:prop_flow}), so no canonical solution of
\eqref{eq:flow_form} exists here. The reference cannot reproduce the decay either, since it predicts by integrating a learned
Hamiltonian vector field, yet it is the more accurate of the two on this task, at MSE $0.017$
against $0.098$ and relative energy error $16.5\%$ against $23.7\%$. The figure shows
the shape of that failure (Fig.~\ref{fig:obs_sensitivity}, Task~3 row): the
degradation is not uniform but concentrated in localised excursions. At them the predicted energy departs markedly from the true value and the orbit leaves the
phase-space band traced by the ground truth. HJN's error also depends more on the conditioning observation here: its across-observation interquartile range is the wider of the two, $[0.062,0.194]$ against $[0.014,0.021]$.

\section{Conclusion and Limitations}
\label{sec:conclusion}

Hamilton--Jacobi theory recasts learning Hamiltonian dynamics as representation
learning: it seeks a canonical transformation that carries every state of a trajectory to the same conserved
quantities $(\alpha,\beta)$, fixed by the initial conditions, and whose inverse returns
the state at any requested time. For autonomous systems with complete flow, such a
transformation exists without any assumption of integrability, and every such transformation takes a common form~\eqref{eq:flow_form}, characterised without a generating function: a single globally defined map, the
object a neural network can parameterise. HJN parameterises it and its exact algebraic inverse as a symplectic-by-construction encoder--decoder pair, trained to reconstruct each trajectory segment from a single observation of it. Its decoder predicts the states at any requested times within the segment from one encoding, in a single forward pass and without a numerical integrator. On the states and times the training data covers, an equivalence holds: HJN achieves zero prediction loss precisely when its encoder takes the form~\eqref{eq:flow_form}. At any loss and on any segment of any trajectory, the departure of the latents from constancy and the prediction error on that segment bound each other, up to local constants. Prediction and the discovery of conserved quantities thus become one objective, while the form those quantities take is
never prescribed.

Two experiments measure, by complementary means, how closely trained models, at the nonzero loss training reaches, carry the structure that zero prediction loss forces exactly on the training data, both there and on states not seen during training. We trained three models from different seeds on each of an ideal mass--spring oscillator, an ideal pendulum and the non-integrable H\'enon--Heiles system. We show that the prediction objective determines the latents only up to a time-independent canonical transformation, which we call their frame. Measured in the frame-free setting, the latents come close to constancy on held-out trajectories at the training energies: for the median trajectory they deviate from their segments' frame-free conserved quantities by 2--14\% of the observed motion in rms, a fraction at most $1.35$ times that on the training trajectories, which the equivalence covers. In absolute terms, this deviation is also below the rms length of one training-noise vector for every model. In the frame that training settles on, the latents are interpretable in an operational sense: on unseen states at the training energies and in every model, a polynomial of the latents of degree at most four approximates the energy (cross-validated $R^2\ge0.944$), and degree-one polynomials approximate the circle coordinates of the initial phase in the two integrable systems ($R^2>0.99$). These read-outs, fit at $t=0$ alone, respond
to shifts of the encoding time approximately as the true quantities do: the per-state slope of the recovered energy has an interquartile range containing zero in every model, with median relative errors of at most $5.2\%$. The energy figures, like the constancy figures above, include the non-integrable H\'enon--Heiles system, consistent with~\eqref{eq:flow_form} carrying no integrability requirement. In the two integrable systems, the recovered phase advances approximately at each system's own rate. For the oscillator, the median slope lies within $0.2\%$ of the true rate. For the pendulum, whose rate falls with energy, the median deviation from each state's own rate is about $1.3\%$ of it; the recovered rates follow that fall in direction but flatten it to roughly $10$--$22\%$ of its extent. The equivalence, the bounds and these experiments all concern single segments. Over rollouts that compose segments from a single observation, HJN's error, on every in-scope test set, is comparable to or below that of a reference method that also learns conserved quantities, under the same protocol.

\paragraph{Limitations.}
The theory is what bounds the method's scope. Dissipative and driven systems lie outside it, as do regions from which trajectories
escape to infinity or reach a singularity of $H$ in finite time. It also requires inputs in the form of conjugate pairs of generalised
coordinates and momenta. This is restrictive in practice, as real-world
measurements typically provide positions and velocities, or raw sensor data,
rather than momenta. We evaluate on one system outside this scope, the measured dissipative pendulum, for which~\eqref{eq:flow_form} admits
no canonical solution: neither HJN nor the reference can reproduce the
decay, yet HJN is the less accurate of the two. Within the scope, the experiments
cover only systems with at most two degrees of freedom, and on every one of them
constancy weakens outside the training energies; for the ideal pendulum and the H\'enon--Heiles system, the latents of the median trajectory deviate there from their segments' initial states by more than the rms length of one training-noise vector. Finally, although the theory needs no segmentation, in practice it matters: fitting grows harder as segments lengthen, while shorter segments mean more re-encodings in a rollout. On H\'enon--Heiles, lengthening segments from the reported length $T_{\max}$ to whole trajectories raised the prediction error of Table~\ref{tab:main_results} on both test sets 1.9- to 3.4-fold, beyond the spread across seeds, while quartering the segment length lowered it by less than that spread and multiplied the network evaluations three- to fourfold (Appendix~\ref{sec:tmax_sweep}). The method
therefore depends on the segment length, which this work sets to a single value
for all of phase space by a rule we do not claim optimal.

\paragraph{Future work.}
Segmentation is the most immediate place to improve: rules for setting the segment length warrant systematic comparison, from single values for all of phase space, such as the one used in this work, to lengths that vary across it, such as those sketched in Appendix~\ref{app:adaptive}. To widen the scope, a learned map from raw observations into a conjugate pair, placed upstream of the encoder, could relax the input requirement, provided it supplies the phase-space structure the construction assumes. Finally, the prediction objective leaves the
frame of the latents free, yet every model, including those trained without $\mathcal{L}_{\mathrm{HSIC}}$, settles on one from which the classical conserved quantities are recovered by polynomial read-outs of degree at most four, pointing to a systematic inductive bias whose source in the model or training setup remains to be isolated.

\section*{Acknowledgements}
Funded by the European Union under the Horizon Europe research and innovation programme (Grant Agreement No.~101298363, Project OPERA). Views and opinions expressed are however those of the author(s) only and do not necessarily reflect those of the European Union or European Health and Digital Executive Agency (HaDEA). Neither the European Union nor the granting authority can be held responsible for them.

\appendix
\section{Proofs}
\label{sec:sm_proof}

\subsection{Canonical transformations to conserved quantities}
\label{app:prop_flow}

We prove the characterisation stated in
Section~\ref{subsec:problem_formulation}, and record one consequence used in
Section~\ref{sec:training}.

\paragraph{Setting.} Let $H$ be autonomous and $C^2$, with complete flow $\phi$, so that
$\phi(z,t)$ is defined for all $z$ and all $t\in\mathbb{R}$ and satisfies the
one-parameter group property $\phi(\phi(z,t),s)=\phi(z,t+s)$. 

Let $\mathcal{U}\subseteq\mathbb{R}^{2n}$ be a nonempty open set invariant under the
flow in both time directions, $\phi(\mathcal{U},t)=\mathcal{U}$ for all
$t\in\mathbb{R}$; since $H$ is conserved, any open sublevel set
$\{z:H(z)<E\}$ has this property.

Let $\Psi(\cdot,t)$ be a canonical transformation of $\mathcal{U}$ for each $t$,
by which we mean a diffeomorphism of $\mathcal{U}$ onto its image whose Jacobian
is symplectic at every point, continuously differentiable in all its arguments. Write $X_H$ for the Hamiltonian vector field,
$\partial\Psi/\partial t$ for the derivative with respect to $t$ at a fixed
phase-space point, and $D_z\Psi$ for the Jacobian with respect to the phase-space
variables at fixed $t$.

\paragraph{Claim.} The following are equivalent:
(i) $\Psi(z(t),t)$ is constant in $t$ along every trajectory $z(\cdot)$ in
$\mathcal{U}$, that is, the transformed variables are conserved;
(ii) $\partial\Psi/\partial t+(D_z\Psi)X_H=0$ at every point of $\mathcal{U}$ and every $t\in\mathbb{R}$;
(iii) $\Psi(z,t)=\Psi_0(\phi(z,-t))$ for all $z\in\mathcal{U}$ and all $t$, where
$\Psi_0(\cdot)=\Psi(\cdot,0)$.
Here $\Psi_0=\Psi(\cdot,0)$ is canonical by hypothesis. Conversely, for every
canonical transformation $\Psi_0$ of $\mathcal{U}$, the family
$\Psi(z,t):=\Psi_0\bigl(\phi(z,-t)\bigr)$ is a canonical transformation of $\mathcal{U}$ for each $t$ and satisfies (i)--(iii). The transformations
characterised above are therefore exactly those obtained in this way from a
canonical $\Psi_0$. In particular, since the identity is canonical, such
transformations exist on every such $\mathcal{U}$.

\paragraph{Proof.}
\emph{(i)$\Leftrightarrow$(ii).} By the chain rule,
$\tfrac{\mathrm{d}}{\mathrm{d}t}\Psi(z(t),t)
=\tfrac{\partial\Psi}{\partial t}(z(t),t)+D_z\Psi(z(t),t)\,\dot z(t)$,
and $\dot z=X_H(z)$ along a trajectory. The derivative vanishes along every
trajectory precisely when (ii) holds for every $(z,t)\in\mathcal{U}\times\mathbb{R}$, since, the system being autonomous, a trajectory passes through every point of
$\mathcal{U}$ at every time.

\emph{(ii)$\Rightarrow$(iii).} Fix $z\in\mathcal{U}$ and set
$w(t)=\Psi(\phi(z,t),t)$, which is defined for all $t$ because the flow is
complete and $\phi(z,t)$ remains in $\mathcal{U}$. Since
$\tfrac{\mathrm{d}}{\mathrm{d}t}\phi(z,t)=X_H(\phi(z,t))$, the chain rule gives
$\dot w(t)=\bigl[\tfrac{\partial\Psi}{\partial t}+(D_z\Psi)X_H\bigr](\phi(z,t),t)$,
which vanishes by (ii). Hence $w(t)=w(0)=\Psi(z,0)=\Psi_0(z)$, that is
$\Psi(\phi(z,t),t)=\Psi_0(z)$ for every $z\in\mathcal{U}$ and every $t$.
Replacing $z$ by $\phi(z,-t)$, which again lies in $\mathcal{U}$, and using the
group property $\phi(\phi(z,-t),t)=\phi(z,0)=z$, gives
$\Psi(z,t)=\Psi_0(\phi(z,-t))$.

\emph{(iii)$\Rightarrow$(i).} Along a trajectory, $z(t)=\phi(z_0,t)$, so
$\Psi(z(t),t)=\Psi_0(\phi(\phi(z_0,t),-t))=\Psi_0(z_0)$, independent of $t$.

\emph{Canonicity of the family.} Let $\Psi_0$ be any canonical transformation of
$\mathcal{U}$ and set $\Psi(z,t):=\Psi_0\bigl(\phi(z,-t)\bigr)$. For each $t$
this is the composition of the diffeomorphism $\phi(\cdot,-t)$ of $\mathcal{U}$
with $\Psi_0$, hence differentiable and invertible, and the chain rule gives
\[
D_z\Psi(z,t)=D\Psi_0\bigl(\phi(z,-t)\bigr)\,D_z\phi(z,-t).
\]
The Hamiltonian phase flow is a canonical
transformation\citep[\S38]{arnold1989mathematical}\citep[\S9.6]{goldstein2002classical},
so $D_z\phi(z,-t)$ is symplectic; $D\Psi_0$ is symplectic because $\Psi_0$ is
canonical. Symplectic matrices form a group under
multiplication\citep[\S9.4]{goldstein2002classical}, so $D_z\Psi(z,t)$ is
symplectic and $\Psi(\cdot,t)$ is canonical for each $t$. By construction
$\Psi$ satisfies (iii), hence (i) and (ii). Since in the forward direction
$\Psi_0=\Psi(\cdot,0)$ is canonical by hypothesis, the transformations
characterised in the Claim are exactly those obtained from a canonical
$\Psi_0$. $\square$

\paragraph{Remark: the transformed Hamiltonian.} For any $\Psi$ satisfying
(i)--(iii), the choice $K=0$ furnishes a transformed Hamiltonian for the given
$H$: by (i) the new variables satisfy $\dot Q_i=0=\partial K/\partial P_i$ and
$\dot P_i=0=-\partial K/\partial Q_i$, so Hamilton's equations hold in the new
variables. This exhibits the $K=0$ property invoked in
Section~\ref{subsec:problem_formulation}.

\paragraph{Independence of the frame.} Let $\Psi$ satisfy (iii).
Its image is
$\Psi(\mathcal{U},t)=\Psi_0(\phi(\mathcal{U},-t))=\Psi_0(\mathcal{U})$, independent
of $t$, and its inverse is
$\Psi^{-1}(\cdot,t)=\phi(\Psi_0^{-1}(\cdot),t)$, since applying it to
$\Psi(z,t)=\Psi_0(\phi(z,-t))$ returns $\phi(\phi(z,-t),t)=z$. Encoding at time
$t$ and decoding at time $t'$ therefore gives
\[
\Psi^{-1}\bigl(\Psi(z,t),\,t'\bigr)
=\phi\bigl(\Psi_0^{-1}(\Psi_0(\phi(z,-t))),\,t'\bigr)
=\phi\bigl(\phi(z,-t),\,t'\bigr)
=\phi(z,\,t'-t).
\]
The result does not involve $\Psi_0$: predictions, and hence the prediction
loss, are the same for every choice of frame. The frame is in this sense a
gauge degree of freedom; what a prediction constrains is $\phi$, not $\Psi_0$.

\subsection{Frame invariance of the prediction loss}
\label{app:frame_invariance}

Let $S:\mathbb{R}^{2n}\to\mathbb{R}^{2n}$ be a bijection of the latent variables
that does not depend on $t$, and let $\mathcal{M}'=S\circ\mathcal{M}$. The decoder is the exact algebraic inverse of the encoder and shares its
parameters (Appendix~\ref{app:invertible}), so composing $S$ onto the encoder
composes $S^{-1}$ onto the decoder $\mathcal{M}'^{-1}=\mathcal{M}^{-1}\circ S^{-1}$ and, for every
$z=(q,p)$ and every pair of times $t,t_{\mathrm{pred}}$,
\begin{equation}
\begin{split}
    \mathcal{M}'^{-1}\bigl(\mathcal{M}'(z,t),\,t_{\mathrm{pred}}\bigr)
    &=\mathcal{M}^{-1}\Bigl(S^{-1}\bigl(S(\mathcal{M}(z,t))\bigr),\,
      t_{\mathrm{pred}}\Bigr)\\
    &=\mathcal{M}^{-1}\bigl(\mathcal{M}(z,t),\,t_{\mathrm{pred}}\bigr),
    \end{split}
    \label{eq:frame_cancel}
\end{equation}
since $S^{-1}\circ S=\mathrm{id}$. If in addition $S$ is canonical, then
$\mathcal{M}'$ is canonical following from the group property of canonical
transformations. Consequently $\mathcal{L}_{\mathrm{pred}}(\mathcal{M}')
=\mathcal{L}_{\mathrm{pred}}(\mathcal{M})$ identically.

The
cancellation in \eqref{eq:frame_cancel} is algebraic, following from the decoder
being the exact inverse of the encoder, so it holds at every parameter setting
visited during training. Two encoders differing by a
$t$-independent $S$ therefore receive the same prediction loss throughout
training. At zero prediction loss, where \eqref{eq:flow_form} holds, such an $S$ is a
change of the frame $\Psi_0$ at fixed $\phi$, consistent with the gauge
freedom of Appendix~\ref{app:prop_flow}.

\subsection{Canonicity Proofs}
We prove that the mapping and inverse networks are both canonical transformations. For the mapping network we do this by proving that the two sub-steps every Hamilton Jacobi layer consists of, both
satisfy the symplectic condition, and then by proving that a composition of symplectic maps itself satisfies the symplectic condition. For the inverse network we first show that the Jacobian of any canonical transformation is invertible, then construct the inverse of each layer in closed form, and finally show that the inverse is itself canonical. The symplectic condition is: $M^T \Omega\, M = \Omega$.

Here $M$ is the $2n \times 2n$ Jacobian of the transformation
and $\Omega$ is the $2n \times 2n$ standard symplectic matrix
\begin{equation}
  \Omega =
  \begin{pmatrix} 0 & I_n \\ -I_n & 0 \end{pmatrix}\,,
  \label{eq:sm_omega}
\end{equation}
with $I_n$ the $n \times n$ identity matrix. We denote the
Hessian matrix of a scalar function $\varphi$ with respect to
variables $\mathbf{u}$ by $\mathcal{H}_\varphi$, i.e.,
$(\mathcal{H}_\varphi)_{ij} =
\partial^2 \varphi / \partial u_i \partial u_j$.

\subsubsection{Step~1}

The transformation at each degree of freedom
$i = 1, \ldots, n$ is
\begin{align}
  Q_i^{(1)} &= a\, q_i + c_{1,i}\,,
  \label{eq:sm_s1q}\\
  P_i^{(1)} &= \tfrac{1}{a}\, p_i
    - \gamma\,\frac{\partial G(\mathbf{q},t)}{\partial q_i}
    + c_{2,i}\,,
  \label{eq:sm_s1p}
\end{align}
where $a \neq 0$ and $G(\mathbf{q},t)$ is a scalar function of
all $n$ generalised coordinates and time.
The additive constants $c_{1,i}$ and $c_{2,i}$ do not appear in
the Jacobian and therefore play no role in the symplectic
condition.
The $2n \times 2n$
Jacobian has the block form
\begin{equation}
  M_1 =
  \begin{pmatrix}
    a\, I_n & 0 \\[4pt]
    -\gamma\, \mathcal{H}_G & \tfrac{1}{a}\, I_n
  \end{pmatrix}\,,
  \label{eq:sm_M1}
\end{equation}
where $\mathcal{H}_G$ is the Hessian of $G$ with respect to
$\mathbf{q}$. Direct computation gives
\begin{equation}
  M_1^T \Omega\, M_1 =
  \begin{pmatrix}
    \gamma a\,(\mathcal{H}_G^T - \mathcal{H}_G)
    & I_n \\[4pt]
    -I_n & 0
  \end{pmatrix}\,.
  \label{eq:sm_M1check}
\end{equation}
This equals $\Omega$ if and only if
$\mathcal{H}_G = \mathcal{H}_G^T$, i.e., the Hessian of $G$ is
symmetric. By Schwarz's theorem, this holds whenever the
second-order partial derivatives of $G$ exist and are continuous.
Since $G$ is parameterised by an MLP with smooth activations
(e.g., GELU, tanh, SiLU)---and compositions and sums of $C^2$
functions are $C^2$---this condition is satisfied
everywhere, and Step~1 is a canonical transformation.

\subsubsection{Step~2}

The transformation at each degree of freedom
$i = 1, \ldots, n$ is
\begin{align}
  P_i^{(2)} &= a'\, P_i^{(1)} + c_{1,i}'\,,
  \label{eq:sm_s2p}\\
  Q_i^{(2)} &= \tfrac{1}{a'}\, Q_i^{(1)}
    + \gamma'\,\frac{\partial F(\mathbf{P}^{(1)}\!,t)}
                     {\partial P_i^{(1)}}
    + c_{2,i}'\,,
  \label{eq:sm_s2q}
\end{align}
where $a' \neq 0$ and $F(\mathbf{P}^{(1)}\!,t)$ is a scalar
function of all $n$ transformed momenta and time.
As before, the additive constants $c_{1,i}'$ and $c_{2,i}'$
drop out of the Jacobian.
The $2n \times 2n$ Jacobian has the block form
\begin{equation}
  M_2 =
  \begin{pmatrix}
    \tfrac{1}{a'}\, I_n
    & \gamma'\, \mathcal{H}_F \\[4pt]
    0 & a'\, I_n
  \end{pmatrix}\,,
  \label{eq:sm_M2}
\end{equation}
where $\mathcal{H}_F$ is the Hessian of $F$ with respect to
$\mathbf{P}^{(1)}$. By an analogous computation,
\begin{equation}
  M_2^T \Omega\, M_2 =
  \begin{pmatrix}
    0 & I_n \\[4pt]
    -I_n &
    a'\gamma'\,(\mathcal{H}_F^T - \mathcal{H}_F)
  \end{pmatrix}\,,
  \label{eq:sm_M2check}
\end{equation}
which equals $\Omega$ whenever
$\mathcal{H}_F = \mathcal{H}_F^T$, again guaranteed by
Schwarz's theorem for smooth MLP activations.

\subsubsection{Composition (HJL-Mapping Network)}
\label{app:composition}
We show by induction that the composition of any finite number of
symplectic maps is itself symplectic. Let $A_1,\ldots,A_K$ be symplectic
matrices and write $A_{(k)} = A_k \cdots A_1$.

\paragraph{Base case.}
For a single map the statement holds by hypothesis:
$A_1^T\Omega\,A_1 = \Omega$.

\paragraph{Inductive step.}
Assume $A_{(k-1)}^T\,\Omega\,A_{(k-1)} = \Omega$, and let $A_k$ be any
further symplectic matrix, i.e.\ $A_k^T\,\Omega\,A_k = \Omega$. Then for
$A_{(k)} = A_k\,A_{(k-1)}$:
\begin{equation}
  A_{(k)}^T\,\Omega\,A_{(k)}
  = A_{(k-1)}^T\,A_k^T\,\Omega\,A_k\,A_{(k-1)}
  = A_{(k-1)}^T\,\Omega\,A_{(k-1)}
  = \Omega\,.
  \label{eq:sm_comp}
\end{equation}
Hence, by induction, $A_{(K)} = A_K\cdots A_1$ is symplectic for
every $K\geq 1$.
Thus the Hamilton--Jacobi Layer, whose Jacobian is $M_2 M_1$ by the chain rule
(with $M_2$ evaluated at the image point $(\mathbf{Q}^{(1)},\mathbf{P}^{(1)})$,
where the symplectic condition of Step~2 holds, as it does pointwise),
and hence the Mapping Network
$\mathcal{M} = \mathrm{HJL}_L \circ \cdots \circ \mathrm{HJL}_1$, both satisfy
the symplectic condition as compositions of symplectic maps.

\subsubsection{Determinant and invertibility}
\label{app:invertible}

\paragraph{The Mapping Network has unit determinant.}
The determinant of each sub-step follows directly from the block-triangular
structure of its Jacobian, without reference to the symplectic condition. For
block matrices with square diagonal blocks,
$\det\!\begin{pmatrix} A & 0 \\ C & D\end{pmatrix}
=\det\!\begin{pmatrix} A & B \\ 0 & D\end{pmatrix}=\det(A)\det(D)$, so
\eqref{eq:sm_M1} and \eqref{eq:sm_M2} give
\begin{equation}
\begin{gathered}
  \det M_1 = \det(a\,I_n)\,\det\!\left(\tfrac{1}{a}\,I_n\right)
           = a^{n}a^{-n} = 1,
  \\
  \det M_2 = \det\!\left(\tfrac{1}{a'}\,I_n\right)\det(a'\,I_n)
           = (a')^{-n}(a')^{n} = 1 .
\end{gathered}
  \label{eq:sm_detsteps}
\end{equation}
The reciprocal scaling of each conjugate pair cancels exactly, and the shear
blocks $-\gamma\,\mathcal{H}_G$ and $\gamma'\,\mathcal{H}_F$ occupy the
off-diagonal position and therefore never enter. The result holds for any
$\gamma,\gamma'$, any Hessians, any $n$, and either sign of $a,a'$; it requires
only $a,a'\neq 0$. By the chain rule and
multiplicativity of the determinant, each Hamilton--Jacobi Layer has
$\det(M_2 M_1)=1$, and hence
\begin{equation}
  \det \mathcal{M}
  = \prod_{k=1}^{L}\det\!\left(M_2^{(k)}M_1^{(k)}\right) = 1 ,
  \qquad
  \det \mathcal{M}^{-1} = \bigl(\det\mathcal{M}\bigr)^{-1} = 1 .
  \label{eq:sm_detM}
\end{equation}
The encoder--decoder pair is therefore exactly volume- and
orientation-preserving, at every point of phase space and for every parameter
setting ($a,a'\neq 0$).

\paragraph{Jacobian invertibility of a general canonical transformation.}
Taking determinants of both sides of the symplectic condition constrains
$\det M$, once $\det(\Omega)$ is known. Swapping row $i$ with row $n+i$ for each $i = 1,\ldots,n$ requires
$n$ transpositions and turns $\Omega$ into the block-diagonal matrix
$\mathrm{diag}(-I_n,\,I_n)$, so
\begin{equation}
  \det(\Omega) = (-1)^{n}\,\det(-I_n)\,\det(I_n)
               = (-1)^{n}(-1)^{n} = 1 \neq 0 .
  \label{eq:sm_detomega}
\end{equation}
Taking determinants of $M^T \Omega\, M = \Omega$ and using multiplicativity
together with $\det(M^T) = \det(M)$ gives
$(\det M)^2\det(\Omega) = \det(\Omega)$, and since $\det(\Omega) \neq 0$ we may
divide to obtain
\begin{equation}
  (\det M)^2 = 1 , \qquad \det M = \pm 1 .
  \label{eq:sm_detpm}
\end{equation}
The Jacobian of any canonical transformation is therefore
invertible, so a $C^1$ canonical map is locally invertible; global invertibility of our layers is established below by explicit construction. This argument fixes the determinant only up to sign, which is all that invertibility requires; for the maps realised
by our architecture the sign is settled by
\eqref{eq:sm_detsteps}--\eqref{eq:sm_detM} above.

In our architecture the Inverse Network
$\mathcal{M}^{-1} = \mathrm{IHJL}_1 \circ \cdots \circ \mathrm{IHJL}_L$
requires that all parameters are shared between $\mathrm{HJL}_k$ and
$\mathrm{IHJL}_k$ for each $k = 1, \ldots, L$.

Within each inverse layer, the triangular structure of the sub-steps dictates
the recovery order. Step~2 is inverted by recovering the momentum first,
\begin{equation}
  P_i^{(1)} = \frac{P_i^{(2)} - c'_{1,i}}{a'},
  \qquad
  Q_i^{(1)} = a'\!\left(Q_i^{(2)} - c'_{2,i}
      - \gamma'\,\frac{\partial F(\mathbf{P}^{(1)}\!,t)}{\partial P_i^{(1)}}\right),
  \label{eq:sm_inv_s2}
\end{equation}
since $\mathbf{P}^{(1)}$ must be known before $\partial F/\partial P_i^{(1)}$ can
be evaluated. Step~1 is inverted in the mirrored order, recovering the coordinate
first,
\begin{equation}
  q_i = \frac{Q_i^{(1)} - c_{1,i}}{a},
  \qquad
  p_i = a\!\left(P_i^{(1)} - c_{2,i}
      + \gamma\,\frac{\partial G(\mathbf{q},t)}{\partial q_i}\right).
  \label{eq:sm_inv_s1}
\end{equation}
Both are defined on all of $\mathbb{R}^{2n}$ whenever $a,a' \neq 0$, since the gradients of the MLPs $G$ and $F$ exist everywhere. Each Hamilton--Jacobi Layer is therefore a global bijection of $\mathbb{R}^{2n}$, as
is their composition.

\subsubsection{Inverse}
\label{app:inverse}
Let $M$ be the Jacobian of a canonical transformation, so that
$M^T\Omega\,M = \Omega$. By the previous subsection $M$ is invertible and the map is $C^1$ (its Jacobian consists of constants and Hessians of the $C^2$ MLPs $G$, $F$), so by
the inverse function theorem the inverse map, which exists globally by construction, is $C^1$ with
Jacobian $M^{-1}$ at the image point.

Right-multiplying the symplectic condition by $M^{-1}$ gives
$M^T\Omega = \Omega\,M^{-1}$, and left-multiplying by
$(M^T)^{-1} = (M^{-1})^T$ yields
\begin{equation}
  (M^{-1})^T\,\Omega\,M^{-1} = \Omega\,.
  \label{eq:sm_inv}
\end{equation}
Hence the inverse of a canonical transformation is itself canonical, and in
particular the Inverse Network $\mathcal{M}^{-1}$ is a canonical
transformation, being the inverse of the (proven canonical) Mapping Network.

\section{Generator function regularities}
\label{app:generator_regularities}
The local non-degeneracy condition of $\det\!\left(\frac{\partial^2 S}{\partial q_i \,\partial \alpha_j}\right)\neq 0$ for the type-2 generating function mentioned above is a regularity condition on the chosen generating-function representation rather than on the underlying dynamics. Different generating-function types correspond to different local projections of a
canonical transformation: each takes a different set of $n$ old and $n$ transformed
variables as independent, and exists only where that choice is locally invertible. Differentiating $p_i=\partial S/\partial q_i(q,P,t)$ with respect to $p_j$ at fixed $q$ gives
$\sum_k \frac{\partial^2 S}{\partial q_i\,\partial P_k}\frac{\partial P_k}{\partial p_j}=\delta_{ij}$,
so the mixed Hessian is the inverse of the block $\partial P/\partial p$ whose
invertibility the type-2 choice requires. Wherever a type-2 representation of a
canonical transformation exists, the non-degeneracy condition therefore holds
automatically, and the failure of the representation appears as divergence of the
Hessian determinant rather than as its vanishing.

The condition therefore differs from type to type, and a particular representation
may become singular even where the underlying canonical transformation remains
regular and another type provides a regular representation. This occurs already for the harmonic oscillator, where the type-1 and type-2 representations of
\eqref{eq:flow_form} cease to exist at alternating quarter periods and never at the
same time, although the transformation is regular throughout
(Appendix~\ref{app:gf_degeneracy}). Moreover, each standard generating-function type fixes a particular choice of  $n$ old and $n$ transformed variables, so a model parameterised by a single generating function of fixed type commits to that choice in its architecture. Consequently, committing to one fixed generating-function type globally may unnecessarily restrict the domain over which that representation is valid. 

\subsection{A generating-function degeneracy example}
\label{app:gf_degeneracy}
Take the one-dimensional harmonic oscillator $H=p^{2}/2m+\tfrac12 m\omega^{2}q^{2}$, and the transformation given by \eqref{eq:flow_form}
with $\Psi_0=\mathrm{id}$, that is $\Psi(\cdot,t)=\phi(\cdot,-t)$:
\begin{equation}
\begin{pmatrix} Q\\ P\end{pmatrix}
=
\begin{pmatrix}
\cos\omega t & -\dfrac{\sin\omega t}{m\omega}\\[2mm]
m\omega\sin\omega t & \cos\omega t
\end{pmatrix}
\begin{pmatrix} q\\ p\end{pmatrix}.
\label{eq:ho_map}
\end{equation}
It is canonical and $(Q,P)$ are conserved, by Appendix~\ref{app:prop_flow}; its
Jacobian is the matrix in \eqref{eq:ho_map}, of determinant $1$ at every $(q,p,t)$.

A type-2 representation $S(q,P,t)$ takes $(q,P)$ as independent variables, so it
exists only where $(q,p)\mapsto(q,P)$ is locally invertible, that is where
$\partial P/\partial p=\cos\omega t\neq0$. There,
\[
S(q,P,t)=qP\sec\omega t-\tfrac12 m\omega q^{2}\tan\omega t
-\frac{P^{2}}{2m\omega}\tan\omega t,
\qquad
\frac{\partial^{2}S}{\partial q\,\partial P}=\sec\omega t,
\]
the additive function of $t$, which is undetermined by the transformation, having been chosen so that the transformed Hamiltonian
vanishes identically. Direct differentiation confirms $p=\partial S/\partial q$,
$Q=\partial S/\partial P$ and $H+\partial S/\partial t=0$. The mixed derivative is the reciprocal of
$\partial P/\partial p$, so it diverges, and the coefficients of $S$ with it, as
$\omega t\to\pi/2+k\pi$: the representation does not merely become degenerate
there, it ceases to exist.

A type-1 representation $F_1(q,Q,t)$ takes $(q,Q)$ instead, and exists where
$\partial Q/\partial p=-\sin\omega t/m\omega\neq0$:
\[
F_1(q,Q,t)=\frac{m\omega}{2}\bigl(q^{2}+Q^{2}\bigr)\cot\omega t
-m\omega qQ\csc\omega t,
\qquad
\frac{\partial^{2}F_1}{\partial q\,\partial Q}=-m\omega\csc\omega t,
\]
failing at $\omega t=k\pi$; it likewise satisfies $p=\partial F_1/\partial q$,
$P=-\partial F_1/\partial Q$, and $H+\partial F_1/\partial t=0$.

The remaining two types add nothing: $F_3(p,Q,t)$ exists where
$\partial Q/\partial q=\cos\omega t\neq0$ and $F_4(p,P,t)$ where
$\partial P/\partial q=m\omega\sin\omega t\neq0$, so they fail exactly where the
type-2 and type-1 representations do. The four types therefore fail in two pairs,
on disjoint sets of times, and no single type is valid for all $t$, while
\eqref{eq:ho_map} is regular throughout: at $\omega t=\pi/2$ it reads $Q=-p/m\omega$, $P=m\omega q$, with unit
Jacobian determinant. The degeneracy belongs to the chosen representation, not to
the transformation or to the dynamics. Nor is the failure an artefact of the choice $\Psi_0=\mathrm{id}$, among linear frames. For any
linear symplectic frame $\Psi_0$ with matrix $A$, the transformation is
$M(t)=A\,R(t)$ with $R(t)$ the matrix in \eqref{eq:ho_map}, so that
\[
\frac{\partial Q}{\partial p}
=-\frac{a_{11}}{m\omega}\sin\omega t+a_{12}\cos\omega t,
\qquad
\frac{\partial P}{\partial p}
=-\frac{a_{21}}{m\omega}\sin\omega t+a_{22}\cos\omega t ,
\]
\[
\frac{\partial Q}{\partial q}
=a_{11}\cos\omega t+a_{12}m\omega\sin\omega t,
\qquad
\frac{\partial P}{\partial q}
=a_{21}\cos\omega t+a_{22}m\omega\sin\omega t .
\]
Each of the four is a sinusoid in $t$ and vanishes identically only if the
corresponding row of $A$ is zero, which $\det A=1$ excludes. Each therefore
vanishes at some time within every period, so no type is valid for all $t$,
whatever the linear frame. Nor do the type-1 and type-2 representations fail
simultaneously: $\partial Q/\partial p=\partial P/\partial p=0$ at one $t$ would
give $A\,(-\sin\omega t/m\omega,\ \cos\omega t)^{\!\top}=0$, contradicting
$\det A=1$.

\section{Scope of the method}
\label{app:scope}

\paragraph{Hamiltonian systems.}Three conditions are inherited from the Hamiltonian setting assumed in Section~\ref{subsec:problem_formulation}. First, the dynamics must preserve the symplectic structure: a Hamiltonian flow
does so, and phase-space volume is preserved in consequence (Liouville's
theorem). The model matches this by construction: every layer satisfies the
symplectic condition (Appendix~\ref{sec:sm_proof}) and in particular has
$\det M = 1$ exactly
(Appendix~\ref{app:invertible}). Second, the flow must be complete, in the sense defined in
Section~\ref{subsec:problem_formulation}: every trajectory is defined for all
$t\in\mathbb{R}$. This excludes systems and regions from which trajectories
escape to infinity, or reach a singularity of $H$, in finite time. Third, the inputs must be a conjugate pair of generalised coordinates
$q \in \mathbb{R}^n$ and momenta $p \in \mathbb{R}^n$, together with time
$t \in \mathbb{R}$.

\paragraph{Autonomous systems.}
Autonomy is assumed from the outset (Section~\ref{subsec:problem_formulation}):
the characterisation \eqref{eq:flow_form} of transformations to conserved
quantities relies on the flow forming a one-parameter group, which holds when $\partial H/\partial t=0$. For a non-autonomous system the evolution from
one time to another depends on both times separately rather than on the elapsed
time alone, so $\phi(\cdot,-t)$ is not a single map and \eqref{eq:flow_form} does
not apply. The temporal shift used in training and the matching
re-encoding used at inference (Sections~\ref{sec:training},
\ref{sec:autoregressive}) are consistent with this restriction.

\section{Design choices for temporal segmentation}
\label{sec:sampling_limitation}
This appendix records the reasoning behind the segmentation of training trajectories and the rule that sets $T_{\max}$, together with the data properties that fix each setting. Neither the segmentation nor the rule is offered as a contribution, and no comparison against alternative schemes is claimed; what is reported is why each choice was made rather than that it is the best available. During training, each segment provides a single observation to the Mapping Network,
which produces the corresponding latents. These are
passed to the Inverse Network together with multiple query times from the same
segment, and the loss compares the resulting predictions against the true
states. What motivates the
limits of the segment length is optimisation. Within a segment the decoder holds the encoded
latents $(Q,P)$ fixed and varies only the query time, so by~\eqref{eq:hj_solution} its
output is the trajectory that $(Q,P)$ parameterises. A small error in the encoded $(Q,P)$ parameterises a neighbouring trajectory instead, so to first order the reconstruction error at elapsed time $t$, the interval between the observation the segment is encoded from and the query time, is the separation between the two. Writing $\delta z$ for the induced error in the phase-space point, that separation is
$\Phi(t)\,\delta z + O(\lVert\delta z\rVert^{2})$, where $\Phi(t)=D_z\phi(z,t)$ is the
fundamental matrix of the linearised flow along the trajectory; it is bounded by $\lVert\Phi(t)\rVert\,\varepsilon$, where $\varepsilon=\lVert\delta z\rVert$ and $\lVert\Phi(t)\rVert\sim \exp(t/t_\lambda)$ with $t_\lambda$ the local Lyapunov time (defined in Section~\ref{sec:tmax_selection}). For a Hamiltonian system the flow is volume preserving, so $\det\Phi(t)=1$ and hence $\lVert\Phi(t)\rVert\ge1$: whatever the system, there is always a direction in which the separation does not contract, and the amplification factor is bounded below by one at every elapsed time.

The growth does not continue indefinitely. For chaotic systems confined to bounded invariant regions of phase space, the separation between two trajectories in that region cannot exceed its diameter $D$:

it grows until
\begin{equation}
    \tau(\varepsilon)=t_\lambda\ln(D/\varepsilon),
    \label{eq:tau_eps}
\end{equation}
and is bounded thereafter. The two regimes constrain the loss differently. 

For $t<\tau(\varepsilon)$ the reconstruction error is of order $\varepsilon \exp(t/t_\lambda)$ and increases with $\varepsilon$, so the loss at that query time is reduced by reducing $\varepsilon$, and hence the error in the latents that induces it.

For $t>\tau(\varepsilon)$ the predicted and true trajectories are separated by a distance of order $D$ and are no longer
close: the reconstruction error is bounded by $D$ and no longer varies systematically with $\varepsilon$, so those query times indicate no consistent direction in which the latents should change. The two regimes are the two faces of the same amplification: within the growth regime the sensitivity, and with it the gradient, is large, which is the exploding gradient reported by \citet{chakraborty2024divide} and \citet{Ribeiro2019OnTS}; beyond it the loss no longer varies systematically with $\varepsilon$, so the gradient carries no reliable information about the error it would correct.

A window of length $T$ constrains the latents through those query times that lie within the growth regime. The whole window does so when $T<\tau(\varepsilon)$, that is for errors
\begin{equation}
    \varepsilon<\varepsilon^\ast(T)=D\exp(-T/t_\lambda).
    \label{eq:basin}
\end{equation}
Above this threshold the informative query times are confined to an initial part of the window, while the remainder contributes terms that indicate no direction of correction. The range of errors the loss can correct therefore shrinks exponentially with the window length. Equation~\eqref{eq:basin} inherits both the linearisation and the exponential bound on $\lVert\Phi(t)\rVert$, so it locates a scale rather than a sharp threshold, and it restates the sensitivity argument of \citet{Ribeiro2019OnTS} and \citet{chakraborty2024divide} in the present setting rather than establishing a new result. The expression is never evaluated: it is used only to identify the scale on which $T_{\max}$ is set.

The discrete-time counterpart of this statement is the one given by
\citet{Ribeiro2019OnTS} for prediction-error estimation, and it is worth
recording why it transfers to a setting in which no state is propagated. There the
prediction at step $k$ is formed by composing a one-step state map $h$ across the
window, and its sensitivity to the initial condition the window is simulated from is
$C_k\prod_{j}A_j$, the product of the one-step state Jacobians $A_j=\partial h/\partial x$
along the window. Their bounds on the Lipschitz constant and $\beta$-smoothness of the objective follow from bounding the one-step Jacobians by a Lipschitz constant $L_h$ of the state map, so that products across the window are bounded by $L_h^{k}$; they therefore grow with the number of steps propagated in the same way that $\lVert\Phi(t)\rVert$ grows with elapsed time here. The two play the same role, and where the one-step map is a discretisation of the flow they are the same object: the derivative of the time-$t$ map with respect to the phase-space point it is taken from, a property of the map rather than of the procedure used to evaluate it: a decoder that reproduces $\phi(z,t)$ in a single
evaluation has the same derivative $D_z\phi(z,t)$ as a composition of one-step maps
that reproduces it, the derivative being fixed by the map and not by the factorisation
used to compute it. What differs between the two settings is only the quantity the
matrix acts on --- there the error in the initial condition, here the phase-space
error $\delta z$ induced by an error in the encoded $(Q,P)$.

Because $\varepsilon^\ast(T)$ decays on the scale of $t_\lambda$, the selection rule sets $T_{\max}$ on that scale. Basing the window length on this scale has precedent. \citet{chakraborty2024divide} state the same principle, that the window be kept within
the fastest Lyapunov time scale of the system, and for the same underlying reason: the
sensitivity of the reconstructed trajectory to the quantities being fitted obeys a linear equation whose homogeneous part is the linearised flow so it is
amplified at the rate the Lyapunov exponent gives regardless of what those
quantities are. In their case the fitted quantities are the parameters of the
vector field and the amplified object is $\partial q/\partial\Theta$, which they
report reaching magnitudes at which the gradient is no longer usable for
optimisation; here the fitted quantities are the encoded $(Q,P)$,
which by~\eqref{eq:hj_solution} parameterises the trajectory, and the amplified
object is the separation $\Phi(t)\,\delta z$ between the parameterised and the true trajectory. In both cases the window must be kept
within the horizon over which that amplification remains bounded.

A related issue is the use of a constant segmentation horizon across the full phase 
space. In more chaotic regions, the underlying $t_\lambda$ is inherently shorter. A single long segment length may therefore exceed the local horizon \eqref{eq:basin} in those regions, while a short one diminishes the efficiency gains from the method's parallel intra-segment prediction. Despite this, we deliberately restrict this work to a single constant $T_{\max}$: it keeps
the focus on the representation-learning method and avoids introducing further mechanisms
that would themselves require separate validation, deferring the adaptive treatment to future
work. This restriction forces the rule to be conservative so the estimated bound falls short of the true window
rather than past it. Finally, for integrable systems the maximal Lyapunov exponent vanishes and nearby
trajectories separate no faster than polynomially, so $\tau(\varepsilon)$ grows as
a power of $D/\varepsilon$ rather than as its logarithm, $\varepsilon^\ast(T)$
decays polynomially in $T$, and the constraint is much more relaxed. The rule
below accounts for this without a separate case: as $\lambda\to 0$ the local
Lyapunov time diverges, the Lyapunov branch of~\eqref{eq:tmax_rule} ceases to
bind, and $T_{\max}$ falls back on the duration of the record. We set out the rule
in Section~\ref{sec:tmax_selection}.

\subsection{Selecting the segment length from data}
\label{sec:tmax_selection}

\paragraph{Selection rule.}
Along the trajectory through $z$, the finite-time Lyapunov exponent $\lambda(z)=t^{-1}\ln\lVert\Phi(t)\rVert$ is the rate at which nearby trajectories
separate over a window of length $t$, and $t_\lambda(z)=1/\lambda(z)$ is the corresponding local Lyapunov time; the long-time limit of $\lambda(z)$ is the maximal Lyapunov exponent. We use the local quantity, since what a segment must respect is the
growth over a window of its own length and in the region of phase space it visits. The finite-time exponent is defined relative to a horizon $t$, which is the
quantity being selected; the estimator of the next section resolves
this by fitting $\lambda(z)$ as the growth \emph{rate} of the separation over an
intermediate scaling region rather than evaluating it at any single horizon, with
that region fixed by properties of the record and not by $T_{\max}$. Subject only to the constraint that a segment cannot outlast the recorded signal,
we set
\begin{equation}
    T_{\max}=\min\bigl(T_{\mathrm{record}},\;t_\lambda^{\min}\bigr),
    \qquad
    t_\lambda^{\min}=\min_{z}\,t_\lambda(z),
    \label{eq:tmax_rule}
\end{equation}
where $T_{\mathrm{record}}$ is the duration of the longest training trajectory and
$t_\lambda^{\min}$ the minimum local Lyapunov time over the training observations
$z=(q,p,t)$. We take the minimum over $z$ deliberately: a single constant $T_{\max}$ is safe only if it respects the least predictable region observed, since
a window valid in the worst observed case is valid throughout the sampled region. Equation~\eqref{eq:tmax_rule} is a conservative default: it fixes $T_{\max}$ from
the training data before optimisation begins rather than by trialling values
against the evaluation. We make no claim that it is the best available choice;
the results reported for each task in Section~\ref{sec:eval_protocol} are those
obtained at the single value the rule returns. What now remains is to define the estimator of $t_\lambda(z)$. Such an estimator
must respect the physics yet work without reference to the system's equations of
motion, so that the rule applies unchanged to data of unknown origin. Appendix~\ref{app:lyap_estimator} specifies the one we use.

\section{Specification of the local Lyapunov-time estimator}
\label{app:lyap_estimator}

The estimator below is assembled from published components and no part of it is proposed as a contribution. It is recorded so that every setting entering $T_{\max}$ is specified and traceable to the property of the record that fixed it. What the rule requires of it is the ordering in~\eqref{eq:tmax_rule}, not an accurate exponent, and the checks reported here concern only whether the dominant finite-sample errors act to shorten $T_{\max}$ rather than lengthen it.

\paragraph{Estimator.}
We require only a conservative horizon bound. The minimum in~\eqref{eq:tmax_rule} is one-sided in this direction: an under-estimate at any single reference shortens $T_{\max}$, whereas lengthening it would require the references near the minimum to be over-estimated, or discarded as regular, together. We accordingly assemble the estimator from components taken from the cited literature, stating below the composition rules and thresholds it requires; what the rule needs of it is the ordering in~\eqref{eq:tmax_rule}, not a tight estimate, and the paragraphs below check that ordering on each task. Because each observation is the full phase-space and time state $(q,p,t)$ rather than a scalar measurement, we work directly in $(q,p)$, omitting the delay-coordinate embedding required by scalar-series methods and its delicate embedding-dimension and delay parameters. Working in phase space also lets us estimate $t_\lambda$ per observation rather than one global exponent per trajectory, which matters because in general a system's phase space is heterogeneous---regular in some regions, chaotic in others---and cannot be described by a single exponent. When the dynamics are regular---as for an integrable system, or on a quasi-periodic orbit of
a non-integrable one---neighbouring trajectories separate at most polynomially rather than
exponentially, so the true $\lambda_{\max}=0$ and the local Lyapunov time is unbounded. On small, possibly noisy datasets an estimator may nonetheless return large but finite values for regular dynamics instead of a strictly unbounded one; we treat such near-zero slopes explicitly in the labelling rule below.

\paragraph{Estimation procedure.}
The estimate proceeds in four steps. \emph{(i)~Scaling.} Each conjugate pair $(q_i,p_i)$ is rescaled by $\alpha_i=\sqrt{\sigma_{q_i}/\sigma_{p_i}}$ ($q_i\!\mapsto\!q_i/\alpha_i$, $p_i\!\mapsto\!\alpha_i p_i$), placing position and momentum on a common scale so Euclidean neighbour distances are meaningful. The rescaling is diagonal and symplectic, preserving the Hamiltonian setting, and---as a smooth invertible coordinate change---leaves the asymptotic $\lambda_{\max}$ invariant \citep{eckmann1985}, introducing no asymptotic freedom. The finite-time local exponent is nonetheless defined relative to the metric the rescaling induces, so we fix $\alpha_i$ from the data before any estimate rather than treat it as a tunable parameter.\emph{(ii)~Neighbours.} For each reference observation
we collect its neighbours within a radius $\varepsilon$ in the scaled phase space using a
KD-tree. Neighbours from other trajectories are always admitted; those from the same trajectory only if separated by more than a Theiler window $w$, so that neighbours are counted only when close by recurrence rather than merely close in time \citep{theiler1986,kantzschreiber2003}.\emph{(iii)~Divergence curve.} Using the common sampling step, we advance the reference and each admitted neighbour by $k$ steps and, for each $k$, record the logarithm of their mean Euclidean separation, $S(k)=\ln\big\langle \lVert x_{\mathrm{ref}}(k)-x_j(k)\rVert \big\rangle_j$, where $\langle\cdot\rangle_j$ averages over the neighbours $j$; averaging distances before taking the logarithm follows \citet{kantz1994}. For a chaotic reference, $S(k)$ grows linearly in $k$ over an intermediate \emph{scaling region}---bounded below by the transient during which a generic separation vector rotates into the most unstable direction, and above by saturation at the scale of the region the trajectories occupy \citep{kantz1994,rosenstein1993practical}. \emph{(iv)~Slope and region detection.} The slope of $S(k)$ over its scaling region---and whether such a region exists---is extracted by the ensemble method of \citet{deshmukh2021}, which fits all sub-intervals of the curve, forms a weighted
distribution of their slopes, and returns the distribution's mode as the slope together with
a width-based verdict that flags the absence of any well-defined region (a broad slope distribution, indicating no interval over which the fits agree). Following \citet{kantz1994}, we repeat steps~(ii)--(iv) over a range of radii $\varepsilon$ and require consistency across them, accepting a positive exponent only when a valid region is recovered at a majority of the radii. The accepted slopes are averaged, and the local Lyapunov time is the reciprocal of the resulting rate, converted from steps to time units by the sampling step. Because neighbour sets at successive radii are nested, adjacent radii yield near-identical slopes and a spuriously small spread; requiring a majority of the radii forces the accepted set to span a substantial part of the geometric grid, so the spread is taken across well-separated scales. A reference is labelled regular ($t_\lambda=\infty$) when its averaged slope is smaller in magnitude than twice its across-$\varepsilon$ spread, or when it fails the cross-radius consistency check (fewer than a majority of radii yield a valid region), and is left undefined when its trajectory affords too few forward steps to form a curve---as at the end of every
trajectory. Two settings vary per task: the minimum scaling-region width $r_{\min}$ and the percentile pair bounding the radius grid. The minimum neighbour count $n_{\min}$, the scaling factors $\alpha_i$, the Theiler window $w$, and the look-ahead horizon $K_{\max}$ are fixed from the data by the rules stated here and below. The remaining settings are held identical across all tasks. The ensemble weight exponents $(\ell,e)=(1,1)$ are the length- and error-weight exponents \citet{deshmukh2021} use for exponent estimation. The number of radii is $n_\varepsilon=8$, large enough that the majority quorum is a meaningful stability check yet small enough that the radii stay well separated on the geometric grid. The region-width threshold rejects a candidate region when the FWHM of its slope distribution exceeds one quarter of the total slope range, using the FWHM that \citet{deshmukh2021} report as an indicator of whether a scaling region is present; the quarter is a fixed constant here, not tuned per task. A reference is declared regular when its averaged slope falls below twice the across-$\varepsilon$ spread, treating a slope no larger than the scatter across radii as not resolvably positive. The cross-radius quorum is a majority of the $n_\varepsilon$ radii, for the nesting reason given above. Each per-reference curve is truncated to $4\,r_{\min}$ steps, fixing the ratio of curve length to minimum fit width, and with it the size of the ensemble of admissible sub-intervals, across tasks; beyond it the average is taken over neighbours whose own trajectories have not yet ended. Only $r_{\min}$ and the two radius-grid percentiles vary between tasks; both are set from the stated principles below.

\paragraph{Application to the training data.}
We now apply the estimator to each task's training set, reporting the chosen settings and resulting values. The minimum neighbour count is held fixed at $n_{\min}=5$ across all tasks, a small count that still admits a slope fit while keeping the neighbourhood local. The minimum scaling-region width $r_{\min}$ is chosen per task under two competing constraints. It must exceed roughly one oscillation of the leading coordinate, since over shorter windows the divergence curve $S(k)$ is dominated by periodic modulation of the separation rather than its exponential trend, so the fit measures the modulation instead of the exponent. It must also remain below the usable look-ahead of the shortest trajectory, leaving enough sub-intervals for the ensemble fit. On the short one-degree-of-freedom Tasks~1--2 the record spans only about one oscillation, so the look-ahead ceiling binds first, and we set $r_{\min}=8$ (Task~1) and $r_{\min}=7$ (Task~2). A sub-oscillation window would matter only if it produced spurious positive exponents; the estimator instead returns a regular verdict on almost every reference ($3$ and $10$ finite values among the $528$ and $525$ references that receive one), and every finite value exceeds $T_{\mathrm{record}}$, so none can bind. The outcome rests on the ordering $t_\lambda^{\min}>T_{\mathrm{record}}$, with margins of $5.8\times$ and $3.3\times$.

On Hénon--Heiles the floor binds and we set $r_{\min}=42$, approximately one oscillation of the leading coordinate estimated from the autocorrelation---the same first-zero-crossing lag used to set the Theiler window, giving a period of $\approx2w=44$ samples; the divergence curves there are long enough ($334$ samples) that the ceiling is inactive. On Task~3 the floor binds as well and we set $r_{\min}=32$, again approximately one oscillation of the leading coordinate at $\approx2w=32$ samples and corroborated by the first autocorrelation peak at $34$; the record is long enough ($444$ samples) that the ceiling is again inactive. The neighbourhood radii $\varepsilon$ are swept over a geometric grid of $n_\varepsilon=8$ points between a low and a high percentile of the pairwise-distance distribution in the scaled phase space, and the slope is averaged over the radii that yield a valid region, as in \citet{kantz1994}. A percentile is not a transferable radius specification across phase spaces of differing dimension: empirically $\varepsilon(p)\propto p^{1/d_{\mathrm{eff}}}$ , with $d_{\mathrm{eff}}\approx 2.0$--$2.25$ on the one-degree-of-freedom Tasks~1--2 and $\approx 3.0$ on the two-degree-of-freedom Task~4, so a common pair of percentiles would place the higher-dimensional grid either below the sampling density or above the linearisation scale. Nor is a percentile transferable across records of different size: because it is read from the distribution of pairwise distances, the neighbours admitted at percentile $p$ number on average $p$ times the sample count, so on the $444$-sample Task~3 the first percentile admits a median of four neighbours, fewer than $n_{\min}$. We therefore fix the transferable quantity---the admitted neighbour count---rather than the percentile.

For each task we read from the pairwise-distance distribution the low and high percentiles whose radii sweep the neighbour count from a few times $n_{\min}$ at $\varepsilon_{\mathrm{lo}}$ (populating the innermost radius for essentially every reference) to a few tens of $n_{\min}$ at $\varepsilon_{\mathrm{hi}}$, realising Kantz's requirement that $\varepsilon$ be populated yet small relative to the attractor \citep{kantz1994,kantzschreiber2003}. This gives $(\varepsilon_{\mathrm{lo}},\varepsilon_{\mathrm{hi}})$ percentiles of $(1,20)$ on Tasks~1--2 and $(0.1,1)$ on Task~4, yielding a comparable neighbour-count sweep---and radii spanning $5$--$26\%$ of the phase-space extent on Tasks~1--2 and $7$--$16\%$ on Task~4. On Task~3 the sampling density just noted raises $\varepsilon_{\mathrm{lo}}$ to the fifth percentile, and we hold $\varepsilon_{\mathrm{hi}}$ at the tenth to keep the top radius at $15\%$ of the extent, near Task~4's; the resulting sweep of $5$--$9\,n_{\min}$ neighbours is narrower than elsewhere, which makes the cross-radius checks less discriminating and so admits finite exponents rather than suppressing them. 

The Theiler window $w$ is set to twice the first zero-crossing lag of the autocorrelation, a common heuristic realisation of the requirement that $w$ exceed the autocorrelation time \citep{theiler1986,kantzschreiber2003}. The maximum look-ahead $K_{\max}$ is bounded by the longest trajectory; the per-reference truncation to $4\,r_{\min}$ steps is one of the fixed constants listed above. Table~\ref{tab:lyap_stats} summarises the outcome. On the two integrable tasks the vast majority of samples are labelled regular and the smallest finite $t_\lambda$ exceeds $T_{\mathrm{record}}$; by \eqref{eq:tmax_rule} this yields $T_{\max}=T_{\mathrm{record}}$, and each trajectory is treated as a single segment. On the non-integrable Hénon--Heiles task the minimum binds: $t_\lambda^{\min}=8.563<50=T_{\mathrm{record}}$, so \eqref{eq:tmax_rule} shortens each trajectory into segments. Here a minority of references, in regions whose estimated horizon is shorter than the recorded span, fixes $T_{\max}$. The same occurs on Task~3 ($t_\lambda^{\min}=15.973<75.833$), where most references are regular and the median finite estimate exceeds $T_{\mathrm{record}}$. That this record is measured and not Hamiltonian costs the argument nothing, since \eqref{eq:basin} needs only the growth rate of separations, which the estimator reads off the record. When trajectories differ in length and $T_{\max}=T_{\mathrm{record}}$, $T_{\max}$ is set by the longest; each shorter trajectory still forms a single segment and, after shifting the time values, occupies the sub-interval $[0,\ell]\subseteq[0,T_{\max}]$ rather than the full range.

\begin{table}[htbp]
\centering
\footnotesize
\caption{Local Lyapunov-time estimation on each task's training set. Per-task settings are the minimum scaling-region width $r_{\min}$ and the two percentiles bounding the radius grid; the minimum neighbour count $n_{\min}=5$ is fixed across tasks. All are set from data properties before the estimate (text), and all other settings are fixed constants. Each sample is labelled with a finite $t_\lambda$, $\infty$ (regular), or undefined (curve too short). ``Finite $t_\lambda$'' columns summarise the finite values only. The final columns apply Eq.~\eqref{eq:tmax_rule}: on the integrable Tasks~1--2 the ordering $t_\lambda^{\min}>T_{\mathrm{record}}$ gives $T_{\max}=T_{\mathrm{record}}$; on Tasks~3 and~4 the minimum binds and $T_{\max}=t_\lambda^{\min}<T_{\mathrm{record}}$.}
\label{tab:lyap_stats}
\setlength{\tabcolsep}{4pt}
\begin{tabular}{lcccccccccccc}
\toprule
& \multicolumn{2}{c}{Knobs}
& \multicolumn{4}{c}{Sample labels}
& \multicolumn{3}{c}{Finite $t_\lambda$}
& & \\
\cmidrule(lr){2-3}\cmidrule(lr){4-7}\cmidrule(lr){8-10}
Task & $n_{\min}$ & $r_{\min}$
& Total & Reg.\ ($\infty$) & Finite & Undef.
& median & max & min
& $T_{\mathrm{record}}$ & $T_{\max}$ \\
\midrule
T1 (mass--spring)    & $5$ & $8$  & $768$   & $525$  & $3$    & $240$
& $59.92$ & $62.03$  & $37.75$ & $6.493$ & $6.493$ \\
T2 (ideal pendulum)  & $5$ & $7$  & $750$   & $515$  & $10$   & $225$
& $19.15$ & $30.45$  & $13.46$ & $4.060$ & $4.060$ \\
T3 (real pendulum)   & $5$ & $32$ & $444$   & $317$  & $92$   & $35$
& $81.24$ & $337.85$ & $15.973$ & $75.833$ & $15.973$ \\
T4 (H\'enon--Heiles) & $5$ & $42$ & $16032$ & $5316$ & $8604$ & $2112$
& $20.82$ & $279.69$ & $8.563$  & $50.0$  & $8.563$ \\
\bottomrule
\end{tabular}
\end{table}
On the integrable tasks the selected $T_{\max}$ is insensitive to the estimator's precision: every finite estimate exceeds $T_{\mathrm{record}}$ by a wide margin ($5.8\times$ on Task~1, $3.3\times$ on Task~2), so an estimate would have to be too small by more than a factor of three before it altered the outcome, and the decision rests on the robust ordering $t_\lambda^{\min}>T_{\mathrm{record}}$ rather than on exact values. On Task~4 the estimate does enter $T_{\max}$ directly, and there the estimator's bias works in the safe direction. In a mixed phase space, neighbours may lie on distinct invariant tori, which separate linearly through frequency dephasing rather than exponentially; over a finite curve a linear separation is indistinguishable from a slow exponential, so where the estimator errs on near-regular references it errs towards positive rates; this accounts for the finite estimates on an otherwise largely regular phase space. What sets $T_{\max}$, however, is the minimum, and two mechanisms act there: the selected value is the minimum over more than eight thousand finite estimates, an order statistic that is itself biased downward by finite-sample scatter; and that minimum is not an isolated spike but the floor of a tight cluster---the ten smallest finite $t_\lambda$ span $8.563$--$8.98$---so it reflects a genuine most-unstable timescale rather than a single noisy reference. As an independent check on Task~4 we computed a ground-truth maximal
Lyapunov exponent for every training trajectory by Benettin's method
\citep{benettin1980} on the variational equations, integrated to $t=5000$.
Seven of the $48$ trajectories are chaotic ($\lambda_{\max}>10^{-2}$),
confirming the mixed phase space the segmentation rule is designed for;
the most unstable trajectory has $\lambda_{\max}=0.053$, a true predictability
horizon of $t_\lambda=18.9$. The check bears on two points. Qualitatively, the phase space is mixed, which is the regime the segmentation rule is designed for. Quantitatively, the selected $T_{\max}=8.563$ is shorter than the true predictability timescale of even the most unstable trajectory, by a factor of $2.2$---the ordering the rule requires. We do not read that factor as a measure of the estimator's error, because the two numbers are not the same quantity: $t_\lambda^{\min}$ is a local, per-observation horizon, which may legitimately fall below a trajectory's long-time average in the regions where the flow stretches most strongly, and it is in addition a minimum over thousands of estimates and therefore a downward-biased order statistic. Both effects shorten the estimate, so the gap is equally consistent with an accurate estimator evaluated at the locally most unstable observation. What the comparison checks is the ordering, not the tightness.

\section{Toward adaptive localities (future work)}
\label{app:adaptive}
Fixing $T_{\max}$ to the minimum local Lyapunov-time of the training phase space \eqref{eq:tmax_rule} is conservative and has real costs. In
regions more stable than the least predictable one, the horizon is shorter than it need be,
so inference performs more re-encodings than necessary and forfeits part of the efficiency
gain that motivates the segment-based rollout. This motivates a fully adaptive mechanism---for
both training and inference---that instead assigns each phase-space region its own segment length, retaining the spatial variation that \eqref{eq:tmax_rule} discards. We sketch directions toward such a scheme below; none is implemented and used in this paper.

\subsection{Training set segmentation}
\paragraph{Lyapunov exponent estimation.}
A natural, training-agnostic starting point for segment-length selection
could be derived from the Lyapunov time, a
well-established diagnostic of local predictability in dynamical systems
theory as described already. The procedure of Section~\ref{sec:tmax_selection}, computes a spatially varying Lyapunov time that reflects
region-dependent predictability---but collapses it to a single value through the minimum
in \eqref{eq:tmax_rule}.  An adaptive scheme would instead retain the field, assigning
each phase-space region its own segment length from the local $t_\lambda$ rather than from
its global minimum; the same estimator therefore serves both the constant rule used here
and its adaptive generalisation.

\paragraph{Intra-segment temporal error profile.}

A second, model-aware approach could leverage the structure of the
prediction error during training to adjust segment boundaries dynamically
as optimisation proceeds.  A high-level practical implementation could be
as follows: at regular intervals during training, each segment is encoded
from its temporal midpoint and decoded at a grid of query times spanning
the full segment.  The resulting prediction error, recorded as a function
of temporal distance from the midpoint, yields an error profile for that
segment.

In principle, tracking such a signature over multiple training epochs
could distinguish segmentation errors from insufficient convergence.  As
training progresses, poor convergence within a given segment would be
expected to produce errors that decrease relatively uniformly across the
entire segment, whereas an excessively long segment would instead give
rise to a pattern in which boundary errors remain systematically higher
than those near the centre, causing the boundary-to-centre error ratio
to persist or widen even as overall error decreases.  Using this signal
and a predefined threshold, segment lengths could then be adjusted
periodically during training, shortening segments with consistently
elevated ratios and expanding those for which the ratio remains close to
unity, yielding spatially varying segment lengths that adapt to the local
dynamics.

\clearpage
\section{Complete Training Algorithm}
\label{app:training_algorithm}

Algorithm~\ref{alg:training} consolidates the calibration phase and the
per-batch training procedure summarised in the main text.
Phase~I estimates the data-driven normalisation constants
$\bar{\mathcal{L}}_{\mathrm{pred}}$ and $H_{\max}$ prior to training.
Phase~II describes a single gradient-update step; it is repeated for every
mini-batch in every epoch of the main training loop.

\begin{ouralgorithm}{Training procedure for the Mapping\,/\,Inverse network pair.}
\label{alg:training}
\begin{algorithmic}[1]

%% ==== PHASE I ====
\Statex \hrulefill
\Statex \textbf{Phase~I\;---\;Calibration (no gradient updates)}
\Statex \hrulefill

\State Run one full epoch (forward passes only) over all training batches,
       computing $\mathcal{L}_{\mathrm{pred}}$ per batch via
       Phase~II Steps~\ref{line:encode}--\ref{line:pred_loss}, and set:
\Statex \quad
       Let $\mu$ and $s$ be the mean and standard deviation of the per-batch
       losses $\{\mathcal{L}_{\mathrm{pred}}^{(b)}\}_{b=1}^{N_{\mathrm{batches}}}$.
       To guard against a noisy calibration epoch, set
\Statex \quad
       $\displaystyle
        \bar{\mathcal{L}}_{\mathrm{pred}} =
        \begin{cases}
          \operatorname{median}\bigl(\{\mathcal{L}_{\mathrm{pred}}^{(b)}\}\bigr),
            & |\mu/s| < \tau,\\[4pt]
          \mu, & \text{otherwise,}
        \end{cases}$
\Statex \quad
       with threshold $\tau=2$.

\setcounter{ALG@line}{0}

\Statex

%% ==== PHASE II ====
\Statex \hrulefill
\Statex \textbf{Phase~II\;---\;Training (per-batch update)}
\Statex \hrulefill

\Statex \textbf{Input:} one observation per segment from $N$ segments:\;
        $q_{\mathrm{in}},\, p_{\mathrm{in}} \in \mathbb{R}^{N \times n}$,\;
        $t_{\mathrm{in}} \in \mathbb{R}^{N}$.

\Statex

\State \label{line:encode}
       Encode observations via the Mapping Network:
\Statex \quad
       $Q,\; P,\; t_{\mathrm{in}}
        = \mathcal{M}(q_{\mathrm{in}},\; p_{\mathrm{in}},\; t_{\mathrm{in}})$,
       \qquad $Q,\, P \in \mathbb{R}^{N \times n}$

\Statex
\State \textbf{Prediction loss.} \label{line:pred_start}
       Using $Q,\,P$ from Step~\ref{line:encode}:

\Statex \quad
       (a)\; Use every sample of each segment as a query time, padding to the
       longest segment length $L_{\max}$ in the batch and recording a mask
       $M\in\{0,1\}^{N\times L_{\max}}$:\;
       $t_{\mathrm{pred}} \in \mathbb{R}^{N \times L_{\max}}$.
       Replicate $Q,\,P$ along a new dimension of size $L_{\max}$:
\Statex \qquad
       $Q_{\mathrm{full}},\; P_{\mathrm{full}}
        \in \mathbb{R}^{N \times L_{\max} \times n}$

\Statex

\Statex \quad
       (b)\; Decode via the Inverse Network:
\Statex \qquad
       $\hat{q},\; \hat{p},\; t_{\mathrm{pred}}
        = \mathcal{M}^{-1}(Q_{\mathrm{full}},\;
                           P_{\mathrm{full}},\;
                           t_{\mathrm{pred}})$,
       \qquad
       $\hat{q},\, \hat{p}
        \in \mathbb{R}^{N \times N_{\mathrm{pred}} \times n}$

\Statex

\Statex \quad \label{line:pred_loss}
       (c)\; Compute the MAE prediction loss:
\Statex \qquad
       $\displaystyle
        \mathcal{L}_{\mathrm{pred}}
        = \frac{\sum_{j=1}^{N}\sum_{m=1}^{L_{\max}}\sum_{i=1}^{n}
                M_{jm}\,\bigl(|\hat{q}_{jmi} - q_{jmi}|
                            + |\hat{p}_{jmi} - p_{jmi}|\bigr)}
               {n\sum_{j=1}^{N}\sum_{m=1}^{L_{\max}} M_{jm}}$,
\Statex \qquad
       where $M_{jm}\in\{0,1\}$ masks padded query times and $L_{\max}$ is the
       longest segment length in the batch.

\Statex

\State \textbf{Independence loss.}
       Using $Q,\,P$ from Step~\ref{line:encode}:

\Statex \quad
       \textbf{for} each degree of freedom $i = 1, \dots, n$ \textbf{do}

\Statex \quad
       (a)\; Extract $Q^{(i)},\, P^{(i)} \in \mathbb{R}^{N}$.
       Compute RBF kernel matrices
       $K^{(i)},\, L^{(i)} \in \mathbb{R}^{N \times N}$:
\Statex \qquad
       $\displaystyle
        K_{rs}^{(i)}
        = \exp\!\left(
            -\frac{\|Q_r^{(i)} - Q_s^{(i)}\|^2}{2\,\sigma_{Q_i}^2}
          \right)$,
       \qquad
       $\displaystyle
        L_{rs}^{(i)}
        = \exp\!\left(
            -\frac{\|P_r^{(i)} - P_s^{(i)}\|^2}{2\,\sigma_{P_i}^2}
          \right)$
\Statex \qquad
       with bandwidths set by the median heuristic:
\Statex \qquad
$\displaystyle
\sigma_{Q_i}^2
= \operatorname{median}\!\bigl(
\{\|Q_r^{(i)} - Q_s^{(i)}\|^2 : r < s\}
\bigr),
\quad
\sigma_{P_i}^2
= \operatorname{median}\!\bigl(
\{\|P_r^{(i)} - P_s^{(i)}\|^2 : r < s\}
\bigr)$

\Statex

\Statex \quad
       (b)\; Set the diagonal entries of $K^{(i)}$ and $L^{(i)}$ to zero,
       yielding $\tilde{K}^{(i)},\, \tilde{L}^{(i)} \in \mathbb{R}^{N \times N}$.

\Statex

\Statex \quad
       (c)\; Compute the unbiased HSIC estimator,
       where $\mathbf{1} \in \mathbb{R}^N$ is a vector of ones:
\Statex
       {\small $\displaystyle
        \mathrm{HSIC}_i
        = \frac{1}{N(N{-}3)}
          \left[
            \operatorname{Tr}\!\bigl(\tilde{K}^{(i)}\tilde{L}^{(i)}\bigr)
            + \frac{
                (\mathbf{1}^{\!\top}\tilde{K}^{(i)}\mathbf{1})\,
                (\mathbf{1}^{\!\top}\tilde{L}^{(i)}\mathbf{1})
              }{(N{-}1)(N{-}2)}
            - \frac{
                2\,\mathbf{1}^{\!\top}
                (\tilde{K}^{(i)}\tilde{L}^{(i)})\,\mathbf{1}
              }{N{-}2}
          \right]$}

\Statex

\Statex \quad
       \textbf{end for}

\Statex

\Statex \quad
       (d)\; Compute the mean HSIC loss:
\Statex \qquad
       $\displaystyle
        \mathcal{L}_{\mathrm{HSIC}}
        = \frac{1}{n} \sum_{i=1}^{n} \mathrm{HSIC}_i$

\Statex
\State \textbf{Per-batch HSIC normalisation.}
       Let $N$ be the number of segments in this batch and
       $\mathbf{s}=[1,2,\dots,N]$. Compute
\Statex \quad
       $H_{\max}=\max\!\bigl(\mathrm{HSIC}(\mathbf{s},\mathbf{s}),\;0.05\bigr)$,
\Statex \quad
       using the same unbiased estimator and median-heuristic bandwidth as
       above; the floor of $0.05$ prevents division instability.
\State Compute the normalised total loss:
\Statex \quad
       $\displaystyle
        \mathcal{L}_{\mathrm{total}}
        = \lambda_{\mathrm{pred}}\,
          \frac{\mathcal{L}_{\mathrm{pred}}}
               {\bar{\mathcal{L}}_{\mathrm{pred}}}
        + \lambda_{\mathrm{HSIC}}\,
          \frac{\mathcal{L}_{\mathrm{HSIC}}}
               {H_{\max}}$

\State Backpropagate $\mathcal{L}_{\mathrm{total}}$ and update
       all network parameters.

\end{algorithmic}
\end{ouralgorithm}

\section{Autoregressive Handoff Inference Protocol}
\label{app:handoff}
\subsection{Full algorithm}
\begin{ouralgorithm}{Autoregressive Handoff Protocol}
\label{alg:handoff}
 
\noindent\textbf{Input:}
\begin{itemize}
  \setlength{\itemsep}{1pt}\setlength{\parskip}{0pt}
  \item $\textit{MappingNet},\ \textit{InverseNet}$: the trained Mapping and Inverse networks.
  \item $z_{\text{obs}} = (q_{\text{obs}},\, p_{\text{obs}},\, t_{\text{obs}})$, with
        $q_{\text{obs}}, p_{\text{obs}} \in \mathbb{R}^n$ and
        $t_{\text{obs}} \in \mathbb{R}$: the single ground-truth observation
        for a system with $n$ degrees of freedom.
  \item $[t_{\min},\, t_{\max}]$: target prediction interval, with
        $t_{\min} < t_{\max}$.
  \item $T_{\text{max}} > 0$: maximum temporal horizon of the
        trained model.
  \item $\mathcal{T} = \{t_1 < t_2 < \cdots\}$: ordered set of query times,
        with all $t_i \in [t_{\min},\, t_{\max}]$.
\end{itemize}
 
\noindent\textbf{Output:}
$\hat{Z}$: predicted trajectory
$\{(\hat{q}_i,\, \hat{p}_i)\}_{t_i \in \mathcal{T}}$,
with $\hat{q}_i, \hat{p}_i \in \mathbb{R}^n$.
 
\medskip
\noindent\textbf{Initialisation:}
\ResetAlgLine
\begin{algorithmic}[1]
\State $\hat{Z}_{\text{obs}} \gets \emptyset$,\quad
       $\hat{Z}_{\text{fwd}} \gets \emptyset$,\quad
       $\hat{Z}_{\text{bwd}} \gets \emptyset$
\State If $t_{\text{obs}} \in \mathcal{T}$:\quad
       $\hat{Z}_{\text{obs}} \gets \hat{Z}_{\text{obs}} \cup
       \{(q_{\text{obs}},\, p_{\text{obs}})\}$
\State $\mathcal{T}' \gets \mathcal{T} \setminus \{t_{\text{obs}}\}$
\State If $\mathcal{T}' = \emptyset$: \textbf{return} $\hat{Z}_{\text{obs}}$
\end{algorithmic}
 
% ----- Case 1 -----
\medskip
\noindent\textbf{Case 1: Direct Prediction}\\[2pt]
\textit{Executed if\/ $t_{\min} \leq t_{\text{obs}} \leq t_{\max}$ and
$t_{\max} - t_{\min} \leq T_{\text{max}}$.
The entire target interval fits within a single normalised window;
no autoregressive propagation is required.}
\ResetAlgLine
\begin{algorithmic}[1]
\State $z_{\text{in}} \gets
       (q_{\text{obs}},\, p_{\text{obs}},\,
       t_{\text{obs}} - t_{\min})$
\State $Q,\, P,\, (t_{\text{obs}} - t_{\min}) \gets
       \textit{MappingNet}(z_{\text{in}})$,
       \quad $Q, P \in \mathbb{R}^n$
\State $t_{\text{rel}} \gets
       \{t_i - t_{\min} \mid t_i \in \mathcal{T}'\}$
\State Create $Q_{\text{full}},\, P_{\text{full}}
       \in \mathbb{R}^{|\mathcal{T}'| \times n}$ by replicating
       $Q,\, P \in \mathbb{R}^n$ along a new leading dimension
       of size $|\mathcal{T}'|$
\State $\hat{q},\, \hat{p},\, t_{\text{rel}} \gets
       \textit{InverseNet}(Q_{\text{full}},\, P_{\text{full}},\,
       t_{\text{rel}})$,
       \quad $\hat{q},\, \hat{p}
       \in \mathbb{R}^{|\mathcal{T}'| \times n}$
\State \textbf{return}
       $\hat{Z}_{\text{obs}} \cup
       \{(\hat{q}_i,\, \hat{p}_i)\}_{t_i \in \mathcal{T}'}$
\end{algorithmic}
 
% ----- Case 2 -----
\medskip
\noindent\textbf{Case 2: Autoregressive Propagation}\\[2pt]
\textit{Executed when direct prediction conditions are not met.}
 
% ----- Phase 1 -----
\medskip
\noindent\textbf{Phase 1: Forward Propagation}\\[2pt]
\textit{Executed if\/ $t_{\max} > t_{\text{obs}}$.}
\ResetAlgLine
\begin{algorithmic}[1]
\State Initialise anchor:
       $q_{\text{curr}} \gets q_{\text{obs}}$,\quad
       $p_{\text{curr}} \gets p_{\text{obs}}$,\quad
       $t_{\text{curr}} \gets t_{\text{obs}}$,
       \quad with $q_{\text{curr}},\, p_{\text{curr}} \in \mathbb{R}^n$
\State $K_{\text{fwd}} \gets
       \lceil (t_{\max} - t_{\text{obs}}) / T_{\text{max}} \rceil$
\For{$k = 0, 1, \ldots, K_{\text{fwd}}-1$}
 
  \State[\textbf{(i)}] Define window boundaries:
  \[
    T_{\text{start}}^{(k)} = t_{\text{curr}}, \qquad
    T_{\text{end}}^{(k)} = \min(t_{\text{curr}} + T_{\text{max}},\; t_{\max})
  \]
 
  \State[\textbf{(ii)}] Identify query times:
  \[
    I_k^{\text{query}} =
    \begin{cases}
      \{t_i \in \mathcal{T}' \mid
        T_{\text{start}}^{(k)} \leq t_i < T_{\text{end}}^{(k)}\}
        & k < K_{\text{fwd}}-1 \\
      \{t_i \in \mathcal{T}' \mid
        T_{\text{start}}^{(k)} \leq t_i \leq T_{\text{end}}^{(k)}\}
        & k = K_{\text{fwd}}-1
    \end{cases}
  \]
 
  \State[\textbf{(iii)}] Augment non-terminal segments with boundary:
  \[
    I_k =
    \begin{cases}
      I_k^{\text{query}} \cup \{T_{\text{end}}^{(k)}\}
        & k < K_{\text{fwd}}-1 \\
      I_k^{\text{query}}
        & k = K_{\text{fwd}}-1
    \end{cases}
  \]
 
  \State[\textbf{(iv)}] Compute normalised times:
  \[
    t_{\text{rel}}^{(k)} =
    \{t_i - T_{\text{start}}^{(k)} \mid t_i \in I_k\}
  \]
 
  \State[\textbf{(v)}] Set segment observation at relative time zero:
  \[
    z_{\text{in}}^{(k)} = (q_{\text{curr}},\; p_{\text{curr}},\; 0)
  \]
 
  \State[\textbf{(vi)}] Encode to latents:
  \[
    Q_k,\; P_k,\; 0 =
    \textit{MappingNet}(z_{\text{in}}^{(k)}),
    \qquad Q_k,\, P_k \in \mathbb{R}^n
  \]
 
  \State[\textbf{(vii)}] Create tensors
  $Q_{\text{full}}^{(k)},\, P_{\text{full}}^{(k)}
  \in \mathbb{R}^{|I_k| \times n}$ by replicating
  $Q_k,\, P_k \in \mathbb{R}^n$ along a new leading dimension
  of size $|I_k|$.
 
  \State[\textbf{(viii)}] Predict coordinates for all times in $I_k$:
  \[
    \hat{q}_k,\; \hat{p}_k,\; t_{\text{rel}}^{(k)} =
    \textit{InverseNet}(Q_{\text{full}}^{(k)},\;
    P_{\text{full}}^{(k)},\; t_{\text{rel}}^{(k)}),
    \qquad \hat{q}_k,\, \hat{p}_k \in \mathbb{R}^{|I_k| \times n}
  \]
 
  \State[\textbf{(ix)}] Store predictions:
  $\hat{Z}_{\text{fwd}} \gets \hat{Z}_{\text{fwd}} \cup
  \{(\hat{q}_{k,i},\; \hat{p}_{k,i})\}_{t_i \in I_k^{\text{query}}}$
 
  \State[\textbf{(x)}] If $k < K_{\text{fwd}}-1$:
  \[
    q_{\text{curr}} \gets \hat{q}_k[-1,:] \in \mathbb{R}^n, \qquad
    p_{\text{curr}} \gets \hat{p}_k[-1,:] \in \mathbb{R}^n, \qquad
    t_{\text{curr}} \gets T_{\text{end}}^{(k)}
  \]
\EndFor
\end{algorithmic}
 
% ----- Phase 2 -----
\medskip
\noindent\textbf{Phase 2: Backward Propagation}\\[2pt]
\textit{Executed if\/ $t_{\min} < t_{\text{obs}}$.}
\ResetAlgLine
\begin{algorithmic}[1]
\State Initialise anchor:
       $q_{\text{curr}} \gets q_{\text{obs}}$,\quad
       $p_{\text{curr}} \gets p_{\text{obs}}$,\quad
       $t_{\text{curr}} \gets t_{\text{obs}}$,
       \quad with $q_{\text{curr}},\, p_{\text{curr}} \in \mathbb{R}^n$
\State $K_{\text{bwd}} \gets
       \lceil (t_{\text{obs}} - t_{\min}) / T_{\text{max}} \rceil$
\For{$k = 0, 1, \ldots, K_{\text{bwd}}-1$}
 
  \State[\textbf{(i)}] Define window boundaries:
  \[
    T_{\text{start}}^{(k)} =
    \max(t_{\text{curr}} - T_{\text{max}},\; t_{\min}), \qquad
    T_{\text{end}}^{(k)} = t_{\text{curr}}
  \]
 
  \State[\textbf{(ii)}] Identify query times:
  \[
    I_k^{\text{query}} =
    \begin{cases}
      \{t_i \in \mathcal{T}' \mid
        T_{\text{start}}^{(k)} < t_i \leq T_{\text{end}}^{(k)}\}
        & k < K_{\text{bwd}}-1 \\
      \{t_i \in \mathcal{T}' \mid
        T_{\text{start}}^{(k)} \leq t_i \leq T_{\text{end}}^{(k)}\}
        & k = K_{\text{bwd}}-1
    \end{cases}
  \]
 
  \State[\textbf{(iii)}] Augment non-terminal segments with boundary:
  \[
    I_k =
    \begin{cases}
      \{T_{\text{start}}^{(k)}\} \cup I_k^{\text{query}}
        & k < K_{\text{bwd}}-1 \\
      I_k^{\text{query}}
        & k = K_{\text{bwd}}-1
    \end{cases}
  \]
 
  \State[\textbf{(iv)}] Compute normalised times:
  \[
    t_{\text{rel}}^{(k)} =
    \{t_i - T_{\text{start}}^{(k)} \mid t_i \in I_k\}
  \]
 
  \State[\textbf{(v)}] Set segment observation at the end of the
  normalised window:
  \[
    z_{\text{in}}^{(k)} =
    \!\left(q_{\text{curr}},\; p_{\text{curr}},\;
    T_{\text{end}}^{(k)} - T_{\text{start}}^{(k)}\right)
  \]
 
  \State[\textbf{(vi)}] Encode to latents:
  \[
    Q_k,\; P_k,\;
    \!\left(T_{\text{end}}^{(k)} - T_{\text{start}}^{(k)}\right) =
    \textit{MappingNet}(z_{\text{in}}^{(k)}),
    \qquad Q_k,\, P_k \in \mathbb{R}^n
  \]
 
  \State[\textbf{(vii)}] Create tensors
  $Q_{\text{full}}^{(k)},\, P_{\text{full}}^{(k)}
  \in \mathbb{R}^{|I_k| \times n}$ by replicating
  $Q_k,\, P_k \in \mathbb{R}^n$ along a new leading dimension
  of size $|I_k|$.
 
  \State[\textbf{(viii)}] Predict coordinates for all times in $I_k$:
  \[
    \hat{q}_k,\; \hat{p}_k,\; t_{\text{rel}}^{(k)} =
    \textit{InverseNet}(Q_{\text{full}}^{(k)},\;
    P_{\text{full}}^{(k)},\; t_{\text{rel}}^{(k)}),
    \qquad \hat{q}_k,\, \hat{p}_k \in \mathbb{R}^{|I_k| \times n}
  \]
 
  \State[\textbf{(ix)}] Store predictions:
  $\hat{Z}_{\text{bwd}} \gets \hat{Z}_{\text{bwd}} \cup
  \{(\hat{q}_{k,i},\; \hat{p}_{k,i})\}_{t_i \in I_k^{\text{query}}}$
 
  \State[\textbf{(x)}] If $k < K_{\text{bwd}}-1$:
  \[
    q_{\text{curr}} \gets \hat{q}_k[0,:] \in \mathbb{R}^n, \qquad
    p_{\text{curr}} \gets \hat{p}_k[0,:] \in \mathbb{R}^n, \qquad
    t_{\text{curr}} \gets T_{\text{start}}^{(k)}
  \]
\EndFor
\end{algorithmic}
 
% ----- Final Assembly -----
\medskip
\noindent\textbf{Final Assembly:}
\[
  \text{Return}\quad
  \hat{Z} = \hat{Z}_{\text{bwd}} \cup
  \hat{Z}_{\text{obs}} \cup \hat{Z}_{\text{fwd}}
\]
 
\end{ouralgorithm}

\subsection{Computational Cost}
When the target prediction interval fits within a single segment, the entire inference
requires exactly two forward passes: one through $\mathcal{M}$ and one through
$\mathcal{M}^{-1}$.
For longer horizons the cost scales to $2(K_{\mathrm{fwd}} + K_{\mathrm{bwd}})$
network calls, where $K_{\mathrm{fwd}}$ and $K_{\mathrm{bwd}}$ are the number of
forward and backward windows from the observation time.

Furthermore, the forward and backward propagation phases in Case~2 are
independent, since both are anchored at the observed state, and can in
principle be executed concurrently. Likewise, independent trajectories
can be processed in parallel, as their encoding and decoding operations
have no sequential dependencies. The current implementation exploits
trajectory-level batching but does not parallelise the forward and
backward propagation phases.

\section{Experimental Details}
\label{app:experimental_details}

\subsection{Task and Dataset Specifications}
\label{app:task_specs}

Table~\ref{tab:hparams_task} summarises dataset sizes, horizons, and
sampling properties for each task. Data-generation scripts are included in
the supplementary code repository.

\begin{table}[htbp]
\centering
\caption{Task specifications: physical system, dataset structure, and
HJN segmentation. Samples/segment $= N_{\mathrm{pred}}$, where $N_{\mathrm{pred}}$ denotes the
number of query times per training segment at which the Inverse
Network reconstructs. Each batch decodes every time point of every
segment. Tasks~3 and~4 use variable-length segments. On Task~4, 192 segments contain
57 samples, 48 contain 58 samples, and 48 contain 48 samples; on Task~3 the
counts differ because the sampling is non-uniform, and the shortest segment is
the remainder of the record. $\Delta t$ is the nominal sampling interval; Task~3 is
non-uniform with dominant step $\Delta t \approx 0.167$
($\Delta t = 0.167$ has 433 occurrences,
$\Delta t = 0.333$ has 8 occurrences,
$\Delta t = 0.5$ has 2 occurrences).}
\label{tab:hparams_task}
\setlength{\tabcolsep}{4pt}

\textbf{(a) System and dataset.}
\begin{tabular}{lcccccc}
\toprule
 & $n$ & Hamiltonian & Integrable & Physical constants & Train traj. & Samples/traj. \\
\midrule
Task~1 (mass--spring)   & 1 & yes & yes & $k=1$, $m=1$          & 24 & 32  \\
Task~2 (ideal pendulum) & 1 & yes & yes & $m=1$, $l=1$, $g=3$   & 25 & 30  \\
Task~3 (real pendulum)  & 1 & no  & n/a & unknown (measured)    & 1  & 444 \\
Task~4 (H\'enon--Heiles) & 2 & yes & no & $\lambda=1$ & 48 & 334 \\
\bottomrule
\end{tabular}

\vspace{0.6em}
\textbf{(b) HJN segmentation and training horizon.}
\begin{tabular}{lcccc}
\toprule
 & Train segments & Samples/segment $= N_{\mathrm{pred}}$ & $T_{\max}$ & $\Delta t$ \\
\midrule
Task~1 & 24 (full trajectory) & 32 & 6.493 & 0.209 \\
Task~2 & 25 (full trajectory) & 30 & 4.060 & 0.140 \\
Task~3 & 5 (within one traj.) & 90/92/94/96/72 & 15.973 & 0.167* \\
Task~4 & 288 (6 per traj.) & 57/58/48 & 8.563 & 0.15 \\
\bottomrule
\end{tabular}

\vspace{0.6em}
\textbf{(c) Test sets.}
\begin{tabular}{lccc}
\toprule
Test set & Trajectories & Samples/traj. & Time horizon \\
\midrule
Task~1 Test~A (rollout extrap.)                    & 24 & 97  & $20.106$ \\
Task~1 Test~B (energy extrap., $E\in[1.118,1.942]$) & 25 & 32  & $6.493$ \\
Task~2 Test~A (rollout extrap.)                    & 20 & 144 & $20.02$ \\
Task~2 Test~B (energy extrap., $E\in[2.4,4.43]$)    & 13 & 37  & $5.04$ \\
Task~3 (single trajectory tail)                     & 1  & 112 & $18.5$ \\
Task~4 Test~A (rollout extrap.)                    & 16 & 334 & $50.0$ \\
Task~4 Test~B (energy extrap., $E\in[0.13,\tfrac{1}{6})$) & 16 & 115  & $17.1$ \\
\bottomrule
\end{tabular}
\end{table}

Task~1 uses the analytic solution
$q(t)=A\sin(\omega t+\varphi_0)$, $p(t)=A\omega\cos(\omega t+\varphi_0)$,
with angular frequency $\omega=\sqrt{k/m}=1$, amplitude $A=\sqrt{2E}$. For Task~2 the ground-truth trajectories are generated with the
Dormand--Prince 8(5,3) adaptive integrator (DOP853) via
\texttt{scipy.integrate.solve\_ivp}~\cite{scipy_sw,virtanen2020scipy} with \texttt{atol}$=10^{-12}$ and
\texttt{rtol}$=10^{-9}$. Task~3 is
taken directly from~\cite{schmidt2009,schmidt2009data} without modification apart
from the train/test split. Task~4 trajectories are integrated with a compiled adaptive Taylor scheme
(heyoka) at the solver's default tolerance (near machine precision in double).
The measured relative energy drift stays at ${\sim}10^{-15}$ over the full
horizon; trajectories exceeding a $10^{-8}$ drift are rejected, and none were on
any split.

\paragraph{Initial-condition sampling.}
For the simulated tasks, initial conditions are drawn by stratifying over
the task's energy range: we place energy levels evenly across the band, and
for each level draw a position uniformly at random from the admissible region
at that energy. The momentum magnitude is then fixed by the energy constraint;
for the single-momentum Tasks~1--2 its sign is chosen uniformly where two roots
exist, while for the two-degree-of-freedom Task~4 its direction is drawn
uniformly on the circle---the $n=2$ analogue of that sign choice. In every task
a single energy-stratified pool is generated and then partitioned into
training, validation, and test sets by an energy-stratified split, so the three
sets are disjoint in trajectories while matched in energy distribution.

\subsection{HJN Architecture and Initialisation}
\label{app:init_details}

The backbone of $L=10$ stacked HJLs (Section~\ref{sec:hjl}) is shared
across all tasks. Each layer's potential functions $F$ and $G$
are MLPs with hidden widths $[33,\,60,\,32]$ and GELU activations on
hidden layers. The MLPs of the \emph{final} HJL (the $L$-th layer) use a linear output
layer. This localises the unconstrained output capacity to the final canonical-transformation step while keeping all intermediate HJLs fully nonlinear. Depending on the input dimensionality, $\mathcal{M}$ contains $82{,}560$ (Tasks~1--3) or $83{,}260$ (Task~4) trainable parameters. Because each
HJL has a closed-form algebraic inverse sharing all of its learnable
parameters, the Inverse Network
$\mathcal{M}^{-1}$ shares the same parameters exactly.

MLP weight matrices are initialised with the orthogonal scheme of
\citet{saxe2013dynamics}. The HJL scaling parameters are drawn from the
following Gaussians $\mathcal{N}(\text{mean},\;\text{variance})$:
\begin{align*}
c_{1,i},\,c'_{2,i} &\sim \mathcal{N}(0.1,\;0.01), &
c_{2,i},\,c'_{1,i} &\sim \mathcal{N}(-0.1,\;0.01), &
\gamma,\,\gamma' &\sim \mathcal{N}(2.0,\;0.04).
\end{align*}

The HJL scaling factor $a$ must remain non-zero for each layer to be
invertible (Appendix~\ref{sec:sm_proof}). To enforce this, we reparameterise $a$ as a bounded sigmoid of
an unbounded raw parameter $\alpha_{\mathrm{raw}}$:
\begin{equation}
a \;=\; \alpha_{\min} + (\alpha_{\max}-\alpha_{\min})\,\sigma(\alpha_k\cdot\alpha_{\mathrm{raw}}),
\end{equation}
with $[\alpha_{\min},\alpha_{\max}]=[0.5,\,2.0]$ and $\alpha_k=0.1$.
The raw parameter $\alpha_{\mathrm{raw}}$ is initialised so that the
induced distribution of $a$ has mean $1.6$ and standard deviation approximately $10\%$ of this mean, resulting in moderate variability across layers
while keeping $a$ bounded away from zero.

\subsection{Training Configuration}
\label{app:training_details}

Table~\ref{tab:hparams_train} summarises the per-task optimisation
settings, including the noise level, loss weights, and training duration.
\begin{table}[htbp]
\centering
\caption{Per-task training configuration. All models use AdamW; weight decay
$\lambda_{\mathrm{WD}}$ is applied only to MLP weights of $F,G$, with
zero weight decay on HJL scaling parameters. Segments per batch equal
the number of training segments; each batch samples one observation
per segment with additive Gaussian noise with
variance $\sigma^2$.}
\label{tab:hparams_train}
\setlength{\tabcolsep}{5pt}
\begin{tabular}{lcccccc}
\toprule
Task & Segments & $\sigma^2$ & $\lambda_{\mathrm{pred}}$ & $\lambda_{\mathrm{HSIC}}$ & $\lambda_{\mathrm{WD}}$ (MLP) & Epochs \\
\midrule
Task~1 (mass--spring)   & 24 & $0.1$ & $2.0$ & $0.10$ & $10^{-4}$ & $507$  \\
Task~2 (ideal pendulum) & 25 & $0.1$ & $2.0$ & $0.35$ & $10^{-6}$ & $1157$ \\
Task~3 (real pendulum)  & 5 & $0$   & $2.0$ & $0.05$    & $10^{-4}$ & $2202$ \\
Task~4 (H\'enon--Heiles) & 288 & $0.0045$ & $2.0$ & $0.35$ & $10^{-6}$ & $2111$ \\
\bottomrule
\end{tabular}
\end{table}

\paragraph{Noise selection.}
For Tasks~1 and~2, we adopt $\sigma^2=0.1$, following the value used
by~\citet{Greydanus2019} for the same mass--spring and pendulum systems,
so that our synthetic-noise regime matches established work. Task~3 uses
the measured trajectory directly and receives no injected noise, since
the real pendulum data already contain measurement noise. For Task~4, no reference noise level is available, and directly reusing
$\sigma^2=0.1$ would be inappropriate. The H\'enon--Heiles state
variables have substantially smaller within-segment variances
(${\approx}0.045$--$0.056$) than the mass--spring and pendulum
coordinates (${\approx}0.57$--$1.70$). These within-segment variances are
nearly identical to the corresponding pooled variances in all tasks, so
the choice is insensitive to this distinction. Under $\sigma^2=0.1$, the
per-channel signal-to-noise ratios (SNRs), defined as the within-segment
signal variance divided by $\sigma^2$, would be far below the range
validated on Tasks~1--2. We consequently select the Task~4 noise level by matching this SNR range.
Since a single shared noise variance most strongly degrades the
lowest-variance channel, we match the $p_y$ channel of the
H\'enon--Heiles system (variance ${\approx}0.045$) to the Task~1--2 SNR
range of $[5.7,\,17.0]$. This yields
$\sigma^2\in[2.7\times10^{-3},\,7.9\times10^{-3}]$. We choose
$\sigma^2=0.0045$ ($\sigma\approx0.067$), which lies near the geometric
mean of this interval. This choice provides a margin against input noise
while avoiding excessive corruption that could be amplified by the
chaotic dynamics over the prediction horizon.

\paragraph{HSIC estimation.}
Per-batch HSIC estimators use the unbiased form (Algorithm~\ref{alg:training}) with RBF kernels and median-heuristic bandwidths computed independently
per batch (Phase~II, step~3 of Algorithm~\ref{alg:training}). The
prediction-loss scale $\bar{\mathcal{L}}_{\mathrm{pred}}$ is estimated in a
single forward-only epoch prior to training (Phase~I of
Algorithm~\ref{alg:training}) and held fixed thereafter. The HSIC
normalisation $H_{\max}$ is instead recomputed per batch from a
maximally correlated reference signal of length equal to the number of
segments in that batch, $\mathbf{s}=[1,2,\dots,N_b]$, as
$H_{\max}=\max\!\bigl(\mathrm{HSIC}(\mathbf{s},\mathbf{s}),\,0.05\bigr)$;
when segments have unequal length, $N_b$ varies across batches within an
epoch.

\paragraph{Validation Criterion.}
\label{app:validation_criterion}
The epoch budgets reported in Table~\ref{tab:hparams_train} were selected
based on a validation prediction loss of the seed-42 run for Tasks~1--2 and~4. For Task~3,
the single available trajectory does not allow a separate validation
split; therefore, the stopping epoch was selected using the test
trajectory. The same procedure is applied to the reference, so the Task~3 errors of both models are optimistically biased. Task~3 results should therefore be read for the behaviour they illustrate
outside the theory's scope (Section~\ref{sec:prediction_accuracy}), not as
estimates of generalisation performance. For Tasks~1,~2, and~4, each validation trajectory matches its
corresponding training trajectory in sampling interval $\Delta t$,
time horizon and number of samples. The validation trajectories are drawn from the same energy distribution as
the training trajectories and are segmented using the task-specific
$T_{\max}$ values established from the training data. The
validation prediction loss is computed as follows: the model receives
the first phase-space point of each validation segment and predicts the
remaining points in the segment. We compute the mean absolute error
(MAE) between the predicted and ground-truth phase-space coordinates
and average this error across all validation segments.

\subsection{Reference model configuration}
\label{app:baseline_details}
The reference, Noether's Razor~\cite{van2024noether}, a symmetry-learning
Hamiltonian neural network, was trained using the authors' official
implementation~\cite{nr_code}. Unless otherwise stated, hyperparameters were chosen to match the
simple harmonic oscillator configuration reported in the original
paper and aligned with the released implementation.

\paragraph{Training configuration inherited from the original paper.}
Following the simple-harmonic-oscillator configuration reported by the authors, the reference used a variational Bayesian Hamiltonian neural network with a $3$-layer MLP Hamiltonian of hidden width $200$ and ELU activations ($\alpha=2$). As with HJN, a single backbone is shared across all tasks. At inference the model uses $41{,}001$ trainable parameters, the variational-posterior means of the Hamiltonian network. Training additionally optimises matrix-normal covariance factors and prior/likelihood hyperparameters, raising the total trained count to ${\approx}121{,}000$. Because the posterior collapses to its mean at test time, none of these covariance or hyperparameter values enter prediction

Symmetrisation used $S=200$ Monte Carlo samples drawn from a uniform
measure $\mu(\tau)$, together with $20$ Euler integration steps and
$20$ symmetry-integration steps. The model used first-order BCH
composition and $M=2$ variational weight samples per optimisation
step. Output variance estimation was fixed, while prior precision was
estimated using the closed-form empirical update derived in
their Appendix~D.3.

The defining contribution
of~\cite{van2024noether} is that the correct number of
conserved quantities need not be specified in advance: the method is
given a surplus of $K$ candidate conserved quantities and relies on the
marginal-likelihood (Occam's razor) objective to retain only those
supported by the data while suppressing the rest. Fixing $K$ to the
true number of conserved quantities of each system would therefore
supply the reference with oracle knowledge that is unavailable in any
realistic setting, and would bypass the very mechanism the method is
designed to exercise. We instead fix $K=3$ across all tasks, exceeding
the number of independent quadratic conserved quantities in every
system, so that the method operates in the over-parameterised regime
for which it was proposed. As a diagnostic, we apply the activity criterion of the generator
diagnostic of~\citet[\S5.1]{van2024noether} at the within scope Tasks~1~2 and ~4, stacking the $K$ learned generators
$G_k = J\nabla C_k$ on homogeneous coordinates and taking the singular values of the resulting
matrix. On all three tasks all $K=3$ singular values exceed the $\lambda>0.05$ threshold used by the authors by more than an order of magnitude, the smallest being $0.80$ (Task~1), $0.66$ (Task~2) and $0.82$
(Task~4); as fractions of the largest singular value of the same task these are $0.31$, $0.25$ and
$0.23$, so the spectra are far from rank-deficient on either an absolute or a scale-free reading. All $K=3$ candidate conserved quantity have been retained to these 3 tasks.

\paragraph{Additional training settings inherited from the released implementation.}
Several optimisation settings used by the released implementation are
not explicitly discussed in the experimental section of the paper but
were retained unchanged in our reproduction from the provided codebase. These include:
The three parameter groups used distinct learning rates: $10^{-3}$ for
the variational parameters of the HNN, $10^{-1}$ for the prior precision
and likelihood (output) variance, and $10^{-2}$ for the
conserved-quantity parameters. A KL warm-up linearly ramps the weight of the KL term from $0$ to $1$
over the first $20$ epochs. The prior variance was initialised to $1.0$
and the posterior scale to $10.0$, with unconstrained basis learning and
MAP-gradient mixing disabled throughout optimisation. Training-time symmetrisation used the path-based formulation of the
released implementation. Variational inference was enabled in all
experiments, cosine learning-rate
annealing, and Adam ($\beta_1=0.9$, $\beta_2=0.999$). On Tasks~1--3 the reference was trained full-batch, matching the released
configuration. For Task~4 this is infeasible: the ${\sim}16{,}000$ training
transition pairs exceed the $12$\,GB of the evaluation GPU.
We therefore minibatch Task~4 at $B{=}3$ trajectories (${\approx}1000$
consecutive pairs per batch). This matches the batch scale the original
authors themselves adopt for their larger $n$-oscillator and $n$-body systems,
for which they likewise depart from full-batch training.

\paragraph{Task-specific modifications aligning the reference with our data and protocol.}
Modifications to the reference's configuration were intentionally minimal
and restricted to quantities required to align the method with the
datasets and training protocol used for HJN. The integration step size
was matched to the nominal sampling interval of each dataset:
$\Delta t=0.209$ (Task~1),
$\Delta t=0.140$ (Task~2),
$\Delta t=0.167$ (Task~3) and
$\Delta t=0.15$ (Task~4).

Additive Gaussian noise with variances $\sigma^2=0.1$ for Tasks~1--2 and $\sigma^2=4.5\times10^{-3}$ for task 4 was injected into the model inputs during training,
while prediction targets remained noise-free, matching the
input-noise model used for HJN. The two training loops differ in
batch construction, so per-epoch noise exposure is not identical;
the per-step noise distribution is. For Task~3, no input noise was added and
the fixed output variance was reduced to $10^{-3}$ in order to reflect
the clean observational setting used for HJN.

Following their reported training schedule, training was run for $2000$ epochs in all experiments. The original work appears to run this
fixed-length schedule and use the model parameters obtained at the
final epoch. We instead apply the same model-selection procedure used
for HJN: at every epoch we evaluate the validation criterion of
Appendix~\ref{app:validation_criterion} and retain the parameters from the epoch with the lowest validation criterion. Both methods therefore
report the parameters selected by the identical validation criterion, rather than those of the final epoch of a fixed schedule.

To check that the trained reference model reached an optimisation regime
comparable to the original work, we monitored the same one-step
prediction error reported in Table~1 of their paper. For
every consecutive pair of observed states $(x_t, x_{t+\Delta t})$ in a
dataset (where $x_t = (q_t, p_t)$), the learned dynamics are integrated forward by one sampling
interval from $x_t$, and the squared error against the true
$x_{t+\Delta t}$ is averaged over all such pairs and all trajectories
in the set. The original paper reports this quantity on both their
training and held-out test sets, approximately $2\times10^{-3}$ on both. We compute
it on our validation set (and the held-out test trajectory for Task~3). The Task~1 model achieved a one-step validation MSE of
$5.75\times10^{-4}$. The two values are not directly comparable, as
they use different data, sampling, and evaluation splits, but their
agreement in scale is consistent with our reproduction having reached a
comparable one-step optimisation regime. The corresponding one-step
validation MSE was $1.97\times10^{-3}$ for Task~2,
$8.2\times10^{-5}$ for Task~3 and $1.7\times10^{-4}$ for Task~4.

\paragraph{Evaluation configuration.}
At evaluation, and following the released implementation, models with quadratic conserved quantities dispatch directly to the matrix-exponential symmetrisation, irrespective of the training-time symmetrisation setting. Consequently, all predictions reported for the reference use closed-form matrix-exponential symmetrisation during
inference, with $S=200$ Monte
Carlo samples per gradient evaluation. Trajectory prediction was implemented to match
the released pipeline as closely as possible: inference
used fixed-step Euler integration with $20$ substeps between
consecutive output samples, predictions used posterior-mean weights,
and the symmetrisation random seed was fixed for deterministic
evaluation.

The released trajectory-prediction function rolls the learned dynamics
forward from the first observation of a trajectory to a set of output
times that all lie after it; it does not support prediction at output
times that lie before the observation, nor does it accept the
observation at an arbitrary index of the trajectory rather than the
first. The evaluation protocol used for HJN
(Section~\ref{sec:eval_protocol}) requires both: for each observation
index $m \in \{1,\dots,N_{\mathrm{obs}}\}$, the observed state
$z_{\mathrm{obs}} = (x_{t_m}, t_m)$ is used to predict the trajectory at
all remaining indices, which generally include both later and earlier
times. To enable evaluation under the same protocol as our method, we extended the released inference pipeline accordingly, with
no modification to the trained model. For times $t_j > t_m$,
predictions are obtained by integrating $\dot{x} = J\nabla H_\theta(x)$
forward from $x_{t_m}$ using the same integrator configuration as the
released code. For times $t_j < t_m$, predictions are obtained via
time-reversibility: integrating $\dot{x} = J\nabla H_\theta(x)$ backward
in time is mathematically identical to integrating $\dot{x} = -J\nabla
H_\theta(x)$ forward in pseudo-time $\tau = t_m - t$, exploited here so
that the integrator's direction remains positive while recovering
exact backward predictions. The learned $H_\theta$ is parameterised as
an MLP of phase-space coordinates alone with no explicit time
dependence, so this equivalence holds for the architectures in our
experiments. The Hamiltonian parameters, posterior-mean weights,
integrator step size, symmetrisation Monte Carlo, and random key are
identical between the forward and backward integration phases.

\section{Supplementary Figures and Tables}
\label{app:supplementary_figures}
\begin{figure}[htbp]
\centering
\includegraphics[width=\textwidth]{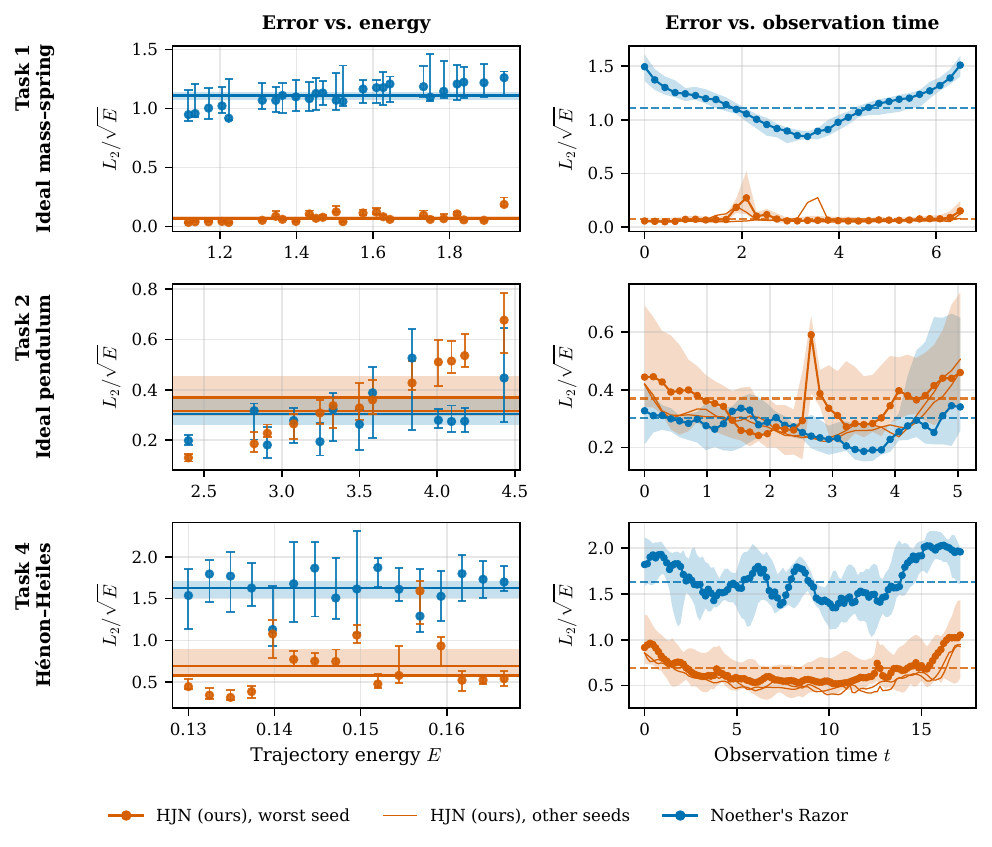}
\caption{Per-task accuracy structure on the Test~B
(energy-extrapolation) sets, HJN
(\textcolor[HTML]{D55E00}{orange}), Noether’s Razor (\textcolor[HTML]{0072B2}{blue}). Rows: Tasks~1, 2 and~4. Columns, left to right: (a)~$\sqrt{E}$-normalised
reconstruction error vs.\ trajectory energy, where each marker is a
trajectory's median error across observation choices, error bars give
the across-observation interquartile range (IQR), the solid horizontal line marks the
cross-trajectory mean of these per-trajectory medians, and the shaded
band is its $95\%$ bias-corrected and accelerated (BCa) bootstrap confidence interval (CI) ($10{,}000$ resamples); (b)~error
vs.\ observation time, with the median across trajectories (solid),
the across-trajectory IQR (shaded band), and the cross-trajectory mean
of medians (dashed). HJN is shown for the seed with the
largest cross-trajectory mean (seed~52 on Tasks~1--2, seed~62 on Task~4); thin
orange lines mark the other two seeds' means in (a) and their medians across
trajectories in (b).}
\label{fig:main_results_testB}
\end{figure}

Figure~\ref{fig:main_results_testB} replicates the format of
Figure~\ref{fig:summary_statistics} for Test~B of Tasks~1, ~2 and~4.

\begin{figure}[htbp]
\centering
\includegraphics[width=\textwidth]{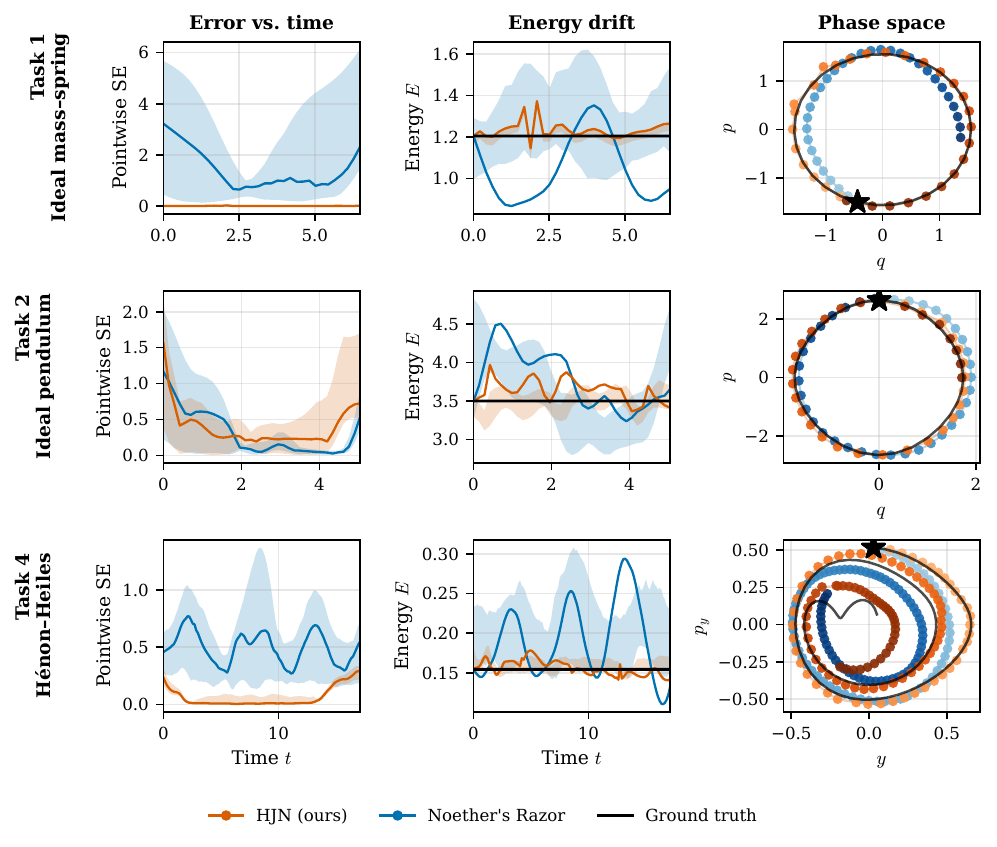}

\caption{Single-trajectory inference diagnostics for one representative
trajectory per task (Tasks~1--2 and 4: Test~B), HJN's seed-42 model (\textcolor[HTML]{D55E00}{orange}), Noether’s Razor (\textcolor[HTML]{0072B2}{blue}), ground truth (black). Inference is run from every observation point; the displayed energy and
phase-space curves use the run initialised from the first observation
(obs.\ index~$0$), the longest rollout and hence the most drift-prone.
Columns, left to right: (a)~pointwise squared error vs.\ time, with the median (solid)
and interquartile range (IQR; shaded) across observation choices; (b)~predicted energy
$E(\hat q,\hat p)$ for the obs-$0$ run (solid) with the across-observation
IQR band (shaded), against the ground-truth energy (black) -- conserved
on Tasks~1--2 and~4; (c)~phase-space orbit for
the obs-$0$ run, colour-graded light$\to$dark with time, with the start
($t{=}0$, the conditioning observation) marked by a star. For Task~4 the panel shows the $(y,p_y)$ projection of the four-dimensional phase space.}
\label{fig:phasespace}
\end{figure}

Similarly, Figure~\ref{fig:phasespace} replicates the format of
Figure~\ref{fig:obs_sensitivity} for the energy extrapolation test set of Tasks~1, ~2 and~4.

\section{Per-task symplecticity diagnostics}
\label{app:symplecticity_diagnostic}
The architectural proofs in Appendix~\ref{sec:sm_proof} guarantee
that $\mathcal{M}$ and $\mathcal{M}^{-1}$ are canonical transformations
for any real-valued parameters and $C^{2}$ activations, under exact
arithmetic. Finite-precision evaluation introduces small numerical
residuals and this experiment bounds them. For the representative test
trajectory per task (Tasks~1~2 and ~4 : Test~A; Task~3: its held-out tail), we partition the trajectory into segments of
length $T_{\max}$ and, at every observation $(q, p, t)$, compute the
Jacobian
$M = \partial(Q, P) / \partial(q, p)\big|_{\,t - t_{\mathrm{start}}}$
of $\mathcal{M}$ at the segment-relative time, consistent with the
inference convention of Section~\ref{sec:autoregressive}. The
Jacobian is obtained via
\texttt{torch.autograd.functional.jacobian}, and we evaluate the
symplectic residual $\|R\|_{\max}$, where
$R := M^{\!\top}\Omega\,M - \Omega$, 
$\Omega = \bigl(\begin{smallmatrix} 0 & I_n \\ -I_n & 0 \end{smallmatrix}\bigr)$,
and $\|\cdot\|_{\max}$ denotes the entrywise maximum
$\|R\|_{\max} := \max_{ij}|R_{ij}|$.
The transformation is exactly canonical if $R = 0$, so $\|R\|_{\max}$
measures the worst-case violation of the symplectic condition at a point. For the Inverse Network the
same residual is evaluated on the Jacobian
$\tilde{M} = \partial(\hat{q}, \hat{p}) / \partial(Q, P)
\big|_{\,t - t_{\mathrm{start}}}$ of $\mathcal{M}^{-1}$,
taken at the corresponding encoder output $(Q, P)$ and the same
segment-relative time.  Table~\ref{tab:symplectic_residuals} reports, per task, the maximum
and median of this residual across all observations of the selected
trajectory, for both networks. All residuals fall near the
single-precision floor, confirming that the constraint holds to
numerical accuracy.

\begin{table}[htbp]
\centering
\caption{Per-point symplectic residual $\|R\|_{\max}$
(with $R = M^{\!\top}\Omega M - \Omega$) computed in single precision
via \texttt{torch.autograd.functional.jacobian} on one representative
test trajectory per task; we report the maximum and median over all
observations of that trajectory.}
\label{tab:symplectic_residuals}
\begin{tabular}{lcccc}
\toprule
& \multicolumn{2}{c}{$\mathcal{M}$} & \multicolumn{2}{c}{$\mathcal{M}^{-1}$} \\
\cmidrule(lr){2-3} \cmidrule(lr){4-5}
Task & max & median & max & median \\
\midrule
Task~1 & $4.77 \times 10^{-7}$ & $1.19 \times 10^{-7}$ & $9.54 \times 10^{-7}$ & $1.19 \times 10^{-7}$ \\
Task~2 & $5.96 \times 10^{-7}$ & $1.19 \times 10^{-7}$ & $4.77 \times 10^{-7}$ & $1.19 \times 10^{-7}$ \\
Task~3 & $9.54 \times 10^{-7}$ & $1.79 \times 10^{-7}$ & $6.44 \times 10^{-7}$ & $1.19 \times 10^{-7}$ \\
Task~4 & $1.25 \times 10^{-6}$ & $2.38 \times 10^{-7}$ & $5.25 \times 10^{-6}$ & $9.54 \times 10^{-7}$ \\
\bottomrule
\end{tabular}
\end{table}

% =====================================================================
%  Appendix: HSIC frame-sweep experiment
% =====================================================================

\section{HSIC motivation}
\label{app:hsic_frame_sweep}
Next we motivate the adoption of the Hilbert--Schmidt Independence Criterion
penalty $\mathcal{L}_{\mathrm{HSIC}}$ on the encoder's latent pairs
$(Q^{(i)}, P^{(i)}), i \in \{1,\ldots,n\}$ during training, by exhibiting the failure mode it is designed to prevent, and by showing that the penalty is non-zero on every frame in the sweep that exhibits it. The failure mode is a chain of three links:
\begin{enumerate}
\item a frame in which $\mathcal{M}_S$ is componentwise ill-conditioned is a frame
      in which $(Q^{(i)}, P^{(i)})$ are statistically dependent, so
      $\mathcal{L}_{\mathrm{HSIC}}$ is non-zero on it;
\item a componentwise ill-conditioned map amplifies single-precision rounding error
      in the latents it produces, and the autoregressive handoff protocol carries
      that error from one segment to the next;
\item the resulting error can degrade prediction accuracy substantially in practice.
\end{enumerate}
We do not claim that the model reported in this paper would
have failed without the penalty; we establish that the mechanism the penalty
targets is real and consequential. 

We run this experiment on a pool of $N_{\mathrm{traj}} = 80$ H\'enon--Heiles
trajectories, drawn from the same energy distribution and integrated at the same
sampling rate as the training set, sharing a common grid of $172$ samples over
$t \in [0, 25.689]$. Link~3 is measured through rollout error, which is
informative only where the rollout has not already diverged: on a diverged
rollout the error is saturated at the attractor scale before any frame is
applied, and no frame-induced increase can be resolved. We therefore evaluate the converged model at $3T_{\max}$, where its double-precision
rollout MAE is $0.183$ (mean over the pool) against a mean state magnitude of
$0.777$: far enough below the attractor scale that a frame-induced increase is
resolvable against it. The cancellation \eqref{eq:frame_cancel} is exact at any
parameter setting, so no property of this particular checkpoint is required for
the contrast; every sweep point is compared against $S=\mathrm{id}$ on the same
frozen weights.

We encode the $t = 0$ state $z_j$ of each trajectory,
\begin{equation}
    (Q_j,\,P_j) = \mathcal{M}(z_j,\,0) \in \mathbb{R}^{n}\times\mathbb{R}^{n},
    \qquad j=1,\dots,80 .
    \label{eq:encode_grid_app}
\end{equation}

Let $S$ be a time-independent canonical map on the latent. We call such an $S$ a
frame, and write $\mathcal{M}_S = S\circ\mathcal{M}$ for the framed encoder and
$(Q^S_j, P^S_j) = S(Q_j, P_j)$ for the framed latents. To expose the link between
statistical dependence and ill-conditioning of $\mathcal{M}_S$, we use three families
of frames, each introducing a different type of dependence between the pairs
$(Q^{S,(i)}, P^{S,(i)})$, and each swept so that the dependence strengthens
progressively along the family:

\begin{itemize}
\item \textbf{Squeeze}, $S:(Q^S_j,P^S_j)=(aQ_j,\,P_j/a)$, with
      $a \in \{1, 10, 10^2, 10^3, 10^4, 10^5\}$. A pure change of per-coordinate
      units, introducing no dependence between the pair at any $a$; it is the
      null arm. The upper limit $a=10^5$ was set
      empirically: it leaves $\min_{i,j} |P_j^{(i)}/a| \approx 1.7\times10^{-6}$, far
      above the single-precision subnormal range, so the arm is not
      contaminated by underflow.

\item \textbf{Shear}, $S:(Q^S_j, P^S_j)=(Q_j + \lambda P_j,\,P_j)$, with
      $\lambda = \tilde\lambda \cdot \mathrm{med}|Q_j|/\mathrm{med}|P_j|$ and
      $\tilde\lambda \in \{0, 0.1, 1, 10, 10^2, 10^3, 10^4\}$. Because a shear
      mixes coordinates with momenta, $\lambda$ carries units of $Q_j/P$; we
      therefore scale it by the measured latent aspect ratio
      $\mathrm{med}|Q_j|/\mathrm{med}|P_j| = 2.374$ so that $\tilde\lambda$ is
      dimensionless and $\tilde\lambda \approx 1$ marks the onset of genuine mixing regardless of the trained latent's scale. The map is symplectic for every $\lambda$, as is the squeeze for every $a \neq 0$.

\item \textbf{Rotation at fixed anisotropy}, $S = R_\theta D_{a_0}$, a squeeze by
      $a_0$ followed by a rotation of each conjugate pair by $\theta$:
      \begin{equation}
        D_{a_0} = \begin{pmatrix} a_0 I_n & 0 \\ 0 & a_0^{-1} I_n \end{pmatrix},
        \qquad
        R_\theta = \begin{pmatrix} \cos\theta\, I_n & \sin\theta\, I_n \\
                                  -\sin\theta\, I_n & \cos\theta\, I_n \end{pmatrix},
        \label{eq:rot_frame}
      \end{equation}
      so that $(Q^S_j, P^S_j) = R_\theta\,(a_0 Q_j,\; P_j/a_0)$. Both factors
      satisfy $X^{\!\top}\Omega X = \Omega$, so $S$ is canonical. Write
      $(u, v) = (a_0 Q_j,\; P_j/a_0)$ for the squeezed pair, so that the frame
      stores $u\cos\theta + v\sin\theta$ and $-u\sin\theta + v\cos\theta$.

      Two properties follow directly from \eqref{eq:rot_frame}. First,
      $R_\theta$ is orthogonal, $R_\theta^{\!\top}R_\theta = I_{2n}$, hence
      \begin{equation}
        M_S^{\!\top} M_S
        = (D_{a_0}M)^{\!\top} R_\theta^{\!\top} R_\theta (D_{a_0}M)
        = (D_{a_0}M)^{\!\top} (D_{a_0}M),
      \end{equation}
      so the singular values of $M_S$ are those of $D_{a_0}M$ for every
      $\theta$: the spectral condition number is pinned along the family, and
      only $\theta$ varies. Second, the decoder recovers the smaller component
      as $v = (u\cos\theta + v\sin\theta)\sin\theta +
      (-u\sin\theta + v\cos\theta)\cos\theta$, in which two contributions from
      $u$, each of magnitude $|u|\sin\theta\cos\theta$, cancel exactly. A
      relative perturbation of the stored pair is therefore amplified in the
      recovered $v$ by
      \begin{equation}
        \frac{|u|\sin\theta\cos\theta}{|v|}
        = a_0^{2}\,\frac{|Q_j|}{|P_j|}\,\sin\theta\cos\theta .
        \label{eq:cancel_ratio}
      \end{equation}

      The sweep is designed around \eqref{eq:cancel_ratio}. Since $P_j = a_0 v$, the
      absolute error the base decoder receives in $P_j$ carries the same $a_0^2$
      factor as the relative error in $v$, so $e_{\mathrm{num}}$ is predicted to
      scale as $a_0^2$ at fixed $\theta$; this is why we run
      $a_0 \in \{10^2, 10^3, 10^4\}$, the quadratic scaling being a test of the
      cancellation mechanism itself and not merely of its consequences. The factor $\sin\theta\cos\theta$,
      which is $\approx\tan\theta$ for small $\theta$ and vanishes at both
      $\theta = 0$ and $\theta = 90^\circ$, is why the grid is logarithmic in
      $\tan\theta$ ---
      $\tan\theta \in \{0, 10^{-6}, 3\!\times\!10^{-6}, 10^{-5},
      3\!\times\!10^{-5}, 10^{-4}, 10^{-3}, 10^{-2}, 10^{-1}, 1\}$, together
      with $\theta = 90^\circ$ --- rather than uniform in degrees: the stored
      component $-u\sin\theta + v\cos\theta$ becomes $u$-dominated, and the pair
      correspondingly dependent, once $\tan\theta \gtrsim a_0^{-2}|P_j|/|Q_j|$,
      so a grid of round angles such as $\{15^\circ,\dots,75^\circ\}$ would lie
      entirely past that onset and resolve the arm into two values. The point
      $\theta = 90^\circ$ is a signed exchange of the pair, for which
      \eqref{eq:cancel_ratio} vanishes; it returns the arm to its baseline and
      so distinguishes genuine cancellation from any monotone artefact of
      increasing $\theta$. We confirm the first property numerically: within
      each rotation arm the spectral condition number is constant to a relative
      spread below $1.1\times10^{-15}$.
\end{itemize}

Together the three families give
$6 + 7 + 3\times 11 = 46$ frames, of which $42$ are
distinct: the arms share their baselines, since $\tilde\lambda = 0$ is the identity
and $\theta = 0^\circ$ at each $a_0$ is the corresponding squeeze. The invariance \eqref{eq:frame_cancel} holds for any $t$-independent bijection
$S$, so restricting to linear $S$ is a choice of experiment, not a requirement of
the argument. It is made because the experiment turns on two quantities being
pinned exactly, and for linear $S$ each is an algebraic identity rather than a
measured property: a squeeze acts on the Jacobian as $M_S = DM$ with $D$
diagonal, and $\mathrm{Skeel}(DM) = \mathrm{Skeel}(M)$ holds identically; a
rotation is orthogonal, so within a rotation arm $M_S$ has the same singular values as $D_{a_0}M$. Under a
nonlinear $S$ neither is available a priori, and whether an arm held its control
fixed would itself become an empirical claim.

For each frame we evaluate $\mathcal{L}_{\mathrm{HSIC}}$ on the $80$ framed
latents $(Q^S_j, P^S_j)$ using exactly the training estimator: unbiased, RBF
kernels, per-DOF median-heuristic bandwidths detached from the graph, zeroed
kernel diagonals, averaged over the $n$ degrees of freedom, and clamped below at
zero (Algorithm~\ref{alg:training}). We evaluate it in double
precision so that the measured value reflects the
frame rather than itself being a casualty of the rounding errors under study. Importantly, the
median heuristic rescales each bandwidth with the data, so HSIC cannot rise
merely because a frame inflates the latent's scale.

To measure the conditioning of the resulting $\mathcal{M}_S$ map we compute the $2n \times 2n$ Jacobian $M = \partial(Q,P)/\partial(q,p)$ of the
base encoder at $t=0$, at each of the $80$ states, in double precision via
\texttt{torch.autograd.functional.jacobian}. Since $S$ is linear and
time-independent, the framed Jacobian is exactly $M_S = SM$, so all $46$ frames
follow from this single differentiation.

We report medians over the pool of
\begin{equation}
\kappa \;=\; \sigma_{\max}(M_S)^2,
\qquad
\mathrm{Skeel}(M_S) \;=\; \bigl\|\,|M_S^{-1}|\,|M_S|\,\bigr\|_\infty ,
\end{equation}
where $|\cdot|$ is entrywise. The square in $\kappa$ reflects the fact that
symplecticity forces the singular values into reciprocal pairs
$(\sigma, 1/\sigma)$ which we verify numerically. The Skeel condition number \citep{Skeel1979ScalingFN} is the
standard \emph{componentwise} conditioning measure: unlike $\kappa$ it is invariant under rescaling of the latent components and sensitive to rotation, matching the invariances the failure mode exhibits here.

As established in Appendix~\ref{app:frame_invariance}, the framed model has the same
prediction error as the base map, $\mathcal{L}_{\mathrm{pred}}(\mathcal{M}_S)
=\mathcal{L}_{\mathrm{pred}}(\mathcal{M})$, for every $S$ at fixed $\mathcal{M}$. Any
observed violation of this identity is therefore attributable to finite-precision
error alone. To measure the errors we run each frame's rollout twice, once in
single and once in double precision, and define
\begin{equation}
\label{eq:e_num}
e_{\mathrm{num}} \;=\; \mathrm{MAE}\bigl(\hat z^{\,\mathrm{f32}},\;
\hat z^{\,\mathrm{f64}}\bigr),
\end{equation}
the discrepancy between the two precisions of the \emph{same} wrapped model,
using the same MAE definition as the evaluation loss. This quantity isolates numerical error amplification and is independent of the model's own approximation error. Every
rollout is conditioned on the first observation index, which maximises the horizon
and therefore the number of encodings ($3$ over the three segments, two of them
handoffs) through which a latent perturbation can compound. We verify numerically
that at $S=\mathrm{id}$ the wrapped rollout is bit-identical to the unwrapped
rollout, confirming that the wrapper introduces no arithmetic of its own.

To show that these floating errors can matter in practice, we report the single-precision rollout MAE ratio
\begin{equation}
\label{eq:r32}
r_{32} \;=\;
\frac{\mathrm{MAE}_{32}(\text{frame})}{\mathrm{MAE}_{32}(\text{arm baseline})},
\end{equation}
aggregated as a ratio of means over the $80$ trajectories, with arm baselines
$a=1$, $\tilde\lambda = 0$, and $\theta = 0^\circ$ at the same $a_0$, respectively. Unlike
$e_{\mathrm{num}}$, this is measured against ground truth and therefore answers
whether the numerical error amplification is large enough to matter relative to the model's
own error. As a control we measure $|r_{64} - 1| \le 3.5\times10^{-8}$ over all $46$ frames: in
double precision every frame is inert to eight digits. This establishes that the entire spread of single-precision outcomes is a numerical phenomenon. Widening the format postpones the onset rather than removing it, since the amplification \eqref{eq:cancel_ratio} is unbounded in $a_0$.

\begin{figure}[H]
\centering
\includegraphics[width=\textwidth]{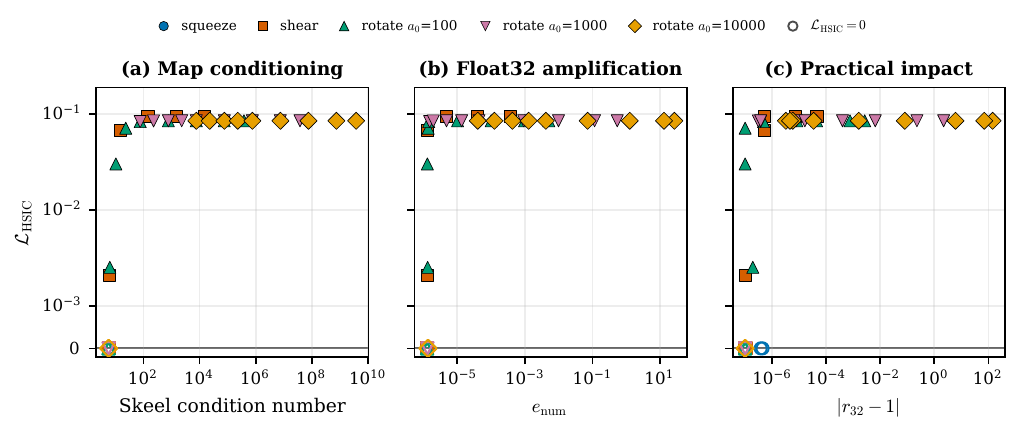}
\caption{$\mathcal{L}_{\mathrm{HSIC}}$ against each link of the chain, over all
$46$ frames. (a)~map conditioning, (b)~single-precision amplification $e_{\mathrm{num}}$ (Eq.~\eqref{eq:e_num}),
(c)~practical impact on the rollout, $|r_{32}-1|$ with $r_{32}$ from Eq.~\eqref{eq:r32}. Marker shape and colour give the frame
family. Filled markers are the $31$ frames on which the penalty is non-zero;
hollow markers the $15$ on which it vanishes exactly. The shared $y$-axis is
symmetric-log below $10^{-3}$ so that those zeros remain resolvable from the
smallest non-zero value ($2.1\times10^{-3}$); $x$-axes are logarithmic, and in
(c) values below $10^{-7}$ are clipped to the left edge. No hollow marker lies
to the right of its panel's harm threshold.}
\label{fig:hsic_frames}
\end{figure}

We characterise a frame as harmful at link~1 if $\mathrm{Skeel}(M_S)$ exceeds ten times
its value at $S = \mathrm{id}$, at link~2 if $e_{\mathrm{num}}$ exceeds twice its
value at $S = \mathrm{id}$, and at link~3 if $|r_{32} - 1| > 10^{-2}$. The
identity frame is the reference for the first two because it is the frame in
which no reparameterisation has been applied, so its $e_{\mathrm{num}}$
($1.28\times10^{-6}$) is the single-precision floor of this model on this pool. None of the three thresholds is derived from $\mathcal{L}_{\mathrm{HSIC}}$, and
the zero-miss result reported below does not depend on where they are set: no
frame on which the penalty vanishes exceeds $1.02$ times the identity frame's
Skeel number, $1.07$ times its $e_{\mathrm{num}}$, or $4.4\times10^{-7}$ in
$|r_{32}-1|$, so all three could be tightened to those values, the link-3 one by
four decades, without a single harmful frame escaping detection. Table~\ref{tab:hsic_frames} reports eight representative frames and
Fig.~\ref{fig:hsic_frames} the full sweep.

\begin{table}[H]
\centering
\footnotesize
\caption{Representative frames from the $46$-point sweep. Rows~1--2 hold the
squeeze arm's extremes; rows~3--8 lie within the single $a_0 = 10^4$ rotation
arm, where $\kappa$ is constant to every digit shown. $r_{64}$ is the
double-precision control. All quantities are medians (conditioning) or means
(errors) over $80$ trajectories.}
\label{tab:hsic_frames}
\setlength{\tabcolsep}{5pt}
\begin{tabular}{llccccc}
\toprule
Frame & Parameter & $\kappa$ & Skeel & $\mathcal{L}_{\mathrm{HSIC}}$
& $e_{\mathrm{num}}$ & $r_{32}$ \\
\midrule
squeeze            & $a=1$                    & $2.781\times10^{1}$  & $5.674\times10^{0}$ & $0.0000$ & $1.28\times10^{-6}$ & $1.000$ \\
squeeze            & $a=10^5$                 & $2.774\times10^{11}$ & $5.674\times10^{0}$ & $0.0000$ & $1.26\times10^{-6}$ & $1.000$ \\
\midrule
rotate $a_0{=}10^4$ & $\theta=0^\circ$                 & $2.774\times10^{9}$ & $5.674\times10^{0}$ & $0.0000$ & $1.36\times10^{-6}$ & $1.000$ \\
rotate $a_0{=}10^4$ & $\theta=5.7\times10^{-5}\,^\circ$ & $2.774\times10^{9}$ & $7.610\times10^{3}$ & $0.0849$ & $4.04\times10^{-5}$ & $1.000$ \\
rotate $a_0{=}10^4$ & $\theta=5.7\times10^{-3}\,^\circ$ & $2.774\times10^{9}$ & $7.605\times10^{5}$ & $0.0849$ & $4.17\times10^{-3}$ & $1.002$ \\
rotate $a_0{=}10^4$ & $\theta=0.573^\circ$              & $2.774\times10^{9}$ & $7.605\times10^{7}$ & $0.0849$ & $1.26\times10^{0}$  & $7.083$ \\
rotate $a_0{=}10^4$ & $\theta=5.71^\circ$               & $2.774\times10^{9}$ & $7.530\times10^{8}$ & $0.0849$ & $2.64\times10^{1}$  & $\mathbf{144.09}$ \\
rotate $a_0{=}10^4$ & $\theta=90^\circ$                 & $2.774\times10^{9}$ & $5.674\times10^{0}$ & $0.0000$ & $1.36\times10^{-6}$ & $1.000$ \\
\bottomrule
\end{tabular}
\end{table}

\paragraph{Results.}
Within the
$a_0 = 10^4$ rotation arm, increasing the rotation angle drives $\mathrm{Skeel}(M_S)$ from
$5.7$ to $7.5\times10^{8}$, $e_{\mathrm{num}}$ from the noise floor to $26.4$,
and $r_{32}$ from unity to $144$ (Table~\ref{tab:hsic_frames}): a rotation of $5.7^\circ$ multiplies the
single-precision rollout error by a factor of $144$ while leaving the predictor
mathematically untouched. That the mechanism is cancellation, and not
some incidental property of large frames, is confirmed by the anisotropy sweep:
at fixed $\theta$, over the seven consecutive-$a_0$ pairs whose $e_{\mathrm{num}}$
lies between $10^{-5}$ and the mean state magnitude at both ends, each decade of
$a_0$ multiplies $e_{\mathrm{num}}$ by a factor in $[78, 118]$, matching the
$a_0^{2}$ ratio predicted by \eqref{eq:cancel_ratio}. Pairs outside that window
are floored at one end or saturated at the other. Across the squeeze arm, $\kappa$ marches over ten decades from $2.8\times10^{1}$
to $2.8\times10^{11}$ while $\mathrm{Skeel}(M_S)$ remains constant at $5.674$, $\mathcal{L}_{\mathrm{HSIC}}$ is exactly zero, and $|r_{32}-1| \le
4.4\times10^{-7}$ throughout: a large $\kappa$ does not imply harm. Within each rotation
arm $\kappa$ is pinned to fifteen significant figures while $r_{32}$ varies
freely, over $[1.000, 144.09]$ in the $a_0 = 10^4$ arm: nor does harm require $\kappa$ to change.

Monitoring the spectral condition
number would therefore have given both false alarms and silent failures. Of the $46$ frames, $26$ are harmful at link~1, $23$ at link~2 and $6$ at link~3;
$\mathcal{L}_{\mathrm{HSIC}}$ is non-zero on every one of them, at all three links,
with no exceptions within this family (Fig.~\ref{fig:hsic_frames}). Conversely, the $15$ frames on which the penalty is
exactly zero are harmless at all three links: among them the largest
$e_{\mathrm{num}}$ is $1.36\times10^{-6}$, indistinguishable from the identity
frame's single-precision floor, and the largest $|r_{32}-1|$ is
$4.4\times10^{-7}$. Those $15$ frames span ten decades of $\kappa$, so the
penalty's silence is a statement about the frame's benignity and not about its
magnitude. This is exactly the asymmetry a training penalty requires: a
regulariser that is occasionally silent on a harmful configuration provides no
gradient at the moment it is needed, whereas one that is never silent supplies
signal whenever the failure mode is present.

Two further properties are visible in the sweep. First, the penalty saturates: across the frames harmful at link~2 it lies within $[0.0849, 0.0941]$ while
$e_{\mathrm{num}}$ spans six decades. HSIC is thus a detector of the failure mode rather than a gauge of its severity. Deep inside the bad region the term is correspondingly flat
and supplies little gradient; what makes it usable as a penalty is that it rises at
the onset, before the geometry has degraded: $25$ frames have non-zero HSIC while still below the
link-3 threshold, the smallest such value being $2.1\times10^{-3}$ at
$\tilde\lambda = 1$, where the rollout is unaffected to seven digits. The lead is
longest at link~3 and shortest at link~1, where only $5$ frames carry a non-zero
penalty while still below threshold: the penalty and the conditioning it targets
turn on together, and it is the damage that lags. For a
penalty this is the desirable direction of error, since it begins opposing the
geometry at the onset of dependence rather than after the accuracy has already
degraded.

Together the three links make the case for the term. The frame is not selected by the prediction objective, which \eqref{eq:frame_cancel} leaves unchanged under any of these reparameterisations, so the choice among frames falls to something else, and the sweep shows the choice is not immaterial. The examined alternatives do not
serve: $\kappa$ gives both false alarms and silent failures, and
$\mathrm{Skeel}(M_S)$, which does track the failure mode, as a training penalty would require the per-state
Jacobian, at a cost of $2n$ additional backward passes per batch element, and an
explicit inverse. Meanwhile $\mathcal{L}_{\mathrm{HSIC}}$ is a single batch statistic of the encoder outputs already computed, and across this sweep it is non-zero on every frame $\mathrm{Skeel}(M_S)$ flags. The prediction loss itself is not a substitute for such a term. It is invariant under a
change of frame in exact arithmetic, so any sensitivity it has to the frame comes
from rounding error alone. That sensitivity
appears only once the conditioning is poor enough for rounding error to survive
the rollout. In the $a_0=10^{4}$ rotation arm of Appendix~\ref{app:hsic_frame_sweep},
componentwise conditioning degrades by five decades before the single-precision
rollout error departs from its baseline by as much as $0.2\%$; and that rollout
compounds three encodings, whereas a training prediction loss involves a single
encode--decode, so the loss is exposed to less of this amplification still.  It therefore supplies no usable signal on the frame until the conditioning has already degraded far. Training and inference are moreover not exposed to
this frame-induced error equally: a training prediction loss involves a single encode--decode within one
segment, whereas in inference, a rollout over a horizon of $K$ segments re-encodes $K$ times,
each re-encoding taking as input the output of the previous one
(Section~\ref{sec:autoregressive}). An amplification too small in
training can therefore accumulate over the handoffs and become consequential at
inference, so the frame cannot be left to be selected by the training loss even
in the regime where that loss is not entirely flat.

\paragraph{Training without the penalty.}
The sweep varies the frame of one trained model. As a complementary check, we
retrain from the same seed and initialisation with $\lambda_{\mathrm{HSIC}}=0$
and evaluate the result at $S=\mathrm{id}$, on the same pool, horizon and
protocol, computing $\mathrm{Skeel}(M)$, $\mathcal{L}_{\mathrm{HSIC}}$ and
$e_{\mathrm{num}}$ as above. We omit $\kappa$, whose role above is that of a
control within a model rather than a quantity with a cross-model reading, and
$r_{32}$, which is defined against an arm baseline frame of the same model and
has no counterpart between two trained models; link~3 is therefore not assessed
here. The unpenalised model's double-precision rollout MAE at $3T_{\max}$ is
$0.187$ against a mean state magnitude of $0.777$, so $e_{\mathrm{num}}$ is
resolvable as before. Table~\ref{tab:hsic_ablation} reports both models at
$S=\mathrm{id}$.

\begin{table}[H]
\centering
\footnotesize
\caption{Penalised and unpenalised models at $S=\mathrm{id}$, same pool and
protocol as Table~\ref{tab:hsic_frames}. Medians (conditioning) or means
(errors) over $80$ trajectories.}
\label{tab:hsic_ablation}
\setlength{\tabcolsep}{5pt}
\begin{tabular}{lccc}
\toprule
Model & Skeel & $\mathcal{L}_{\mathrm{HSIC}}$ & $e_{\mathrm{num}}$ \\
\midrule
$\lambda_{\mathrm{HSIC}}>0$ & $5.674\times10^{0}$ & $0.0000$ & $1.28\times10^{-6}$ \\
$\lambda_{\mathrm{HSIC}}=0$ & $6.129\times10^{0}$ & $0.0125$ & $2.00\times10^{-6}$ \\
\bottomrule
\end{tabular}
\end{table}

Training without the penalty left the encoder in a frame carrying
measurable within-pair dependence, yet one that meets neither harm criterion
applicable across models: its Skeel number is $1.08$ times and its
$e_{\mathrm{num}}$ $1.56$ times the penalised model's, against thresholds of
$10$ and $2$. A non-zero penalty on a harmless frame is consistent with the results of the
above: no frame exceeding a harm threshold had a vanishing penalty, while
several frames on which the penalty was non-zero stayed below all three. On
this seed, then, the penalty was not required for training to reach a
well-conditioned frame, consistent with our claim that the penalty is a safeguard rather
than a requirement.

\section{Results without the independence term}
\label{zero_hsic_experiments}

$\mathcal{L}_{\mathrm{HSIC}}$ acts directly on the latents, so the properties
that Section~\ref{sec:eval_protocol} attributes to the prediction objective could
in principle be its product. To test this, we retrain HJN on Tasks~1, 2 and~4
with $\lambda_{\mathrm{HSIC}}=0$, denoted HJN ($\lambda_{\mathrm{HSIC}}{=}0$),
keeping the architecture, data, seed and checkpoint-selection criterion
(Appendix~\ref{app:validation_criterion}) of the seed-42 run. The epoch budgets
of Table~\ref{tab:hparams_train} were selected on validation loss for the
objective that includes the term. Without the term the optimisation proceeds
differently, and a budget that ends before the retrained model reaches its
validation optimum could degrade its results through under-training, a
degradation that could be mistaken for an effect of removing the term. We
therefore train for $600$, $1200$ and $2200$ epochs on Tasks~1, 2 and~4
respectively, each above the corresponding budget of
Table~\ref{tab:hparams_train}.

We repeat the experiments of Section~\ref{sec:experiments} except the
single-trajectory diagnostics of Section~\ref{sec:prediction_accuracy}, which
follow a single representative trajectory in time and would be expected to add
little to the attribution beyond the aggregate comparison, which spans all test
trajectories and observation choices. The time-shift test uses, for both models, the read-out
degrees selected for the seed~$42$ run (energy: $2$ for the oscillator, $3$ for
the pendulum, $4$ for H\'enon--Heiles; phase, integrable systems only: $1$), and
Table~\ref{tab:nohsic_recovery} reports the recovery fits of both models at the
same degrees, so that the two models are evaluated identically.

\begin{table}[htbp]
\centering
\footnotesize
\caption{Within-segment constancy as in Tables~\ref{tab:constancy_D0} and~\ref{tab:constancy_abs}, for HJN (seed~42) and HJN ($\lambda_{\mathrm{HSIC}}{=}0$): median $[\,p_{25},p_{75}\,]$ over trajectories of $D_0$ and $\delta_{\mathrm{rms}}$. Rows T$n$-Train, T$n$-A and T$n$-B denote Task~$n$ on its training trajectories, Test~A and Test~B.}
\label{tab:nohsic_consistency}
\setlength{\tabcolsep}{3pt}
\begin{tabular}{lcccc}
\toprule
& \multicolumn{2}{c}{$D_0$} & \multicolumn{2}{c}{$\delta_{\mathrm{rms}}$} \\
\cmidrule(lr){2-3}\cmidrule(lr){4-5}
Task & Seed~42 & $\lambda_{\mathrm{HSIC}}{=}0$ & Seed~42 & $\lambda_{\mathrm{HSIC}}{=}0$ \\
\midrule
T1-Train & $0.0235\,[0.0219,\,0.0307]$ & $0.0215\,[0.0189,\,0.0267]$ & $0.0833\,[0.0718,\,0.0988]$ & $0.0731\,[0.0684,\,0.0776]$ \\
T1-A     & $0.0232\,[0.0214,\,0.0275]$ & $0.0235\,[0.0193,\,0.0276]$ & $0.0791\,[0.0657,\,0.0957]$ & $0.0740\,[0.0655,\,0.0803]$ \\
T1-B     & $0.0358\,[0.0269,\,0.0573]$ & $0.0380\,[0.0268,\,0.0548]$ & $0.212\,[0.145,\,0.324]$ & $0.203\,[0.150,\,0.306]$ \\
\midrule
T2-Train & $0.0381\,[0.0272,\,0.0518]$ & $0.0403\,[0.0340,\,0.0539]$ & $0.175\,[0.153,\,0.222]$ & $0.200\,[0.155,\,0.237]$ \\
T2-A     & $0.0379\,[0.0333,\,0.0491]$ & $0.0413\,[0.0340,\,0.0526]$ & $0.187\,[0.175,\,0.233]$ & $0.199\,[0.182,\,0.230]$ \\
T2-B     & $0.203\,[0.159,\,0.323]$ & $0.218\,[0.177,\,0.344]$ & $1.50\,[1.04,\,2.22]$ & $1.55\,[0.995,\,2.37]$ \\
\midrule
T4-Train & $0.0673\,[0.0616,\,0.0839]$ & $0.0664\,[0.0571,\,0.0909]$ & $0.303\,[0.268,\,0.403]$ & $0.316\,[0.245,\,0.435]$ \\
T4-A     & $0.0908\,[0.0640,\,0.136]$ & $0.101\,[0.0721,\,0.144]$ & $0.451\,[0.274,\,0.659]$ & $0.503\,[0.320,\,0.688]$ \\
T4-B     & $0.308\,[0.194,\,0.382]$ & $0.300\,[0.189,\,0.389]$ & $1.80\,[1.01,\,2.01]$ & $1.82\,[0.984,\,2.11]$ \\
\bottomrule
\end{tabular}
\end{table}

\begin{table}[htbp]
\centering
\footnotesize
\caption{Recovery of $(E,\phi)$ and time-shift test for HJN (seed~$42$) and
HJN ($\lambda_{\mathrm{HSIC}}{=}0$), with the metrics of
Tables~\ref{tab:energy_recovery}--\ref{tab:anharmonic_curve} at the read-out degrees
selected on seed~$42$ (Section~\ref{sec:latent_recovery}). Recovery rows: $5$-fold
cross-validated $R^2$ and median error at $t=0$, with $|\Delta E|/E$ in \%. Time-shift
rows: expected value in parentheses; $|\hat E-E|/E$ (\%) is the per-$\tau$ median across
states summarised over $\tau$, the other rows are summarised across states. Bracketed
entries are median $[p_{25},p_{75}]$.}
\label{tab:nohsic_recovery}
\setlength{\tabcolsep}{4pt}
\begin{tabular}{lcc}
\toprule
& HJN (seed 42) & HJN ($\lambda_{\mathrm{HSIC}}{=}0$) \\
\midrule
\multicolumn{3}{l}{\emph{Ideal mass--spring}} \\
Energy (deg.~2): $R^2$; $|\Delta E|/E$ (\%) & 0.963; 6.1 & 0.922; 8.3 \\
Phase (deg.~1): $R^2$ ($\cos$, $\sin$) & 0.998, 0.999 & 0.991, 0.997 \\
Phase (deg.~1): $\Delta\phi$ (deg) & 1.83 & 2.71 \\
$\mathrm{d}\hat E/\mathrm{d}\tau$ ($0$) & 0.001 [$-$0.003, 0.003] & $-$0.001 [$-$0.004, 0.004] \\
$|\hat E-E|/E$ ($0$) & 5.0 [4.7, 5.5] & 8.1 [7.7, 8.8] \\
$\mathrm{d}\hat\phi/\mathrm{d}\tau$ ($-1$) & $-$0.998 [$-$1.004, $-$0.991] & $-$0.998 [$-$1.008, $-$0.989] \\
Residual (deg) ($0$) & 1.80 [0.87, 3.10] & 2.53 [1.24, 4.29] \\
Intercept error (deg) ($0$) & 1.33 [0.58, 2.07] & 1.91 [0.91, 3.17] \\
\midrule
\multicolumn{3}{l}{\emph{Ideal pendulum}} \\
Energy (deg.~3): $R^2$; $|\Delta E|/E$ (\%) & 0.954; 2.5 & 0.912; 3.6 \\
Phase (deg.~1): $R^2$ ($\cos$, $\sin$) & 0.995, 0.996 & 0.997, 0.997 \\
Phase (deg.~1): $\Delta\phi$ (deg) & 2.50 & 2.41 \\
$\mathrm{d}\hat E/\mathrm{d}\tau$ ($0$) & $-$0.009 [$-$0.018, 0.016] & $-$0.002 [$-$0.023, 0.024] \\
$|\hat E-E|/E$ ($0$) & 3.9 [3.2, 5.3] & 5.7 [5.2, 6.4] \\
$|\mathrm{d}\hat\phi/\mathrm{d}\tau+\Omega(E)|$ ($0$) & 0.020 [0.010, 0.032] & 0.019 [0.009, 0.029] \\
Residual (deg) ($0$) & 2.05 [1.02, 3.46] & 1.99 [0.97, 3.48] \\
Intercept error (deg) ($0$) & 1.39 [0.68, 2.33] & 1.46 [0.88, 2.57] \\
Bin-median slope, lowest$\to$highest bin ($-1.623\to-1.551$) & $-1.592\to-1.581$ & $-1.598\to-1.583$ \\
\midrule
\multicolumn{3}{l}{\emph{H\'enon--Heiles}} \\
Energy (deg.~4): $R^2$; $|\Delta E|/E$ (\%) & 0.944; 2.1 & 0.939; 2.1 \\
$\mathrm{d}\hat E/\mathrm{d}\tau$ ($0$) & 0.0000 [$-$0.0003, 0.0003] & 0.0000 [$-$0.0002, 0.0002] \\
$|\hat E-E|/E$ ($0$) & 2.4 [2.2, 2.8] & 2.3 [2.1, 2.5] \\
\bottomrule
\end{tabular}
\end{table}

\begin{table}[htbp]
\centering
\footnotesize
\caption{Prediction accuracy, as in Table~\ref{tab:main_results}: cross-trajectory
mean of per-trajectory median $\mathrm{L}_2/\sqrt{E}$, $95\%$ BCa bootstrap
confidence interval in brackets ($10{,}000$ resamples).}
\label{tab:nohsic_prediction}
\setlength{\tabcolsep}{4pt}
\begin{tabular}{lcc}
\toprule
Setting & HJN (seed 42) & HJN ($\lambda_{\mathrm{HSIC}}{=}0$) \\
\midrule
T1-A & 0.083 [0.080, 0.087] & 0.076 [0.073, 0.080] \\
T1-B & 0.064 [0.056, 0.074] & 0.061 [0.051, 0.074] \\
T2-A & 0.162 [0.131, 0.195] & 0.160 [0.135, 0.191] \\
T2-B & 0.316 [0.248, 0.385] & 0.307 [0.246, 0.378] \\
T4-A & 0.508 [0.396, 0.648] & 0.526 [0.408, 0.670] \\
T4-B & 0.575 [0.450, 0.757] & 0.594 [0.469, 0.756] \\
\bottomrule
\end{tabular}
\end{table}

\paragraph{Results.}

For HJN ($\lambda_{\mathrm{HSIC}}{=}0$) (Table~\ref{tab:nohsic_consistency}), the median $D_0$ and $\delta_{\mathrm{rms}}$ fall inside the range of the three main seeds' medians (Tables~\ref{tab:constancy_D0}--\ref{tab:constancy_abs}) in $12$ of $18$ cases (three tasks, three sets, two metrics). In $5$ cases they are lower than all three seeds, and in $1$ case higher ($\delta_{\mathrm{rms}}$ on T4-B: $1.82$ against $1.80$). For
HJN ($\lambda_{\mathrm{HSIC}}{=}0$), the degree-selection rule of
Section~\ref{sec:latent_recovery} picks the same energy and phase degrees as for
seed~$42$ on the integrable systems, and a lower energy degree on
H\'enon--Heiles, $3$ ($R^2=0.923$) against $4$, as it also does for seed~$52$
(Section~\ref{sec:latent_recovery}). Between the two models, the median energy slope
$\mathrm{d}\hat E/\mathrm{d}\tau$ differs by at most $0.007$ across the three
systems, while the median phase slope on the oscillator and the median
deviation from $-\Omega(E)$ on the pendulum each differ by at most $0.001$
(Table~\ref{tab:nohsic_recovery}). The confidence
intervals of prediction accuracy overlap in five of the six settings; on T1-A
they share only their endpoint at the reported precision ($0.080$), with the
lower error obtained by HJN ($\lambda_{\mathrm{HSIC}}{=}0$)
(Table~\ref{tab:nohsic_prediction}). In both within-segment consistency and prediction accuracy the change has no
common sign across tasks, so removing the term neither systematically degrades
nor systematically improves them. The ablation uses a single seed, so the differences between the two models
cannot be separated from seed-to-seed variation; within that limitation, the
results of Section~\ref{sec:experiments} examined here show no dependence on
$\mathcal{L}_{\mathrm{HSIC}}$.

\section{Ablation: sensitivity to segment length}
\label{sec:tmax_sweep}

The segment length $T_{\max}$ is fixed before optimisation by the rule of
Appendix~\ref{sec:sampling_limitation}, and we do not claim that the value
it returns is optimal. This appendix measures how strongly the reported
results depend on that value. We ablate on Task~4 for two reasons. It is the
hardest system in the suite, being non-integrable and chaotic, and it is the
only task within scope for which the estimator returns a $T_{\max}$ shorter
than the record length, so that segmentation is actually exercised:
Tasks~1 and~2 are trained on full trajectories, and Task~3 lies outside the
scope of the method. Around the selected value $T_{\max}^{\star}=8.563$ we retrain at four
additional segment lengths: $T_{\max}^{\star}/4$, $T_{\max}^{\star}/2$,
$2T_{\max}^{\star}$, and $T_{\max}=50$. The last equals the full record
length, so that each trajectory forms a single segment. Every configuration
segments the same training trajectories by the same procedure as the
reported run; only $T_{\max}$ changes. The resulting segmentations are
summarised in Table~\ref{tab:tmax_ablation_setup}.

Under the training rule of Section~\ref{sec:training}, an epoch consists
of $L_{\max}$ batches, where $L_{\max}$ is the longest segment. An epoch is therefore not a fixed unit of work: $T_{\max}$
determines the segmentation, hence $L_{\max}$, hence the number of optimiser
updates per epoch, which varies by a factor of about $22$ across the
configurations. Holding the number of epochs fixed would thus confound
segment length with training budget. We instead fix the number of optimiser
steps at $122{,}438$, the step at which the seed-42 run attained its
selected checkpoint (epoch $2111$ at $L_{\max}=58$), and set
$\mathrm{epochs}=\lfloor 122{,}438/L_{\max}\rceil$ for each configuration,
where $\lfloor\cdot\rceil$ denotes rounding to the nearest integer. Since
the seed-42 run was trained under a larger epoch cap and its selected
checkpoint was interior, this is a budget at which at least one member of
the family is known to have stopped improving. The same consideration applies to the learning-rate scheduler, which reduces
the learning rate on a validation plateau and is stepped once per epoch. Its
patience is therefore counted in epochs, that is, in multiples of $L_{\max}$
optimiser steps. The seed-42 run used a patience of $200$ epochs, i.e.\
$200\times 58=11{,}600$ steps. We hold this fixed in steps by setting
$\mathrm{patience}=\lfloor 11{,}600/L_{\max}\rceil$, which permits at most
ten reductions within the budget of every run. All remaining hyperparameters, including the
architecture, initialisation and seed, are those of
Appendix~\ref{app:training_details}. No run is early-stopped, and each
retains the checkpoint minimising the validation criterion of
Appendix~\ref{app:validation_criterion}, evaluated on its own segmentation.

\begin{table}[htbp]
\centering
\caption{Task~4 $T_{\max}$ ablation: segmentation, training budget and
scheduler patience. Samples/segment lists the distinct segment lengths, in
samples, that occur in each segmentation; the largest is $L_{\max}$. Epochs
and patience are set so that every run takes $122{,}438$ optimiser steps
with a patience of $11{,}600$ steps.}
\label{tab:tmax_ablation_setup}
\setlength{\tabcolsep}{5pt}
\begin{tabular}{lccccc}
\toprule
$T_{\max}$ & Segments & Samples/segment & $L_{\max}$ & Epochs & Patience \\
\midrule
$2.141$          & 1152 & 15/14/6     & 15  & 8163 & 773 \\
$4.282$          & 576  & 29/28/20    & 29  & 4222 & 400 \\
$8.563^{\star}$    & 288  & 58/57/48    & 58  & 2111 & 200 \\
$17.126$           & 144  & 115/114/105 & 115 & 1065 & 101 \\
$50.000$           & 48   & 334         & 334 & 367  & 35  \\
\bottomrule
\end{tabular}
\end{table}

We repeat the aggregate prediction-accuracy experiment of
Section~\ref{sec:prediction_accuracy} on both Task~4 test sets, following
the protocol described there. In addition, for each observation index we
record the number of network evaluations ($\mathcal{M}$ and
$\mathcal{M}^{-1}$) required to reconstruct the batched trajectories from
that observation (NFE). Table~\ref{tab:tmax_ablation_results} reports the
error together with the range of NFE across observation indices; its
$T_{\max}^{\star}$ column reproduces the seed-42 Task~4 entries of
Table~\ref{tab:main_results}.

\begin{table}[H]
\centering
\footnotesize
\caption{Ablation over the segment length $T_{\max}$ on Task~4. $T_{\max}^{\star}=8.563$ is the segment length of the models reported in
Table~\ref{tab:main_results}. For each test setting, the first row reports
$\mathrm{L}_2/\sqrt{E}$ as the cross-trajectory mean of per-trajectory
median errors, with $95\%$ BCa bootstrap confidence intervals
($10{,}000$ resamples) in brackets; bold marks the lowest mean in each row.
The second row reports the minimum and maximum, across observation indices,
of the number of network evaluations ($\mathcal{M}$ and $\mathcal{M}^{-1}$)
required to reconstruct the batched trajectories from one observation
(NFE); a single value indicates that the minimum and maximum coincide. Rows
T4-A and T4-B denote rollout (A) and energy (B) extrapolation.}
\label{tab:tmax_ablation_results}
\label{tab:tmax_ablation}

\setlength{\tabcolsep}{3pt}

\begin{tabular}{llccccc}
\toprule
Task & & $T_{\max}=2.14075$ & $T_{\max}=4.2815$ & $T_{\max}^{\star}=8.563$ & $T_{\max}=17.126$ & $T_{\max}=50$ \\
\midrule
T4-A & $\mathrm{L}_2/\sqrt{E}$ & \textbf{0.417} [0.312, 0.560] & 0.457 [0.352, 0.595] & 0.508 [0.396, 0.648] & 1.037 [0.941, 1.130] & 1.748 [1.670, 1.797] \\
 & NFE & [48,50] & [24,26] & [12,14] & [6,8] & 2 \\
\addlinespace
T4-B & $\mathrm{L}_2/\sqrt{E}$ & \textbf{0.514} [0.370, 0.729] & 0.541 [0.403, 0.747] & 0.575 [0.450, 0.757] & 0.898 [0.789, 1.036] & 1.099 [0.989, 1.218] \\
 & NFE & [16,18] & [8,10] & [4,6] & 2 & 2 \\
\bottomrule
\end{tabular}
\end{table}
Comparing the differences with the across seed variation: shortening $T_{\max}$ below $T_{\max}^{\star}$ lowers the mean error by at most $0.091$ on T4-A and $0.061$ on T4-B (Table~\ref{tab:tmax_ablation_results}), less than the difference between the largest and smallest of the three seeds' means at $T_{\max}^{\star}$, $0.244$ and $0.118$ (Table~\ref{tab:main_results}); lengthening it raises the error to at least $1.037$ and $0.898$, above the largest of those means, $0.752$ and $0.693$.

\section{Inference runtime and GPU memory}
\label{sec:inference_efficiency}
We measured wall-clock inference runtime and peak GPU memory of the models
trained with seed~42 on an NVIDIA GeForce RTX 3060 (12\,GB). Both methods were evaluated in
strict FP32 precision with TF32 disabled, so that the comparison is unaffected by
framework-dependent mixed-precision acceleration. On the PyTorch side, cuDNN
autotuning (\texttt{cudnn.benchmark}) was enabled and restricted to
deterministic algorithms (\texttt{cudnn.deterministic}). Runtime was measured with \texttt{time.perf\_counter()} and explicit device
synchronisation before and after each measurement (\texttt{torch.cuda.synchronize()}
for HJN, \texttt{jax.block\_until\_ready()} for the reference), ensuring that
asynchronous kernel launches are fully accounted for. Each benchmark comprises $20$ timed evaluation sweeps. A sweep iterates over
all observation indices $\{1,\dots,N_{\mathrm{obs}}\}$ and executes inference
at a given index as a single batched call over all $N_{\mathrm{traj}}$
trajectories, so the resulting runtimes are batched-call costs rather than
per-trajectory latency. To prevent initialisation, autotuning, and compilation
overheads from contaminating the measurements,each sweep is preceded by an
untimed inference call whose runtime is discarded. Each timed sweep runs in a
separate operating-system process, and the $20$ timed sweeps are preceded by
one additional, discarded sweep that populates caches persisting across
processes, notably the JAX compilation cache (location fixed via
\texttt{JAX\_COMPILATION\_CACHE\_DIR}). For each observation index, runtime is averaged across the $20$ timed sweeps.
The runtime values reported in Table~\ref{tab:timing_results} are the mean of
these per-observation runtimes, and the bracketed values give the observed
range $[\min,\max]$ across observation indices. Peak GPU memory is measured as the maximum incremental memory allocation
during each timed sweep relative to the allocation immediately preceding the
sweep. For the reference, JAX's default GPU memory preallocation was disabled
(\texttt{XLA\_PYTHON\_CLIENT\_PREALLOCATE=false}) so that allocation is
incremental and this quantity is well defined. The memory values reported in Table~\ref{tab:timing_results} are the mean
across the $20$ sweeps. We note that the two implementations differ in framework and compilation: the reference is JIT-compiled using JAX (\texttt{@jit}), while HJN is executed in eager-mode PyTorch.\footnote{Efforts to enable compilation via \texttt{torch.compile} (PyTorch~2.5.1) failed because higher-order autodifferentiation operations (\texttt{torch.autograd.grad}) in the forward pass are not yet supported.} The runtimes are therefore not like-for-like.

\begin{table}[H]
\centering
\footnotesize
\caption{
Inference runtime and peak incremental GPU memory on the test sets.
Reported runtimes are mean per-observation batched-call runtimes in
milliseconds, with ranges across observation indices shown in brackets.
GPU memory is the maximum incremental allocation within a sweep, averaged
over the $20$ timed sweeps. Rows Tn-A and Tn-B denote Task~$n$
under rollout (A) and energy (B) extrapolation. Ref: the reference, included to give the runtime a scale.}

\label{tab:timing_results}

\setlength{\tabcolsep}{3pt}

\begin{tabular}{lcccc}
\toprule
& \multicolumn{2}{c}{\begin{tabular}[c]{@{}c@{}}Runtime\\(ms)\end{tabular}}
& \multicolumn{2}{c}{\begin{tabular}[c]{@{}c@{}}GPU mem.\\(MB)\end{tabular}} \\
Task
& HJN & Ref
& HJN & Ref \\
\midrule

T1-A
& 94.29 [91.08, 115.14]
& 6713.86 [3567.73, 8094.87]
& 2.8
& 0.2
\\

T1-B
& 20.98 [20.74, 21.81]
& 5982.31 [2953.55, 6916.90]
& 2.0
& 0.1
\\

T2-A
& 129.33 [109.51, 132.17]
& 7105.70 [3952.03, 9138.52]
& 3.0
& 0.4
\\

T2-B
& 49.01 [42.11, 63.43]
& 5346.31 [2581.46, 6175.17]
& 1.7
& 0.1
\\

T4-A
& 155.17 [136.58, 195.61]
& 8907.08 [5326.24, 12692.90]
& 5.4
& 0.9
\\

T4-B
& 65.37 [45.53, 66.56]
& 6449.69 [3381.70, 7921.40]
& 3.1
& 0.3
\\

\bottomrule
\end{tabular}

\end{table}

\section{Software and Hardware}
\label{app:compute}

All evaluation runs reported in this paper were executed
on a single NVIDIA GeForce RTX 3060 (12\,GB VRAM) with
CUDA version 12.4, PyTorch version 2.5.1~\cite{pytorch_sw,paszke2019pytorch} for our method and JAX/jaxlib version 0.10.0~\cite{jax2018github} for the reference. Training wall-clock times and peak
GPU memory during training were not systematically logged. All code is available at \url{https://github.com/thomasGR9/Hamilton_Jacobi_NN}~\cite{hjn_code}.

\bibliographystyle{unsrtnat}
\bibliography{references}

\end{document}